\documentclass{article}

\usepackage[preprint,nonatbib]{neurips_2025}

\usepackage[utf8]{inputenc} % allow utf-8 input
\usepackage[T1]{fontenc}    % use 8-bit T1 fonts
\usepackage{hyperref}       % hyperlinks
\usepackage{url}            % simple URL typesetting
\usepackage{booktabs}       % professional-quality tables
\usepackage{amsfonts}       % blackboard math symbols
\usepackage{nicefrac}       % compact symbols for 1/2, etc.
\usepackage{microtype}      % microtypography
\usepackage{float}
\usepackage{graphicx}
\usepackage{tabularx}  
\usepackage[table,dvipsnames,svgnames,x11names]{xcolor}  % 合并所有xcolor选项

\usepackage{makecell}
\usepackage{enumitem}
\usepackage{mdframed}

\usepackage{subcaption}
\usepackage{bbding}
\usepackage{soul}
\usepackage{longtable}

\usepackage{wrapfig}
\usepackage{bbding}
\usepackage{soul}
\definecolor{softblue}{RGB}{230,230,230}
\definecolor{codegreen}{rgb}{0,0.6,0}
\definecolor{codegray}{rgb}{0.5,0.5,0.5}
\definecolor{codepink}{RGB}{252, 142, 172}
\definecolor{codepurple}{rgb}{0.58,0,0.82}
\definecolor{backcolour}{RGB}{245,245,245}
\definecolor{darkblue}{rgb}{0, 0, 0.5}
\hypersetup{colorlinks=true, citecolor=darkblue, linkcolor=darkblue, urlcolor=darkblue, linktoc=page}

\usepackage{listings}
\lstdefinestyle{finmaster_style}{
    backgroundcolor=\color{backcolour},   
    commentstyle=\color{magenta},
    keywordstyle=\color{blue},
    numberstyle=\tiny\color{codegray},
    stringstyle=\color{codepurple},
    basicstyle=\fontfamily{\ttdefault}\footnotesize,
    breakatwhitespace=false,        
    breaklines=true,                
    keepspaces=true,    
    frame=single,
    numbersep=5pt,                  
    showspaces=false,              
    showstringspaces=false,
    showtabs=false,               
    tabsize=2,
    classoffset=1, % starting new class
    keywordstyle=\color{violet},
    classoffset=0,
    xleftmargin=0.03\textwidth,
    xrightmargin=0.03\textwidth
}
\usepackage[numbers]{natbib}
\usepackage{amsmath}
\usepackage{amssymb}
\usepackage{mathtools}
\usepackage{amsthm}
\usepackage{tcolorbox}
\usepackage{bm}
\usepackage{multirow}

\theoremstyle{plain}

\theoremstyle{definition}

\theoremstyle{remark}

\title{FinMaster: A Holistic Benchmark for Mastering Full-Pipeline Financial Workflows with LLMs}

\author{
Junzhe Jiang$^{1,}$\thanks{Equal contribution, alphabetical order}~, Chang Yang$^{1,}$\footnotemark[1]~, Aixin Cui$^{2}$, Sihan Jin$^{3}$,  Ruiyu Wang$^{4}$, \\ 
\textbf{Bo Li$^{1}$, Xiao Huang$^{1}$, Dongning Sun$^{5,}$\thanks{Corresponding author}~, Xinrun Wang$^{6,}$\footnotemark[2]}\\
$^1$The Hong Kong Polytechnic University, $^2$The Chinese University of Hong Kong
, \\ $^3$The Hong Kong University of Science and Technology, $^4$KTH Royal Institute of Technology, \\ $^5$Peng Cheng Laboratory, $^6$Singapore Management University
 \\
\texttt{xrwang@smu.edu.sg}}

\begin{document}

\maketitle

\begin{abstract}

Financial tasks are pivotal to global economic stability; however, their efficient execution faces persistent challenges, including labor intensive processes, low error tolerance, data fragmentation, and limitations in existing technological tools. Although large language models (LLMs) have shown remarkable success in various natural language processing (NLP) tasks and have shown potential in automating workflows through reasoning and contextual understanding, current benchmarks for evaluating LLMs in finance suffer from insufficient domain-specific data simplistic task design, and incomplete evaluation frameworks. To address these gaps, this article presents \textbf{FinMaster}, a comprehensive financial benchmark designed to systematically assess the capabilities of LLM in financial literacy, accounting, auditing, and consulting. Specifically, \textbf{FinMaster} comprises three main modules: i) \emph{FinSim}, which builds simulators that can generate synthetic, privacy-compliant financial data sets for different types of companies to replicate real-world market dynamics; ii) \emph{FinSuite}, which provides a variety of tasks in core financial domains, spanning 183 tasks of various types and difficulty levels; and iii) \emph{FinEval}, which develops a unified interface for streamlined evaluation. Extensive experiments over state-of-the-art LLMs, including GPT-4o-mini, Claude-3.7-Sonnet, and DeepSeek-V3, reveal critical capability gaps in financial reasoning, with accuracy dropping from over 90\% on basic tasks to merely 40\% on complex scenarios requiring multi-step reasoning. This degradation exhibits the propagation of computational errors, where single-metric calculations initially demonstrating 58\% accuracy decreased to 37\% in multimetric scenarios. To the best of our knowledge, \textbf{FinMaster} is the first benchmark that comprehensively covers full-pipeline financial workflows with challenging and realistic tasks. We hope that \textbf{FinMaster} can bridge the gap between the research community and industry practitioners, driving the adoption of LLMs in real-world financial practices to enhance both efficiency and accuracy.

% As the first benchmark that simulates multistep financial workflows beyond simple question-answering (QA) tests, \textbf{FinMaster} presents more challenging and realistic tasks requiring sophisticated financial reasoning. By providing guidance aligned with actual industry challenges and a robust foundation for financial AI advancement, \textbf{FinMaster} encourages development that addresses the core complexities of professional financial practice.

% This paper presents \textbf{FinMaster}, a financial benchmark for LLMs. Specifically, \textbf{FinMaster} has three main components: i) \emph{FinSim}: which builds a simulator to generate financial tasks for accounting, auditing and consulting; ii) \emph{FinSuite}: which provides a unified interface to evaluate the tasks with API-based online models; and iii) \emph{FinEval}: which provides a evaluation suite to analyze the performances of LLMs over different tasks.
\end{abstract}
\vspace{-10pt}
\section{Introduction}
\vspace{-5pt}
Finance serves as the cornerstone of the global economic system, where financial tasks are critical for ensuring accuracy, compliance, and efficiency in economic activities, e.g., capital allocation, risk management, and investment strategic decision-making. However, their execution faces significant challenges: i) labor intensive: traditional financial tasks such as accounting and auditing rely heavily on manual operations, and financial professionals require extensive training to master complex regulations, which are repetitive and time-consuming; ii) low error tolerance: minor mistakes in financial statements, e.g., decimal errors or misclassifications, can trigger compliance risks or market volatility; iii) data fragmentation:  financial data always originates from diverse sources, each with unique data structures and update frequencies. However, the latency in real-time scenarios and the poor compatibility between systems in integrating heterogeneous data may lead to data silos; iv) limitations of technological tools: existing fintech tools and rule-based systems often exhibit limitations in interpreting implicit logic in financial data and adapting to evolving regulations, making them difficult to handle complex causal reasoning tasks. 
The advancements in large language models (LLMs), e.g., GPT-4~\citep{achiam2023gpt} and DeepSeek-v3~\citep{liu2024deepseek}, have demonstrated remarkable successes in tasks requiring reasoning, contextual understanding, and multi-step problem-solving across various domains (e.g., code generation and math reasoning). 
The general-purpose capabilities of LLMs make them well-suited for automating financial workflows, reducing human effort, and minimizing errors.

\begin{wraptable}{r}{0.5\textwidth}
\centering
\vspace{-10pt}
\setlength{\tabcolsep}{3pt} 

\caption{{Comparison of Financial Benchmarks}}
\resizebox{0.5\textwidth}{!}{
\begin{tabular}{c|ccc}
\toprule
Benchmark & \makecell{Complex \\ tasks} & \makecell{
Infinite \\ data} & \makecell{Holistic \\ evaluation}\\
\midrule
FinQA~\citep{chen2021finqa} &
{\color{red!75}\XSolidBrush}
& {\color{red!75}\XSolidBrush} & {\color{red!75}\XSolidBrush} \\
PIXIU~\citep{xie2023pixiu} & {\color{Green3}\Checkmark} & {\color{red!75}\XSolidBrush} & {\color{red!75}\XSolidBrush}

\\
FinanceBench~\citep{islam2023financebench}     & {\color{red!75}\XSolidBrush} &{\color{red!75}\XSolidBrush} & {\color{red!75}\XSolidBrush}\\
% BizBench~\citep{koncel2023bizbench}     & \\
FinBen~\citep{xie2024finben} &  {\color{Green3}\Checkmark}  &{\color{red!75}\XSolidBrush} & {\color{Green3}\Checkmark}\\
FinEval~\citep{zhang2023fineval} &{\color{Green3}\Checkmark} & {\color{red!75}\XSolidBrush} &  {\color{red!75}\XSolidBrush}\\
SECQUE~\citep{yoash2025secque} &{\color{Green3}\Checkmark} & {\color{red!75}\XSolidBrush} & {\color{Green3}\Checkmark}\\
\midrule
\textbf{FinMaster} & {\color{Green3}\Checkmark} & {\color{Green3}\Checkmark} & {\color{Green3}\Checkmark}\\
\bottomrule
\end{tabular}}
\label{tab:benchmark_comparision}
\vspace{-10pt}
\end{wraptable}
Several recent attempts that apply LLMs to finance have demonstrated promising capabilities. 
FinQA~\citep{chen2021finqa} introduces complex numerical reasoning over financial reports but is constrained by static data and structured formats. FinBen~\citep{xie2024finben} offers broader evaluation across diverse financial tasks but lacks granularity for domain-specific reasoning nuances. FinTSB~\citep{hu2025fintsb} specializes in time-series forecasting but neglects external factors. Other benchmarks like PIXIU~\citep{xie2023pixiu}, FinanceBench~\citep{islam2023financebench}, and BizBench~\citep{koncel2023bizbench} demonstrate narrow scope, primarily focusing on conventional financial NLP tasks while overlooking complex financial reasoning and real-world applications. 
SECQUE~\citep{yoash2025secque} advances LLMs evaluation by focusing on practical financial tasks requiring multi-step reasoning, but still relies on a static dataset and fails to fully capture the reality of financial workflows.
These limitations highlight the need for more comprehensive benchmarks that can address the dynamic nature of financial markets, complex reasoning requirements, and practical application scenarios.

% , these existing attempts remain some main issues that constrain their operational viability in finance. i) Simplistic task design: current efforts mainly focus on tasks far from practice. Tasks like auditing multi-year financial statements remain unexplored, making LLMs fail to capture the complexity of real-world financial workflows. ii) Lack of domain-specific data: financial data is inherently heterogeneous, regulated, and sensitive, which limits its availability for developing and evaluating LLMs for financial tasks. iii) Incomplete evaluation frameworks: 
% there is no comprehensive evaluation task suite for financial tasks, hindering the systematic evaluation and practical deployment of LLMs in finance.

% These limitations highlight the need for more comprehensive benchmarks that can address the dynamic nature of financial markets, complex reasoning requirements, and practical application scenarios.

\begin{figure}[t]
    \centering
    \vspace{-20pt}
    \includegraphics[width=\textwidth]{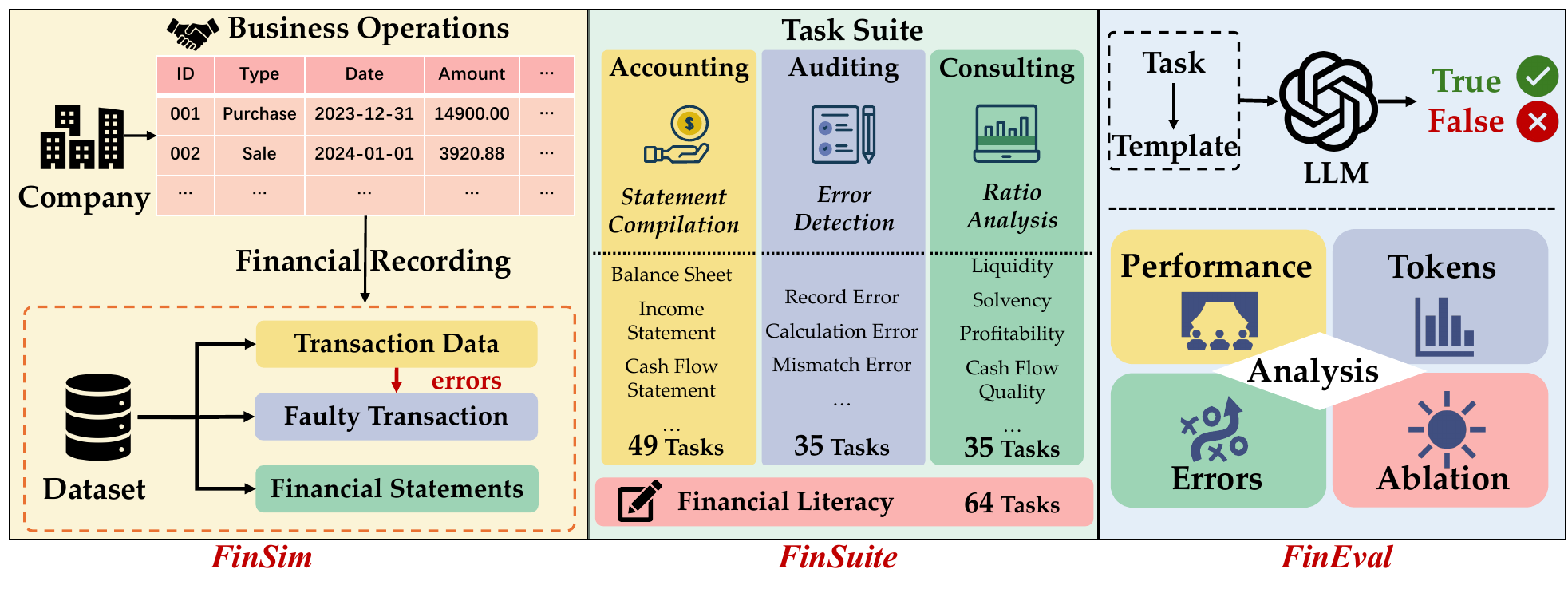}
    \vspace{-20pt}
    \caption{The three main modules of \textbf{FinMaster}.}
    \label{fig:overview}
    \vspace{-20pt}
\end{figure}

To address these issues, we present \textbf{FinMaster}, a holistic benchmark for mastering full-pipeline financial workflows with LLMs. Specifically, \textbf{FinMaster} has three main modules:
i) \emph{FinSim}, which builds a simulator that can automatically generate financial data, including transaction records and financial statements, for financial literacy, accounting, auditing and consulting. It can produce infinite synthetic datasets for various types of companies to replicate real-world market dynamics, addressing the issue of lacking of financial data due to privacy constraints.
ii) \emph{FinSuite}, which provides a task suite, incorporating various types of tasks in the fields of accounting, auditing, and consulting, to comprehensively evaluate the performance and capabilities of LLMs in different financial tasks.
iii) \emph{FinEval}, 
which develops a unified interface, i.e., a streamlined framework, for systematic evaluation of LLMs and provides a comprehensive evaluation that quantifies the performances of online LLMs across different tasks with in-depth analysis such as accuracy and the number of tokens. 

Extensive experiments over widely-used LLMs, e.g., GPT-4o-mini, Claude-3.7-Sonnet, DeepSeek-V3, and OpenAI o3-mini, show the systemic gaps in financial reasoning capabilities.
While these models achieve over 90\% accuracy on basic tasks, their performance drops sharply to just 40\% in complex scenarios that require multi-step reasoning. 
Moreover, general-purpose LLMs lack domain-specific knowledge, leading to hallucinated conclusions or statistically invalid outputs when professional judgment is needed. 
To the best of our knowledge, \textbf{FinMaster} is the first financial benchmark that simulates multi-step financial operations for LLMs, serving as the fundamental testbed for the advanced LLMs. The code is released at \url{https://github.com/JiangInsight/Finmaster} and the leaderboard is shown at \url{https://jianginsight.github.io/Finmaster_leaderboard/}.

\section{Preliminaries}

\textbf{Large Language Models.}
% \paragraph{Large Language Models (LLMs).} 
Large Language Models (LLMs) are autoregressive models that learn from textual data through unsupervised learning, modeling the joint probability of variable-length token sequences by decomposing it into a product of conditional probabilities. 
The factorization is expressed as: $P(x)=\prod\nolimits^n_{i=1}P(s_i|s_1,...,s_{i-1})$, where $(s_1, s_2, ..., s_n)$ represents a sequence of tokens. Trained on vast corpora and parameterized by billions of weights, LLMs encode extensive common knowledge, enabling them to generalize robustly across diverse NLP tasks with simple prompting and in-context learning, and without task-specific fine-tuning~\citep{touvron2023llama,openai2023gpt4}. 

\textbf{Transactions and Financial Statements.}
Transactions refer to the economic event or business activity that result in measurable changes in a company’s financial position and are systematically recorded to populate accurate, reliable financial statements. There are three key financial statements that form the bedrock of financial reporting and analysis: i) income statement, which dynamically reports a company's revenues, expenses, and profits/losses over a specific period;
ii) balance sheet, which provides a snapshot of a company's financial position at a specific moment; and
iii) cash flow statement, which tracks the actual cash movements, i.e., cash inflows and outflows, over a period.
These financial statements are intrinsically interconnected,
where the net income from the income statement flows into the retained earnings on the balance sheet, and the changes in the balance sheet affect the cash flow statement.
Three financial statements are prepared under standardized guidelines, e.g., IFRS~\citep{soderstrom2007ifrs} and GAAP~\citep{epstein2009wiley}, to ensure consistency, transparency, and comparability, providing a comprehensive view of a company's performance and financial stability. Recording transactions and preparing financial statements are legal and managerial obligations that every enterprise must fulfill.

% At its core, accounting serves primarily to track, report, and analyze financial transactions, facilitating informed decision-making \citep{santos2021experimental}.

\textbf{Financial Workflows.}
\textbf{Accounting} is the systematic recording, analysis, and reporting of financial transactions to ensure transparency, compliance, and informed decision-making \citep{godfrey2010accounting}. It supports business sustainability by preparing financial reports and resolving transactions to aid decision-making \citep{santos2020experimental}. Daily operations are recorded, classified, and adjusted to align with accounting principles, e.g., matching income and expenses, forming the basis for financial statements.
% \textbf{Auditing} involves independent examination of financial statements, processes, and compliance. Modern audits extend beyond financial reports to include operational data analysis, assessing internal controls and efficiency using advanced techniques \citep{ovami2023data}.
\textbf{Auditing} is an independent assurance activity that verifies the compliance of a company’s economic activities and the reliability of its financial information through systematic review and professional evaluation. Its core objective is to ensure the accuracy and completeness of financial data, with the verification of transaction data serving as the foundation of the audit process. 
\textbf{Consulting} provides expert analysis to improve business performance \citep{biech2019new, bruhn2018impact}. Financial diagnostics, i.e., evaluating profitability, efficiency, and solvency via frameworks like DuPont Analysis \citep{soliman2008use} and Altman Z-scores \citep{altman2017financial}, identify inefficiencies and competitive positioning. This analysis bridges financial data to strategic decisions, distinguishing consultants as data-driven problem solvers.

\section{FinMaster}

\subsection{\emph{FinSim}: Financial Data Simulator}

\begin{wrapfigure}{r}{0.38\textwidth}
\centering
\vspace{-3cm}
\includegraphics[width=0.38\textwidth]{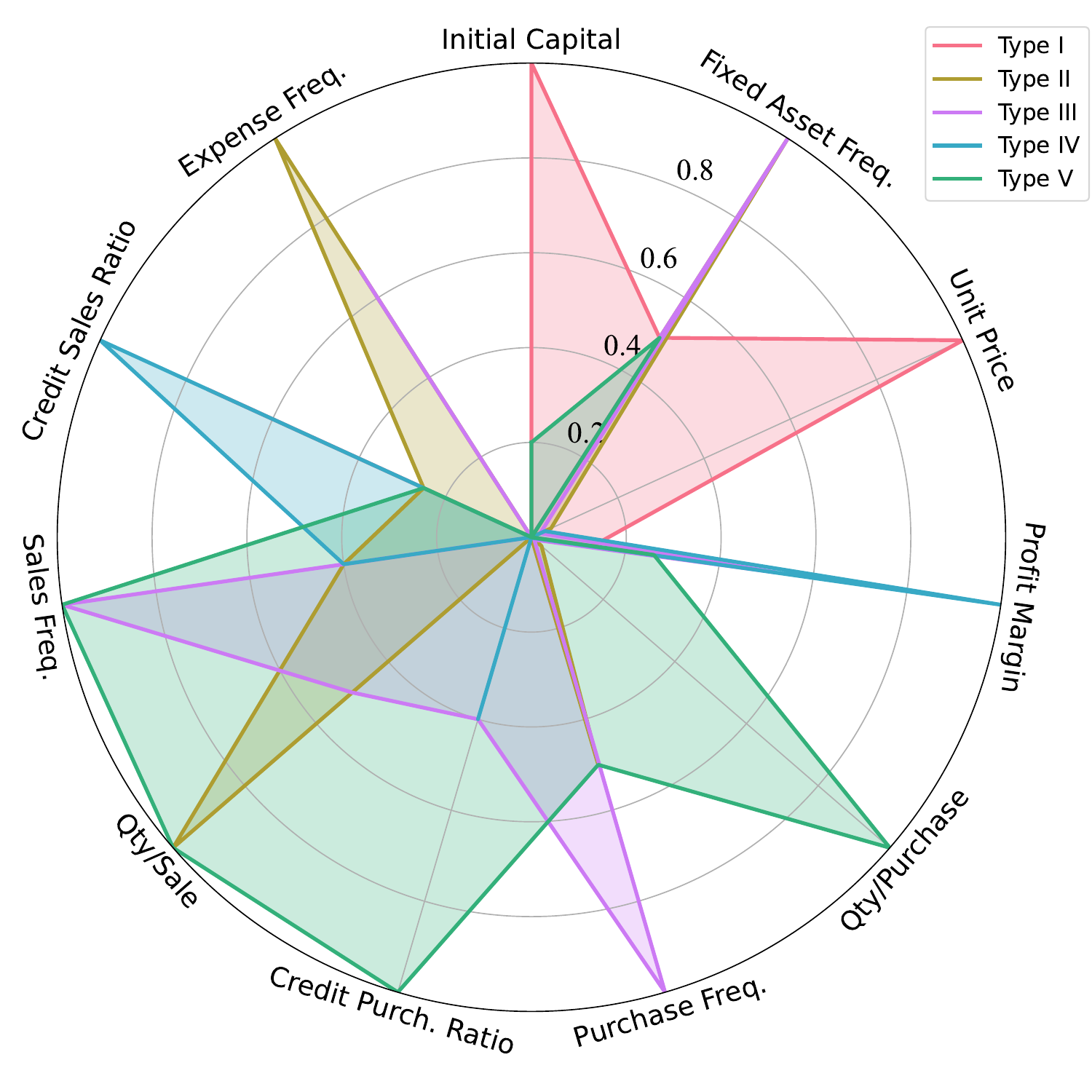}
\vspace{-15pt}
\caption{Companies Comparison}
\label{fig:radar}
\vspace{-20pt}
\end{wrapfigure}
We develop a financial data simulator, which simulates the daily financial activities of different types of companies and generates transaction records and financial statements.

\textbf{Types of Companies.}
To mirror real-world market diversity, \emph{FinSim} incorporates simulators of five distinct company types: Type \uppercase\expandafter{\romannumeral 1} considers capital goods manufacturers with capital-intensive operations, low sales frequency, and high-value products; 
Type \uppercase\expandafter{\romannumeral 2} considers transaction-driven companies with stable costs, low pricing power, and reliance on bulk sales despite high operating expenses; Type \uppercase\expandafter{\romannumeral 3} considers high value-added consumer goods producers with high margins, low production costs, and heavy branding investments; 
Type \uppercase\expandafter{\romannumeral 4} considers asset-light companies with minimal fixed assets, high margins, and procurement via credit; 
Type \uppercase\expandafter{\romannumeral 5} considers high-turnover companies with frequent low-price sales, large volumes, and a broad customer base.
The detailed characteristic financial profiles of different types of companies are shown in Figure~\ref{fig:radar} and Table~\ref{tab:company} in Appendix~\ref{app:company}, comprehensively covering diverse business models across industries.

\textbf{Types of Transactions.}
Several types of transactions or financial records are considered in \emph{FinSim} to simulate the complexity of realistic operations while maintaining the integrity of financial systems. 
i) Asset data: including the initializations of cash deposit, bank deposit, and fixed assets. 
ii) Operational data: including purchase management, which refers to the purchase-related transactions with different payment methods, e.g., cash and bank transfer; sales management, which refers to the sales-related transactions, corresponding to the purchase-related transactions; and fixed assets management, which comprises fixed asset purchase transactions, i.e., recording the fixed asset acquisitions, and fixed asset depreciation, i.e., generating monthly depreciation entries on the 1st of each month. 
iii) Financial data: including cash flow management, which ensures the cash balance, triggering cash-bank transfers, i.e., bank to cash transfer transactions and cash to bank transfer transactions, when cash balances fall below a threshold; expense processing, which logs different types of expenses, including administrative expenses, sales expenses, and financial expenses; and some other transactions such as interest receivables, which create interest receivable entries for bank deposits.

% i) Initial balances: which include the initializations of cash deposit, bank deposit, and fixed assets. 
% ii) Purchase management: which refers to the purchase-related transactions with different payment methods, e.g., cash and bank transfer. 
% iii) Sales management: which refers to the sales-related transactions, corresponding to the purchase-related transactions. 
% iv) Expense processing: which logs the expenses, including administrative expenses, sales expenses, and financial expenses. 
% v) Fixed assets management: which includes fixed asset purchase transactions, i.e., recording the fixed asset acquisitions, and fixed asset depreciation, i.e., generating monthly depreciation entries on the 1st of each month. 
% vi) Cash flow management: which ensures the cash balance, triggering cash-bank transfers, i.e., bank to cash transfer transactions and cash to bank transfer transactions, if cash balances fall below a threshold. 
% vii) Other transactions: e.g., interest receivables, which create interest receivable entries for bank deposits.
\vspace{-5pt}
\begin{figure}[ht]
    \centering
    \vspace{-10pt}
    \includegraphics[width=\textwidth]{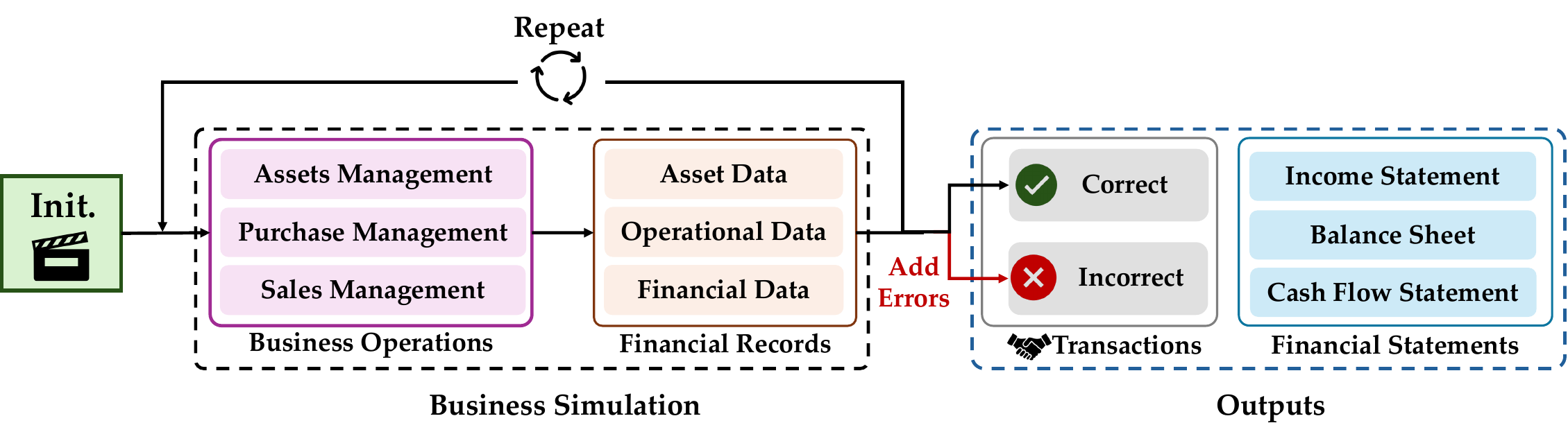}
    \vspace{-15pt}
    \caption{The workflow of \emph{FinSim}.}
    \label{fig:simulator}
    \vspace{-10pt}
\end{figure}

\textbf{Generation Process.}
Figure~\ref{fig:simulator} illustrates the  simulation workflow, where the architecture of \emph{FinSim} follows a multi-stage process to generate transactions and financial statements.
\emph{FinSim} begins with initialization, where the simulator is configured to accurately model a specific type of company.
Then the simulator proceeds to business simulation, where the business operations model financial activities, including assets management, purchase management, and sales management. These business operations further produce financial records, consisting of asset data, operational data, and financial data. This business simulation process will be repeated for the financial outputs.
Transactions are derived from financial records, including both correct transactions and incorrect transactions, where the system adds typical errors to simulate real-world mistakes for auditing.
For the financial statements, the income statement is generated by aggregating revenue and expense transactions from the financial data; the balance sheet combines asset positions from asset data with liability and equity information derived from operational data; and the cash flow statement synthesizes all cash-related transactions, categorizing them into operating, investing, and financing activities.

\subsection{\emph{FinSuite}: Financial Task Suite}

% \begin{figure}
% \centering
% \includegraphics[width=0.5\textwidth]{figures/task.pdf}
% \caption{Task Design}
% \label{fig:task}
% \end{figure}

{\emph{FinSuite} offers a comprehensive suite of 183 financial tasks—64 in financial literacy, 49 in accounting, 35 in auditing, and 35 in consulting. The composition and proportions of tasks are shown in Figure 4(a). The interdependencies among accounting, auditing, and consulting tasks are illustrated in Figure 4(b), highlighting their interconnected nature, which is crucial for evaluating LLM performance. Figure 4(c) provides examples of various tasks to give a clearer understanding of the task formats. }

\textbf{Task Configuration.}
For each task, we construct an innovative three-dimensional metric system $\langle\alpha,\beta,\gamma\rangle$ to precisely characterize task features and complexity. Specifically:
 i) Computational base cardinality $\alpha$: It refers to the number of fundamental data items required to solve the task, capturing the depth of the computational path;
ii)  Cross-source integration level $\beta$: It denotes the number of distinct input data sources involved, reflecting the complexity of data integration;
iii) Output dimensionality breadth $\gamma$: It measures the number of target outputs, indicating the structural complexity of result generation.
This multi-dimensional framework overcomes the limitations of traditional single-metric evaluations, enabling fine-grained error attribution. It also exhibits strong cross-domain adaptability, providing a unified complexity benchmark applicable to accounting, auditing, and consulting tasks, with potential for extension to other domains.

\begin{figure}[htbp]
    \centering
    % First row with two subfigures
    \begin{subfigure}[b]{0.35\textwidth}
        \centering
        \includegraphics[width=\textwidth]{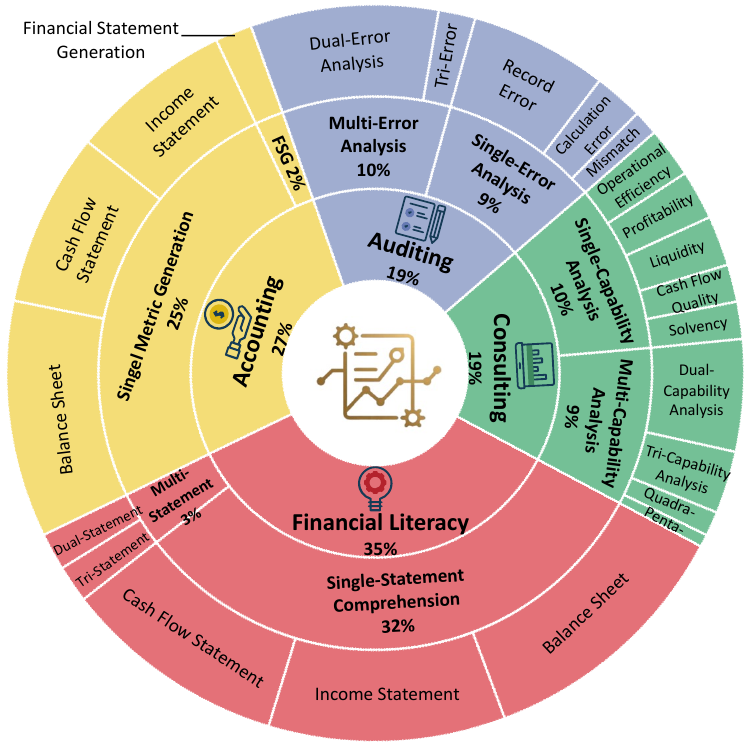}
        \caption{Task Composition and Distribution }
        \label{fig:sub1}
    \end{subfigure}
    \hfill
    \begin{subfigure}[b]{0.625\textwidth}
        \centering
        \includegraphics[width=\textwidth]{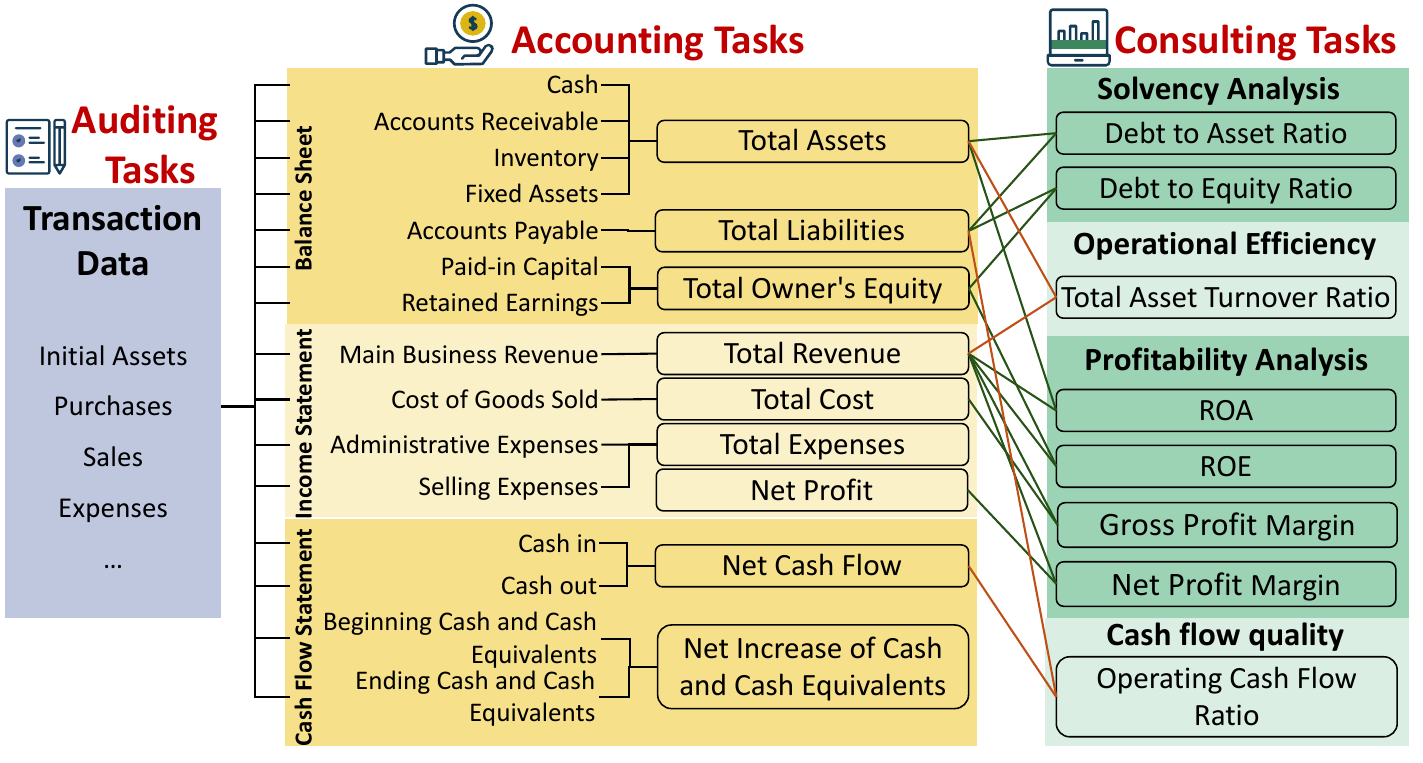}
        \caption{Logical Relationships Among Financial Task Categories}
        \label{fig:sub2}
    \end{subfigure}
        \begin{subfigure}[b]{\textwidth}
        \centering
        \includegraphics[width=\textwidth]{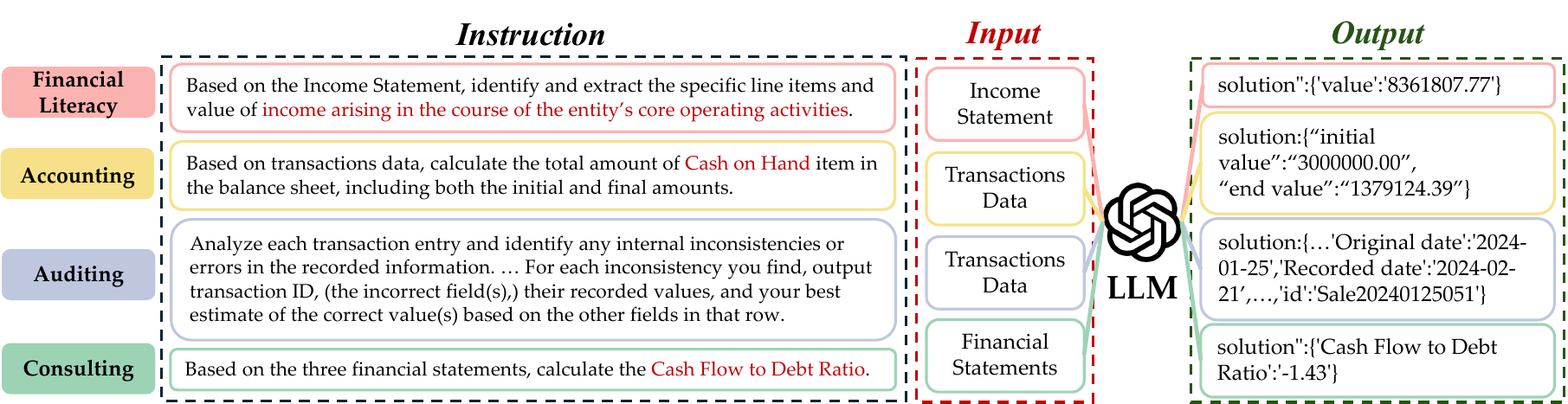}
        \caption{Task Structure Examples Across Four Financial Domains}
        \label{fig:sub3}
    \end{subfigure}
    \caption{Task taxonomy and architecture}
    \label{fig:task}
    \vspace{-10pt}
\end{figure}

\textbf{Financial Literacy Tasks.}
The possession of fundamental financial literacy and an understanding of financial statements forms the foundation for addressing complex tasks in accounting, auditing, and consulting. {Before tackling domain-specific tasks,} \emph{FinSuite} introduces the Financial Literacy task to evaluate whether LLMs possess the requisite competencies to handle sophisticated financial tasks. Leveraging simulator-generated financial reports as input, this task employs a definition-matching query methodology, where models are provided with {concept definitions instead of specific terms and requiring them to accurately locate and extract the corresponding values}. Tasks are categorized based on the number of input statements and output items, progressing from simple to complex configurations, to evaluate LLMs' proficiency in terminology comprehension, logical reasoning, and complex data processing. By structuring the task along a progressive complexity continuum, this evaluation establishes a robust benchmark for downstream expert-level applications while identifying potential deficiencies in the financial knowledge foundations of the models.

\textbf{Accounting Tasks.}
Financial statement generation represents a core accounting function. \emph{FinSuite} conceptualizes this process as the systematic transformation of transaction records into standardized financial statements through a two-tiered framework, as illustrated in Figure \ref{fig:sub1}: first generating disclosure items via elementary computational operations involving single transaction types and complex computational operations requiring cross-transaction analysis, then synthesizing these components into comprehensive financial statements. This paradigm evaluates LLMs' capabilities in generating standardized values according to accounting principles, achieving precision in multisource data integration, and maintaining consistency in applying accounting standards, establishing a benchmark for reliable financial outputs that support audit procedures and financial analysis applications.

% \begin{figure}[ht]
%     \centering
%     \vspace{-10pt}
%     \includegraphics[width=\textwidth]{figures/example.pdf}
%     \vspace{-10pt}
%     \caption{Example}
%     \label{fig:example}
%     \vspace{-10pt}
% \end{figure}

\textbf{Auditing Tasks.}
Based on diverse audit focus areas in practice, \emph{FinSuite} conceptualizes the audit task as a transaction record verification process within the financial audit framework, as shown in the Fugure \ref{fig:sub2}. The experimental paradigm constructs dual core components: generating realistic invoice-format transaction data to simulate authentic business environments, and systematically embedding error samples within these records. Following typical error classifications in audit, twelve basic error types grouped into three categories, were established and embedded into the dataset through randomized generation algorithms as primary evaluation targets. To deepen the measurement dimensions, error types were constructed in two levels: single error analysis and multi-error analysis, as shown in Figure \ref{fig:sub1}. This paradigm evaluates LLMs' performance in identifying audit errors of varying complexity and their semantic understanding of financial textual information.

\textbf{Consulting Tasks.} Analyzing a company's financial performance through financial indicators is a critical service in financial consulting. \emph{FinSuite} conceptualizes the consulting task as an analytical framework based on financial indicators, constructing a diagnostic matrix comprising 18 key indicators in five dimensions: profitability, operational efficiency, liquidity, solvency and cash flow quality. By applying calculations on either single indicators or combinations of multiple indicators, this paradigm systematically evaluates LLMs' accuracy in understanding financial indicator formulas, traceability in data extraction, and computational robustness, as shown in the Figure \ref{fig:sub3}.

\subsection{\emph{FinEval}: LLM Evaluation}
We introduce a unified interface with a designed prompt template to facilitate the evaluations.

\textbf{Prompt Template.}
The prompt template is a structured framework that should be predefined to guide LLMs in solving specific tasks. For \textbf{FinMaster},
we carefully design the prompt template to be simple but general, where no task-specific knowledge is provided, intentionally avoiding task-specific instructions to ensure the consistency for tasks across all modules, i.e., accounting, auditing and consulting. The prompt template is composed of several components. i) Task description: which provides the name and a concise description of the task. ii) Examples: which provides at least one concrete example to explain the input-output schema to guide LLMs to generate the correct solution. iii) Problem to solve: which provides the specific target task to be solved. iv) Instruction: which defines strict JSON output format for easy analysis. The full prompt template is displayed in~\ref{template}.

\textbf{Completion with LLMs.}
To standardize the response processing across diverse LLMs, we develop \emph{FinEval}, which offers a unified interface for the completion of API-based LLMs, enabling direct performance benchmarking and consistent, reproducible generation. Specifically, for the completion with LLMs, \emph{FinEval} includes: i) adaptive prompt construction, which dynamically generates task-specific prompts based on the predefined template, ii) unified LLM execution, which leverages LiteLLM~\citep{litellm} for the response generation of online LLMs, iii) structured answer parsing, which employs regular expressions to extract structured data from JSON-formatted responses.

\section{Results}

\subsection{Analysis of Financial Literacy}
\begin{wrapfigure}{r}{0.4\textwidth}
\centering
\vspace{-0.55cm}
\includegraphics[width=0.4\textwidth]{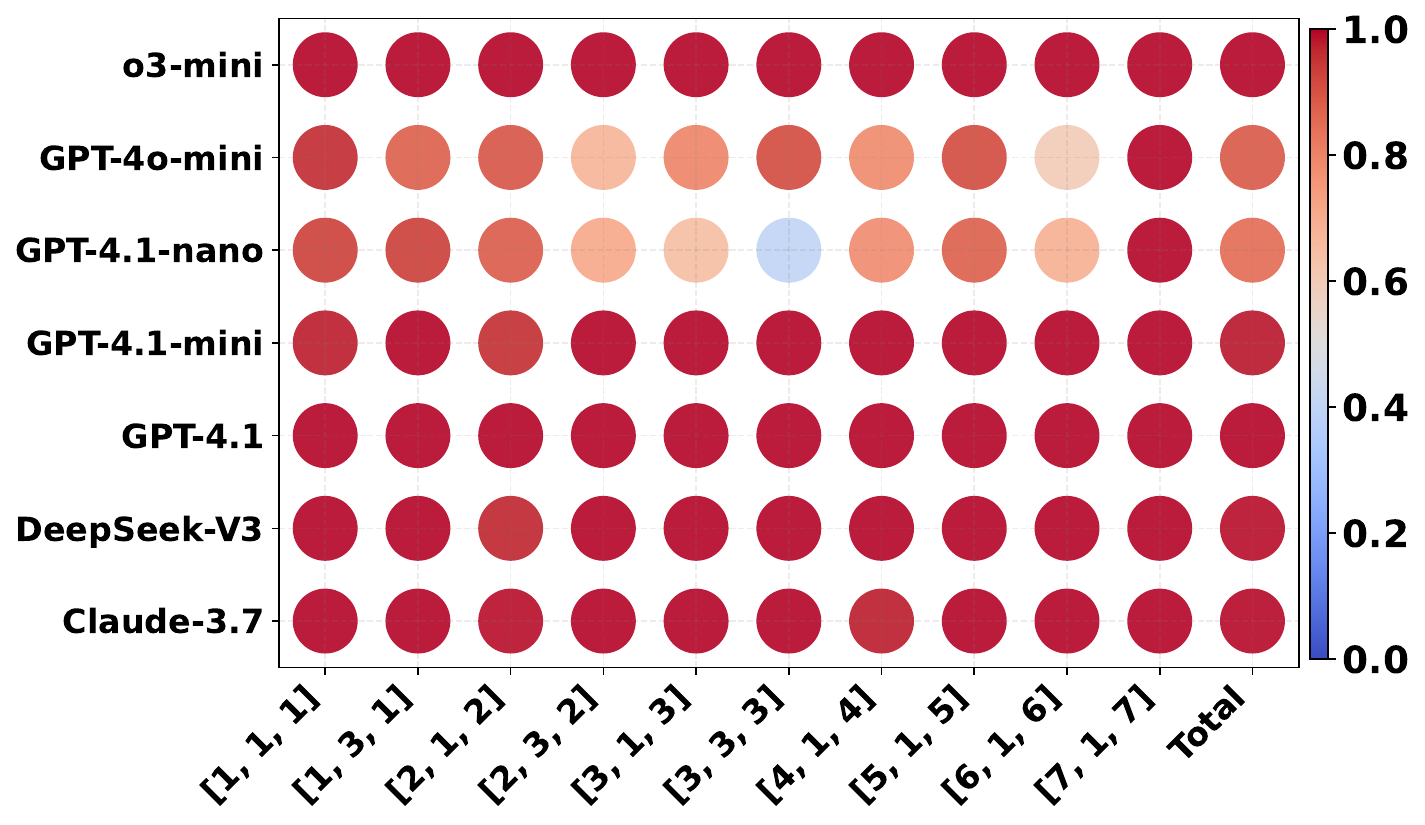}
\caption{Financial literacy result}
\label{fig:accuracy_single}
\vspace{-15pt}
\end{wrapfigure}
\textbf{FinMaster} first assesses the basic financial knowledge and understanding of the financial statements of the selected models. As shown in Figure~\ref{fig:accuracy_single}, LLMs demonstrate a strong performance in basic financial literacy tasks, achieving an average accuracy of 96\%. GPT-4.1, DeepSeek-V3, o3-mini, and Claude-3.7-Sonnet, reach nearly 100\% accuracy on all test questions. However, GPT-4.1-nano and GPT-4o-mini demonstrate lower performance with accuracy between 40-60\%, especially in tasks requiring multiple outputs. These two models show particular difficulty with multistep reasoning and larger-scale inputs.

% \vspace{-10pt}
\begin{tcolorbox}[
    arc=3pt,
    boxrule=0.5pt,
    colback=softblue,
    colframe=black,
    width=\textwidth,
    top=2pt,
    bottom=2pt,
    left=0pt,
    right=5pt
]
\vspace{-2pt}
\textbf{Takeaways}
\begin{itemize}[
    leftmargin=7pt,
    itemsep=0pt,     % 移除项目间的负间距
    topsep=0pt,      % 移除顶部额外间距
    parsep=0pt,      % 保持段落间距为0
    label=\textbullet
]
    \item LLMs demonstrate strong basic financial literacy,  with top models achieving 100\% 
    \item Multi-step tasks and source integration remain challenging for GPT-4.1-nano and GPT-4o-mini 
\end{itemize}
\vspace{-4pt}
\end{tcolorbox}

\subsection{Analysis of Performance}

\begin{figure}[htbp]
    \centering
    \vspace{-10pt}
    \includegraphics[width=1\textwidth]{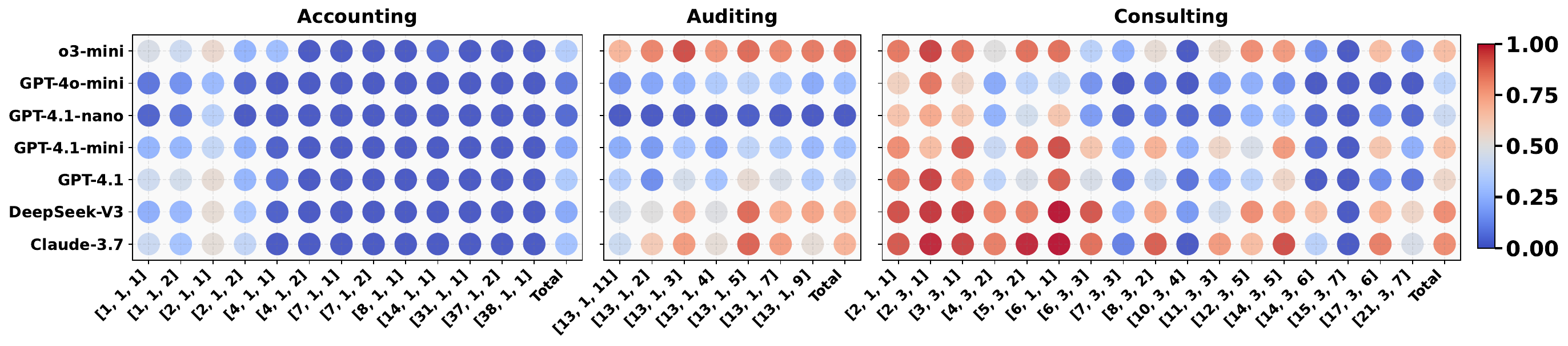}
    \vspace{-15pt}
    \caption{Performance comparison across task types (color bars: model accuracy)}
    \label{fig:accuracy_all}
    \vspace{-10pt}
\end{figure}

The systematic evaluation of LLMs in financial tasks, as shown in Figure~\ref{fig:accuracy_all}, reveals distinct performance patterns. \textbf{FinMaster} demonstrates significant variations across accounting, auditing, and consulting tasks. \textbf{Accounting} emerges as the most challenging type of task, focusing on transforming raw transaction data into financial statements. Models like o3-mini and GPT-4.1 achieve 40-60\% accuracy on basic tasks, demonstrating reasonable proficiency in handling straightforward, isolated calculations. However, their performance significantly deteriorates when faced with complex scenarios: accuracy drops sharply below 20\% for multistep calculations and drop to 3\% on statement generation tasks. This performance reveals the limitations of LLMs in integrating data and comprehending accounting entries, especially when these tasks must be performed simultaneously.

\textbf{Auditing} tasks reveal distinct challenges in error recognition and internal consistency verification. Figure~\ref{fig:accuracy_all} shows that LLMs, especially DeepSeek-V3 and Claude-3.7, achieve relatively higher accuracy on auditing tasks involving multiple simultaneous errors, suggesting that these models effectively leverage patterns and correlations  to detect inconsistencies. However, their accuracy declines when confronted with scenarios containing only a single, isolated error. This may be because isolated anomalies resemble valid entries and lack reinforcing cues, indicating that LLMs rely on detectable error patterns and struggle with sparse or subtle inconsistencies.
\textbf{Consulting} tasks present a unique perspective on model capabilities through cross-statement analysis and complex financial metrics calculation. Analysis reveals that task complexity substantially impacts model performance, with o3-mini showing significant accuracy drops of 35\% in scenarios when analyzing interconnected calculations across financial statements for various financial metrics. DeepSeek-V3 and Claude-3.7-Sonnet demonstrate high but unstable performance ranging from 40-85\%, while GPT-4.1 maintains consistent performance at 55-65\% in most cases. In particular, GPT-4.1-nano achieves good comparable results of 42\% despite its poor performance in other financial tasks, suggesting specialized capabilities in cross-statement analysis and interpretation of financial metrics.

\begin{tcolorbox}[
    arc=3pt,
    boxrule=0.5pt,
    colback=softblue,
    colframe=black,
    width=\textwidth,
    top=2pt,
    bottom=2pt,
    left=0pt,
    right=5pt
]
\vspace{-2pt}
\textbf{Takeaways}
\begin{itemize}[
    leftmargin=7pt,
    itemsep=0pt,     % 移除项目间的负间距
    topsep=0pt,      % 移除顶部额外间距
    parsep=0pt,      % 保持段落间距为0
    label=\textbullet
]
    \item \textbf{FinMaster} effectively reveals LLMs' limitations in realistic financial services
    \item o3-mini and Claude-3.7-Sonnet are the strongest LLMs across all considered financial tasks 
    \item LLMs struggle with multi-step reasoning and multi-source data integration
    \item LLMs' audit capabilities rely on explicit patterns, with weakness in isolated anomaly detection.
\end{itemize}
\vspace{-4pt}
\end{tcolorbox}

\subsection{Analysis of Tokens}
\begin{figure}[htbp]
\vspace{-15pt}
    \centering
    \includegraphics[width=1\textwidth]{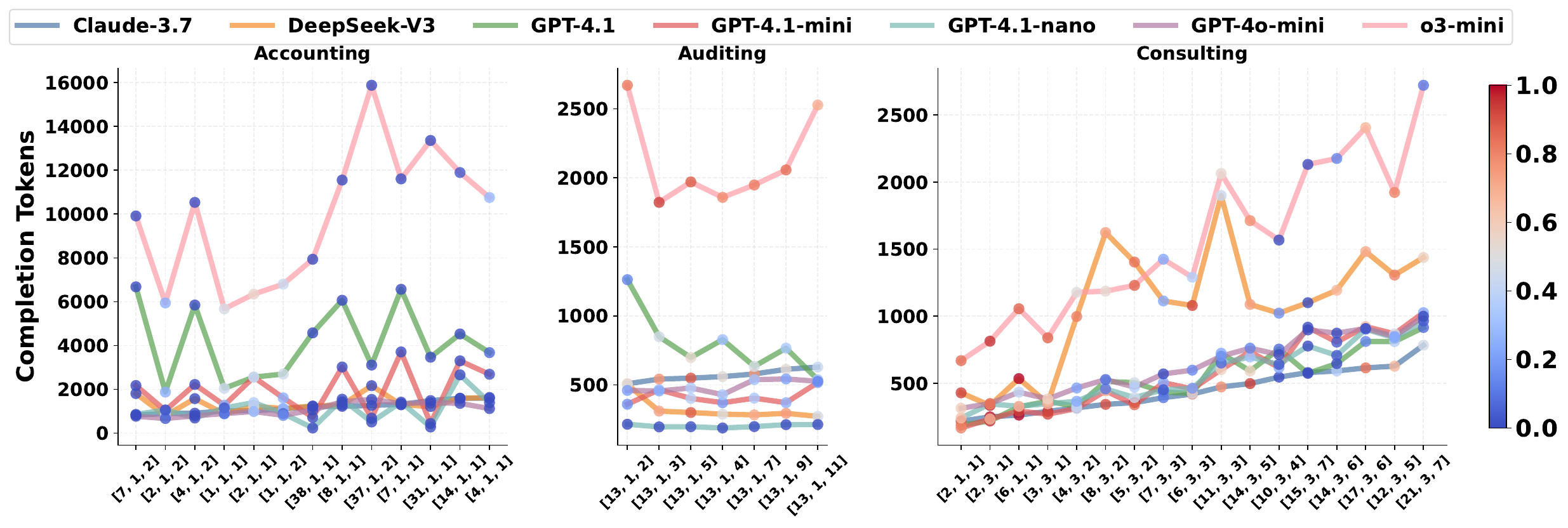}
    \caption{Token usage and performance analysis across task types (color bars: model accuracy)}
    \label{fig:tokens_all}
    \vspace{-8pt}
\end{figure}

Figure~\ref{fig:tokens_all} illustrates token usage and performance patterns in LLMs in financial tasks. Although token consumption varies greatly among different models, it generally tends to increase with the level of task complexity. For example, in accounting tasks, o3-mini uses up to 16,000 tokens in accounting tasks, significantly more than the 2,000 of DeepSeek-V3. While both models demonstrate increasing token usage with rising $\alpha$ values. However, higher token usage does not necessarily translate into improved performance. In auditing tasks, o3-mini reaches 85\% accuracy with lower token usage, surpassing the accuracy of Claude-3.7-Sonnet and DeepSeek-V3 from 70\% to 69\%. This observation highlights a critical insight: model efficiency, reasoning quality, and effective information utilization appear to be more important drivers of performance than sheer token volume. Excessive token consumption may reflect redundant computations, ineffective reasoning processes, or inefficiencies in handling contextual information rather than deeper analytical capabilities.

\begin{tcolorbox}[
    arc=3pt,
    boxrule=0.5pt,
    colback=softblue,
    colframe=black,
    width=\textwidth,
    top=2pt,
    bottom=2pt,
    left=0pt,
    right=5pt
]
\vspace{-2pt}
\textbf{Takeaways}
\begin{itemize}[
    leftmargin=7pt,
    itemsep=0pt,     % 移除项目间的负间距
    topsep=0pt,      % 移除顶部额外间距
    parsep=0pt,      % 保持段落间距为0
    label=\textbullet
]
    \item Token usage grows with task complexity, especially in tasks requiring multi-step reasoning
    \item LLMs performance depends on clear reasoning and efficiency rather than token consumption

\end{itemize}
\vspace{-2pt}
\end{tcolorbox}

\subsection{Ablation}

\begin{figure}[htbp]
    \centering
    \vspace{-10pt}
    \includegraphics[width=1\textwidth]{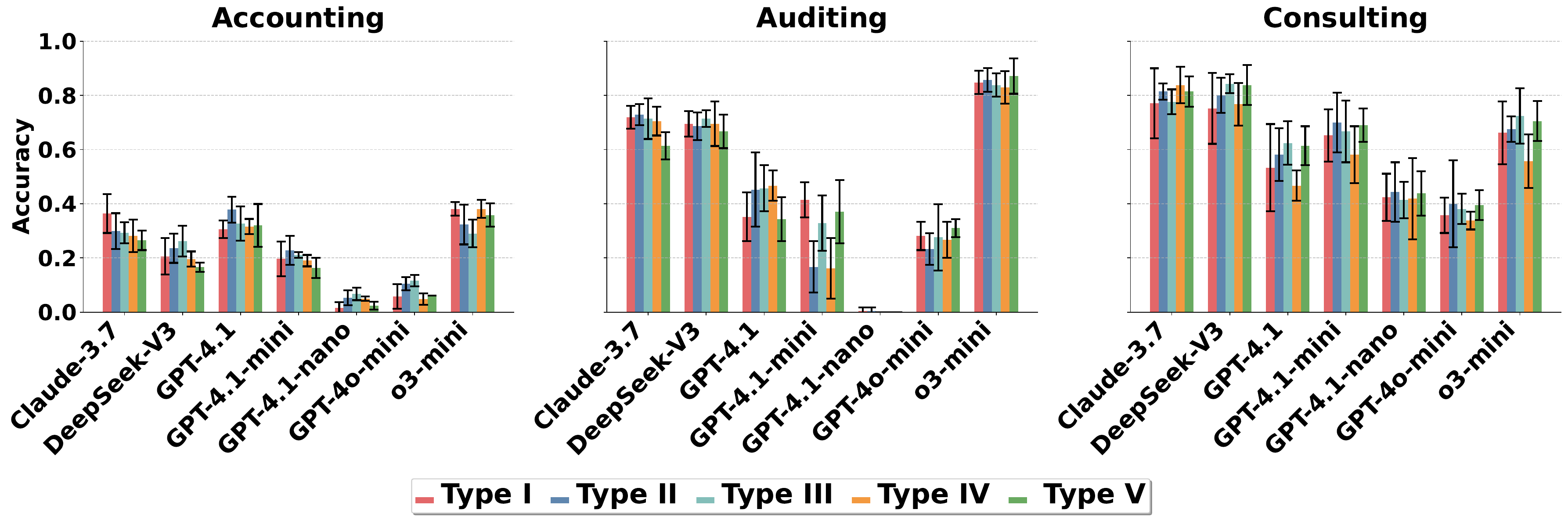}
    \caption{Model performance comparison for different companies (error bars: standard deviation)}
    \label{fig:company_type}
    \vspace{-15pt}
\end{figure}

\paragraph{Different Companies Comparison.}
In the simulation process, \emph{FinSim} designed five different types of companies, each with distinct operational characteristics, to investigate how organizational settings influence model performance across tasks. As shown in Figure~\ref{fig:company_type}, models with higher accuracy such as Claude-3.7-Sonnet and DeepSeek-V3 demonstrate more consistent performance with shorter error bars, while models with lower accuracy such as GPT-4.1-mini, GPT-4.1-nano and o3-mini show greater performance variability with longer error bars. The figure also reveals noticeable differences in accuracy and standard deviation across company types on the same tasks, highlighting that variations in business operations can impact both model effectiveness and stability. Moreover, each model tends to perform better on different company types, and while strong models maintain relatively stable performance across all five settings, weaker models are more affected by organizational differences. These findings indicate that company-specific operations increase model variability, especially in weaker models, highlighting the need to match model choice with real-world business complexity.
These findings indicate that company-specific operations increase model variability, especially in weaker models, highlighting the need to match model choice with real-world business complexity.
\begin{tcolorbox}[
    arc=3pt,
    boxrule=0.5pt,
    colback=softblue,
    colframe=black,
    width=\textwidth,
    top=2pt,
    bottom=2pt,
    left=0pt,
    right=5pt
]
\vspace{-2pt}
\textbf{Takeaways}
\begin{itemize}[
    leftmargin=7pt,
    itemsep=0pt,     % 移除项目间的负间距
    topsep=0pt,      % 移除顶部额外间距
    parsep=0pt,      % 保持段落间距为0
    label=\textbullet
]
    \item \textbf{FinMaster} designed tasks reflect general business scenarios independent of industry specifics

    \item High-accuracy models show consistent results while lower-accuracy models vary more
    \item Company-specific factors increase performance variability, especially for weaker models

\end{itemize}
\vspace{-4pt}
\end{tcolorbox}

\paragraph{Companies Operation Duration Comparison.}

\begin{wrapfigure}{r}{0.5\textwidth}
    \vspace{-15pt}
    \centering
    \includegraphics[width=\linewidth]{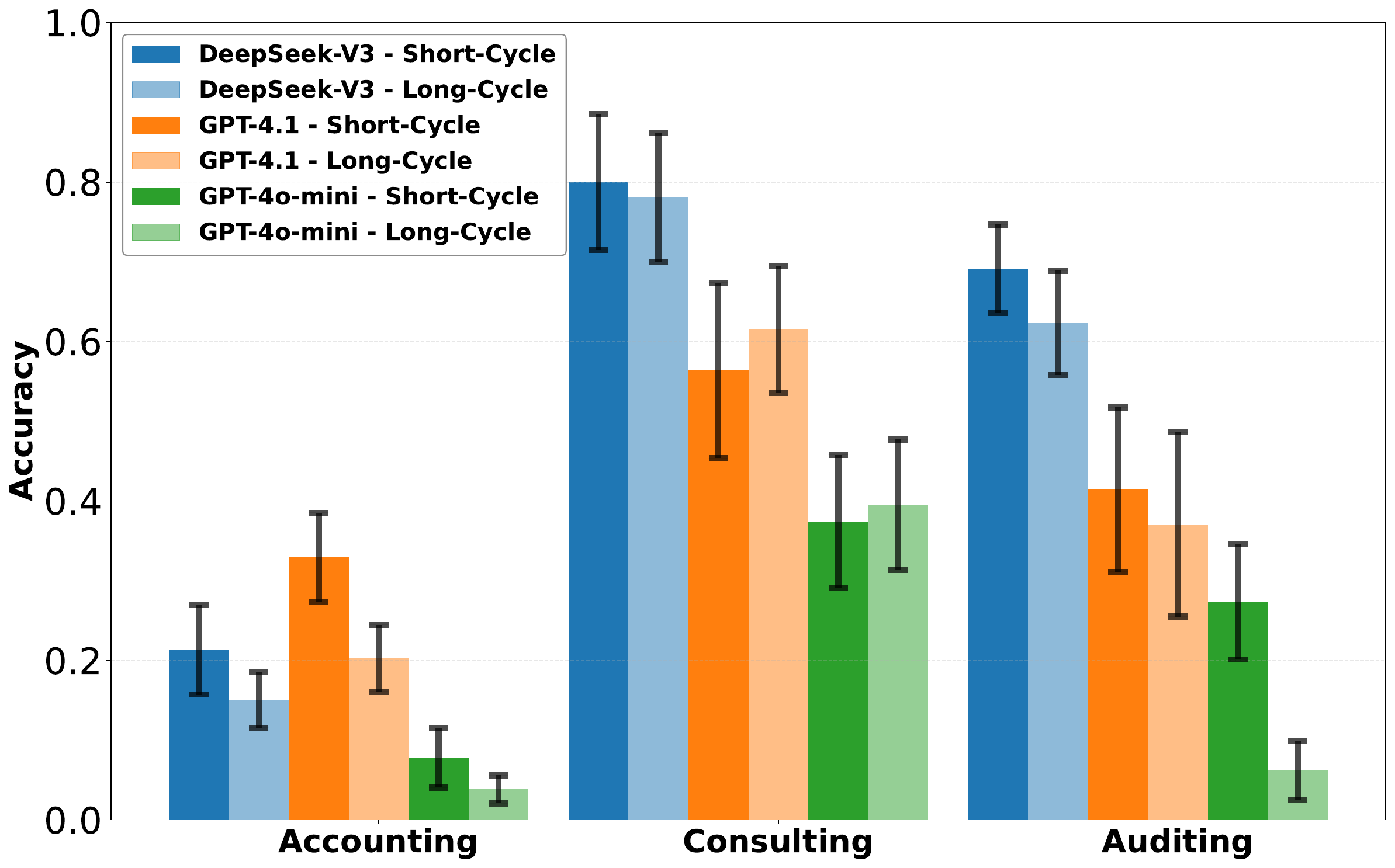}
    \caption{Model accuracy for different operation time (error bars: standard deviation)}
    \label{fig:transaction_compare}
    \vspace{-10pt}
\end{wrapfigure}

In the simulation process, \emph{FinSim} uses transaction volume as a proxy for operational time periods, comparing short-cycle scenarios with 200 transactions versus long-cycle scenarios with 400 transactions per company. As shown in Figure~\ref{fig:transaction_compare}, consulting tasks demonstrate remarkable stability, GPT-4o-mini maintaining the performance of 37\% to 39\% in both cycles and GPT-4.1 showing a slight improvement from 56\% to 61\% in long cycles. In contrast, transaction processing tasks exhibit notable performance degradation: DeepSeek-V3's accuracy in accounting decreases from 21\% to 15\%, while its auditing performance declines from 69\% to 62\% in long cycles. These findings indicate that although models maintain consistent performance in analyzing standardized financial statements, their ability to perform reasoning and calculation declines as operational periods expand. Also, LLMs error bars show no significant difference in variance for different cycles and tasks, suggesting the uncertainty stems from the inherent stochasticity of large language models rather than reasoning input length.

\begin{tcolorbox}[
    arc=3pt,
    boxrule=0.5pt,
    colback=softblue,
    colframe=black,
    width=\textwidth,
    top=2pt,
    bottom=2pt,
    left=0pt,
    right=5pt
]
\vspace{-2pt}
\textbf{Takeaways}
\begin{itemize}[
    leftmargin=7pt,
    itemsep=0pt,     % 移除项目间的负间距
    topsep=0pt,      % 移除顶部额外间距
    parsep=0pt,      % 保持段落间距为0
    label=\textbullet
]
    \item Increasing input transactions weaken LLMs' reasoning and multi-step calculation ability
    \item Model output variance in LLMs stays consistent across cycles, suggesting inherent uncertainty
\end{itemize}
\vspace{-4pt}
\end{tcolorbox}

\subsection{Analysis of Reasoning Failure Cases}

\begin{table}[ht]
\centering
\vspace{-10pt}
\sethlcolor{yellow!25}
\caption{Examples of Failure Cases for o3-mini}
\label{tab:r1_failure_cases}
\begin{tabular}{m{0.9\textwidth}}
\toprule
\scriptsize{\textbf{\colorbox{blue!10}{Failure Reason 1: Domain Knowledge Gap}}} \\
\scriptsize{... All Sale transactions were summed to yield a Main Business Revenue of 6936375.28.  \hl{ Adding positive interest‐earning "Interest Receivables" (net 1735.58) gives Total Revenue of 6938110.86.}  The cost side was computed by summing the "cost\_amount" on all sale rows (totaling 5,700,000.00) which is taken as the Cost of Goods Sold (and Total Cost)...} \\
\midrule
\scriptsize{\textbf{\colorbox{blue!10}{Failure Reason 2: Critical Data Missing}}} \\
\scriptsize{... The sum of all cash outflows (all payments made "by Cash"—for expenses, \hl{ cash purchases, and all "Cash to bank transfer" items) is about 6,361,868.78 and all "Cash to bank transfer" items) is about 6,361,868.78. } Subtracting, the ending balance is  9,333,895.58 – 6,361,868.78 = 2,972,026.80. Thus, in our simulation the cash on hand starts at 3,000,000.00...}\\

\midrule
\scriptsize{\textbf{\colorbox{blue!10}{Failure Reason 3: Floating Point Error}}} \\
\scriptsize{To calculate ROA, we first compute the average total assets and then divide the net profit by this average... Calculate the Return on Assets (ROA): ROA = (Net Profit/Average Total Assets)×100 = (–1,342,040.76/14,043,024.825)×100 \hl{ $\approx$ –9.55\% }. Thus,the ROA is approximately –9.55\%.The solution in the required JSON format is : \{"solution": \{"Return on Assets (ROA)": "-9.55\%"\}...} \\

\midrule
\scriptsize{\textbf{\colorbox{blue!10}{Failure Reason 4: Reasoning Consistency Error}}} \\
\scriptsize{... That is, all cash-using transactions reduced the pool roughly by 14.92 million. Thus the ending cash and cash equivalents is calculated as: \hl{20,000,000.0 - 14,917,367.15 = 5,082,632.85}... Below is our answer in the required JSON format: \hl{\mbox{"solution":{"Ending Balance":"5072632.85"}}}. That is the final ending cash balance. Below is our answer...}\\

\bottomrule
\end{tabular}
\vspace{-5pt}
\end{table}% % 

% \fi

For o3-mini, the reasoning content of the failure cases shows several notable reasons that lead to wrong answers. i) domain knowledge gap: o3-mini, while demonstrating the highest overall accuracy among models, still exhibits a financial knowledge deficiency by incorrectly categorizing Interest Receivables into Main Business Revenue, ultimately resulting in calculation errors; ii) critical data missing: the absence of essential computational data. For example, in the calculation of end value of cash, the omission of "bank to cash transfer" leads to inaccurate results; iii) floating point error: primarily addressing computational deviations caused by data precision errors and numerical rounding processing issues. For instance, when calculating ROA, discrepancies in decimal precision led to a miscalculation, where the correct value of -9.56\% was inaccurately rounded to –9.55\% ; iv) reasoning consistency error: primarily reflected in the inconsistency of the model's logical flow, such as in the case where the initial calculation of cash and cash equivalents was determined to be 5082632.85, the LLMs demonstrated a clear and systematic deviation in output by generating an incorrect value of 5072632.85, which completely differs from the previously calculated numerical value. We list 4 key failure cases for o3-mini in Table \ref{tab:r1_failure_cases} and more examples are shown in Table \ref{tab:model_error} in Appendix ~\ref{app:reasoning_failures}. In additional to explicit model failure, current \textbf{FinMaster} tasks are limited to text-based inputs, lacking the multimodal data processing capabilities essential for realistic financial analysis. Furthermore, general-purpose LLMs face constraints regarding context window size and specialized financial domain knowledge, increasing inaccuracies and hindering comprehensive financial reasoning.

\section{Conclusion}
Existing LLM benchmarks fall short of addressing the complexity, dynamics, and domain-specific nuances of real-world finance. 
Therefore, we introduce \textbf{FinMaster}, a comprehensive benchmark designed to rigorously evaluate LLMs across accounting, auditing, and consulting tasks. Specifically, \textbf{FinMaster} comprises three core modules: i) \emph{FinSim}, a simulator that generates synthetic financial data, including transaction records and financial statements. By simulating real-world market dynamics for various types of companies, \emph{FinSim} overcomes data scarcity and privacy concerns, enabling infinite data generation. ii) \emph{FinSuite}, a comprehensive task suite covering 183 diverse tasks across financial literacy, accounting, auditing, and consulting. It rigorously evaluates the performance and capabilities of LLMs across financial tasks. iii) \emph{FinEval}, a unified evaluation framework with a streamlined interface for systematic model assessment.
Although extensive experiments show that advanced LLMs, including Claude-3.7-Sonnet, DeepSeek-V3, and o3-mini, achieve impressive accuracy of 96\% on fundamental financial literacy tasks, their performance drops sharply to 40\% on complex tasks requiring rigorous multi-source data integration and domain-specific reasoning, indicating that computational error propagation significantly impairs accuracy, indicating computational error propagation worsens performance.
To the best of our knowledge, \textbf{FinMaster} is the first benchmark specifically designed to rigorously evaluate LLMs reasoning capabilities in real-world financial workflows, providing a foundational testbed for advanced LLMs to guide future research directions toward enhancing model robustness, precision, and practical utility in finance.

\bibliography{paper}
\bibliographystyle{plain}
% \section*{References}

% References follow the acknowledgments in the camera-ready paper. Use unnumbered first-level heading for
% the references. Any choice of citation style is acceptable as long as you are
% consistent. It is permissible to reduce the font size to \verb+small+ (9 point)
% when listing the references.
% Note that the Reference section does not count towards the page limit.
% \medskip

% {
% \small

% [1] Alexander, J.A.\ \& Mozer, M.C.\ (1995) Template-based algorithms for
% connectionist rule extraction. In G.\ Tesauro, D.S.\ Touretzky and T.K.\ Leen
% (eds.), {\it Advances in Neural Information Processing Systems 7},
% pp.\ 609--616. Cambridge, MA: MIT Press.

% [2] Bower, J.M.\ \& Beeman, D.\ (1995) {\it The Book of GENESIS: Exploring
%   Realistic Neural Models with the GEneral NEural SImulation System.}  New York:
% TELOS/Springer--Verlag.

% [3] Hasselmo, M.E., Schnell, E.\ \& Barkai, E.\ (1995) Dynamics of learning and
% recall at excitatory recurrent synapses and cholinergic modulation in rat
% hippocampal region CA3. {\it Journal of Neuroscience} {\bf 15}(7):5249-5262.
% }

%%%%%%%%%%%%%%%%%%%%%%%%%%%%%%%%%%%%%%%%%%%%%%%%%%%%%%%%%%%%
% \clearpage

% \section{Technical Appendices and Supplementary Material}
% Technical appendices with additional results, figures, graphs and proofs may be submitted with the paper submission before the full submission deadline (see above), or as a separate PDF in the ZIP file below before the supplementary material deadline. There is no page limit for the technical appendices.

%%%%%%%%%%%%%%%%%%%%%%%%%%%%%%%%%%%%%%%%%%%%%%%%%%%%%%%%%%%%

\newpage

\clearpage
\appendix
\section{Frequently Asked Questions (FAQs)}
\subsection{Why Using Simulated Data is Enough?}

Simulated data addresses several critical challenges in financial LLM research:
\begin{itemize}[leftmargin=*, topsep=0pt, partopsep=0pt, parsep=0pt, itemsep=0pt]
    \item Privacy and Compliance: Real financial data often contains sensitive information, e.g., client records and proprietary transactions, subject to various strict regulations. Simulated transaction records and financial statements generated by \emph{FinSim} 
   can avoid exposing confidential information to eliminate privacy risks while replicating real-world complexity.
    \item Scalability and Diversity: Financial data is inherently sensitive, regulated, and fragmented across institutions, making real-world datasets difficult to obtain and limited in scope.
    \emph{FinSim} can dynamically generate limitless datasets for diverse types of companies and market conditions, enabling robust LLM evaluation without data scarcity.
    \item Evaluation Control: Simulated data makes evaluation controllable. i) Ground truth control: \emph{FinSim} establishes precise, verifiable ground truth for each specific task, which is critical for evaluating the accuracy and robustness of LLMs in financial tasks.
    ii) Complexity control: \emph{FinSim} can systematically modulate task difficulty, which allows benchmarking LLMs' scalability across operational environments.
    iii) Error injection control:
    \emph{FinSim} allows precise manipulation of variables, e.g., for auditing tasks, \emph{FinSim} can simulate transactions with injecting errors, especially for the errors that are impractical to replicate with real data, to evaluate LLMs’ anomaly detection capabilities without exposing real-world misconduct. 
    \item Generalizability: Financial tasks, e.g., accounting, auditing and consulting, rely on standardized formats rather than unpredictable market dynamics, e.g., trading, allowing \emph{FinSim} to replicate realistic data structures, logical relationships, and error patterns that mirror real-world complexity.
    This ensures LLM performance on simulated data transfers reliably to real-world scenarios, with minimal out-of-distribution (OOD) divergence.
\end{itemize}

\subsection{Why Focusing on Accounting, Auditing and Consulting? }
We focus on accounting, auditing and consulting due to three fundamental considerations that collectively establish them as both critical and uniquely positioned for LLM-driven transformation in financial services:
\begin{itemize}[leftmargin=*, topsep=0pt, partopsep=0pt, parsep=0pt, itemsep=0pt]
    \item Critical roles in financial workflows: Accounting, auditing and consulting underpin global financial systems, where accounting ensures accurate record-keeping and compliance, auditing provides the verification mechanism to ensure the reliability and integrity of financial reporting, and consulting translates financial data into actionable business insights and investment decisions. Their temporal dependencies create a comprehensive evaluation framework where LLMs should demonstrate proficiency across data processing, verification, and strategic interpretation.
    \item High automation potential:
    Accounting and auditing involve rule-based but labor-intensive processes, e.g., intercompany reconciliations and anomaly detection in ledgers, ideal for LLM automation to reduce human effort and errors. Consulting leverages LLMs for rapid insights from financial statements or regulatory documents. 
    \item Under-explored complexity in existing LLM benchmarks: Existing benchmarks, e.g., FinQA~\citep{chen2021finqa}, prioritize narrow tasks like numerical question answering from static reports, which only require single-step reasoning and fail to capture the full spectrum of financial reasoning, i.e., from data processing to strategic decision-making. We present \textbf{FinMaster}, which involves tasks requiring multi-step reasoning across the entire financial workflow, to fill this under-explored complexity gaps in existing LLM benchmarks.
\end{itemize}
% emulating temporal dependencies

\subsection{Model Selection}
Due to budget constraints, our evaluation focuses on a curated set of representative models. Specifically, we select online advanced non-reasoning models, e.g., GPT-4o-mini, Claude-3.7-Sonnet, and DeepSeek-V3, and online reasoning model, i.e., o3-mini. For more recent models, we
plan to incorporate them in the next update of our benchmark.

\subsection{Discussion about Limitations and Future Works}
\label{faq:limitation}
\textbf{Multimodal Financial Analysis.}
Current financial tasks in \textbf{FinMaster} are text-based, but real-world financial analysis sometimes involves multimodal data, e.g., charts and scanned documents. Therefore, expanding \emph{FinSim} to generate multimodal financial data and considering multimodal LLMs (MLLMs) can
extend \textbf{FinMaster} to include multimodal financial reasoning, where a multimodal version of \textbf{FinMaster}
would better simulate real-world scenarios.

\textbf{Retrieval Augmented Generation (RAG) for Financial Reasoning.} Financial reasoning often requires access to large-scale datasets, while current LLMs struggle with limited context window size and long-context retention. Therefore, exploring dynamic data retrieval, i.e., integrating RAG to fetch relevant financial data, provides a promising direction for future works to reduce hallucination and improve accuracy.

\textbf{Domain-Specific Training for Financial Expertise.}
General-purpose LLMs lack deep financial knowledge, leading to misinterpretation of financial concepts. \textbf{FinMaster} provides high-quality financial training data through \emph{FinSim}, supporting LLM specialized fine-tuning, which can improve LLM financial reasoning capability and even develop finance domain-specific foundation models.

%没有一步到位会计,不彻底,相对于原本的qa
%受限于 真实的财务报表 窗口大小
%multi-modal

%multi-modal
%database/rag
%训练
%RL 

\subsection{Negative Impacts}
\label{app:negative_impact}
We do not foresee any negative impacts.

\clearpage
\section{Related Work}

% FinQA~\citep{chen2021finqa} is introduced as a large-scale explanatory QA dataset for complex numerical reasoning over financial reports, but it mainly focuses on structured numerical reasoning tasks and relies on static, historical financial reports, which may limit its applicability to dynamic and broader financial scenarios under evolving market conditions. 
% Beyond numerical reasoning, ~\citep{xie2024finben} designs a financial benchmark, FinBen, to evaluate LLMs across diverse financial tasks, e.g., textual analysis, risk management, and market forecasting; however, it lacks granular evaluation for domain-specific reasoning nuances. Other financial benchmarks, including PIXIU~\citep{xie2023pixiu}, FinanceBench~\citep{islam2023financebench}, BizBench~\citep{koncel2023bizbench}, etc, exhibit a narrow scope, primarily emphasizing conventional financial NLP tasks such as question answering for investment advice, sentiment analysis of market news, and basic arithmetic validation in accounting. 

\textbf{Financial Benchmarks.}
We provide a review of existing financial benchmarks for LLMs evaluation.
FinQA~\citep{chen2021finqa} introduces a novel dataset for complex financial QA, requiring models to interpret hybrid data (text/tables) from financial reports, perform multi-step arithmetic operations, and generate program-like reasoning chains to derive answers. While FinQA advances domain-specific QA, it struggles with complex tables and implicit numerical relationships requiring contextual reasoning beyond arithmetic. Furthermore, its narrow focus on structured numerical QA tasks and the dependence on predefined report structures limit its adaptability to evolving real-world financial tasks.
FinBen~\citep{xie2024finben} is a benchmark for financial reasoning that integrates numerical analysis, textual comprehension, and multi-modal data from financial reports, emphasizing tasks like ratio computation, trend prediction, and decision making. However, FinBen overemphasizes on quantitative tasks and underrepresents qualitative reasoning. And its reliance on idealized document formats ignores real-world noises and limits its generalizability.
FinTSB~\citep{hu2025fintsb} focuses on time-series forecasting with high-frequency trading data but overlooks exogenous factors.
Other related financial benchmarks such as FinanceBench~\citep{islam2023financebench}, PIXIU~\citep{xie2023pixiu}, and BizBench~\citep{koncel2023bizbench}, offer limited evaluation task diversity and emphasize NLP capabilities, e.g., information extraction and QA, overlooking complex financial reasoning or practical application scenarios.
SECQUE~\citep{yoash2025secque} is a novel benchmark that advances the evaluation of LLMs in finance by simulating real-world financial challenges. It focuses on practical financial tasks and requires multi-step reasoning to handle noisy data akin to real-world scenarios. However, SECQUE still relies on a static dataset and may not fully capture the dynamic reality of financial workflows.

\textbf{Financial LLMs and Agents.} 
Recent advancements in LLMs have spurred the development of domain-specific LLMs and agents.
FinGPT~\citep{liu2023fingpt} is an open-sourced and data-centric framework, which uses real-time market data and leverages techniques such as Reinforcement Learning with Stock Prices (RLSP) to help models adapt to financial trends.
Concurrently, FinAgent~\citep{zhang2024finagent} introduces a multimodal, agent-based system enhanced with tools, e.g., data retrieval mechanism and chain-of-thought (COT) reasoning, for diverse financial trading tasks.
FinCon~\citep{yu2024fincon} is an LLM-based multi-agent system designed for complex financial decision-making tasks such as stock trading and portfolio management. It employs a hierarchical manager-analyst framework inspired by real-world investment firms, enabling synchronized agent collaboration through natural language.
Baichuan4-Finance~\citep{zhang2024baichuan4} is a specialized LLM optimized for financial applications, which is built upon Baichuan's general AI capabilities and is fine-tuned with extensive financial data to enhance performance in financial tasks, e.g., financial analysis, risk assessment and market prediction.
DianJin-R1~\citep{zhu2025dianjin} is a financial LLM enhanced through Group Relative Policy Optimization (GRPO)~\citep{shao2024deepseekmath}, a reinforcement learning method that incorporates dual reward signals, i.e., format reward and accuracy reward, guiding the model to excel in complex financial reasoning tasks.

\clearpage
\section{Preliminaries of Finance Management Workflows}
\textbf{Accounting}~\citep{godfrey2010accounting} is the systematic practice of recording, summarizing, analyzing, and reporting financial transactions to ensure transparency, compliance, and informed decision-making. It is an essential component of business growth and sustainability. The main purpose of the program is to prepare and disseminate financial reports, speak from the essence, and to fundamentally relate to recording, reporting and resolving financial transactions to support reasonable decisions~\citep{santos2020experimental}. After completing daily business operations, financial activities are recorded and classified as relevant accounts and adjusted to comply with the principles of the meeting to ensure that income and fees are matched over the corresponding meeting period. Finally, the adjusted account balance provides basic data for preparing comprehensive financial reports.

\textbf{Auditing} refers to the systematic and independent examination and evaluation of an organization's financial statements, operational processes, and regulatory compliance conducted by internal or external auditing entities or personnel. In contemporary practice, the scope of review has exceeded the scope of traditional financial reports. Nowadays, people are paying more and more attention to the review and analysis of data generated by the organization's daily operations. This evolution reflects the shift of audits to a more holistic approach, and modern audits not only emphasize the evaluation of financial reporting, but also use advanced data analysis to evaluate daily operational processes and internal controls~\cite{ovami2023data}.

\textbf{Consulting} refers to professional services that help organizations solve problems and achieve objectives through systematic analysis~\citep{biech2019new}. Clients typically seek consulting support to improve business performance or address operational challenges~\citep{bruhn2018impact}. The core of consulting services resides in diagnostic analysis of client operational status, with financial diagnostics emerging as the most strategically critical analytical dimension. This process involves deconstructing financial statements to build a quantitative evaluation framework. Key metrics include profitability (gross/net margins), operational efficiency (inventory/receivables turnover), and solvency (current/quick ratios). Leveraging established analytical paradigms such as DuPont Analysis~\citep{soliman2008use} and Altman Z-score models~\citep{altman2017financial}, this financial diagnostic methodology achieves dual objectives: i) precise identification of resource allocation inefficiencies in corporate operations, and ii) revelation of competitive positioning within industry landscapes through benchmarking analysis against peer comparables. These insights enable data-driven decisions for strategic restructuring, cost optimization, and capital allocation. Financial analytics not only differentiates consultants from domain experts but also validates the feasibility of cross-disciplinary solutions. Thus, financial analysis acts as both a strategic foundation for consulting and a bridge connecting financial data to business realities.

\begin{table}[ht]
\centering
\caption{Balance sheet is a financial status report for a company at a specific time, reflecting all assets, debts and shareholder rights owned by the company}
\renewcommand{\arraystretch}{1.5}
\begin{tabular}{l r r}

\multicolumn{3}{c}{\textbf{BALANCE SHEET}} \\[0.5em]
\hline
\hline
\textbf{Assets} & \textbf{Initial Amount} & \textbf{End Amount} \\ \cline{1-3}
Current Assets & & \\ 
\quad Cash on Hand & \underline{3000000} & \underline{270005.9} \\ 
\quad Bank Deposits & \underline{5000000} & \underline{164645.57} \\ 
\quad Interest Receivable & \underline{0} & \underline{2672.87} \\ 
\quad Accounts Receivable & \underline{0} & \underline{2429482.13} \\ 
\quad Inventory & \underline{0} & \underline{5090000} \\ 
\textbf{Total Current Assets} & \underline{8000000} & \underline{7956806.47} \\ \cline{1-3}
Non-Current Assets & & \\ 
\quad Fixed Assets & \underline{5000000} & \underline{5305354.43} \\ 
\quad Accumulated Depreciation & \underline{0} & \underline{(45751.41)} \\ 
\quad Net Fixed Assets & \underline{5000000} & \underline{5259603.02} \\ 
\textbf{Total Non-Current Assets} & \underline{5000000} & \underline{5259603.02} \\ \cline{1-3}
\textbf{\textit{Total Assets}} & \underline{13000000} & \underline{13216409.49} \\ 
\hline
\hline
\textbf{Liabilities} & & \\ 
\hline
Current Liabilities & & \\ 
\quad Accounts Payable & \underline{0} & \underline{1590000} \\ 
\quad Taxes Payable & \underline{0} & \underline{271550.92} \\ 
\textbf{Total Current Liabilities} & \underline{0} & \underline{1861550.92} \\ \cline{1-3}
\textbf{\textit{Total Liabilities}} & \underline{0} & \underline{1861550.92} \\ 
\hline
\hline
\textbf{Owner’s Equity} & & \\ 
\quad Paid-in Capital & \underline{13000000} & \underline{13000000} \\ 
\quad Retained Earnings & \underline{0} & \underline{-1645141.46} \\ 
\textbf{\textit{Total Owner's Equity}} & \underline{13000000} & \underline{11354859} \\ 
\hline
\hline
\textbf{Total Liabilities and Equity} & \underline{13000000} &\underline{13216409} \\ 
\hline
\hline
\end{tabular}

\end{table}

\begin{table}[ht]
\centering
\caption{Income statement is a financial report that shows a company's revenues, costs, and expenses over a period of time, reflecting the company's profitability and performance}
\renewcommand{\arraystretch}{1.5}
\begin{tabular}{l r}

\multicolumn{2}{c}{\textbf{INCOME STATEMENT}} \\[0.5em]
\hline
\hline
\textbf{Revenue} & \\ 
\cline{1-2}
\quad Main Business Revenue & \underline{5431018.59} \\ 
\textbf{Total Revenue} & \underline{5431018.59} \\ 
\hline
\hline
\textbf{Cost} & \\ 
\cline{1-2}
\quad Cost of Goods Sold & \underline{(4410000)} \\ 
\textbf{Total Cost} & \underline{(4410000)} \\ 
\hline
\hline
\textbf{\textit{Gross Profit}} & \underline{1021018.59} \\ 
\hline
\hline
\textbf{Expense} & \\ 
\quad Administrative Expenses & \underline{(1425164.2)} \\ 
\quad Selling Expenses & \underline{(493854.67)} \\ 
\quad Depreciation & \underline{(45751.41)} \\ 
\quad Financial Expenses & \underline{(432511.69)} \\ 
\textbf{Total Expenses} & \underline{(2397281.97)} \\ 
\hline
\hline
\textbf{Other Revenue} & \\ 
\quad Interest Income & \underline{2672.87} \\ 
\hline
\hline
\textbf{Profit Before Tax}  & \underline{-1373590.51} \\ 
Tax Expense & \underline{271550.92} \\ 
\hline
\hline
\textbf{\textit{Net Profit}} & \underline{-1645141.43} \\ 
\hline
\hline
\end{tabular}
\end{table}

\begin{table}[ht]
\centering
\caption{Cash flow statement is a financial report that shows a company's cash inflows and outflows over a period of time, reflecting how the company generates and uses its cash through operating, investing, and financing activities}
\renewcommand{\arraystretch}{1.5}
\begin{tabular}{l r}

\multicolumn{2}{c}{\textbf{CASH FLOW STATEMENT}} \\[0.5em]
\hline
\hline
\textbf{Cash Flows from Operating Activities} & \\ \cline{1-2}
Net profit &       \underline{-1645141.43} \\ 
Depreciation & \underline{45751.41} \\ 
(Increase) Decrease in Current Assets & \\ 
\quad Accounts Receivable & \underline{(2429482.13)} \\ 
\quad Interest Receivable & \underline{(2672.87)} \\ 
\quad Inventory & \underline{(5090000))} \\ 
Total (Increase) Decrease in Current Assets & \underline{(7522155)} \\ \cline{1-2}
Increase (Decrease) in Current Liabilities &\\ 
\quad Accounts Payable & \underline{1590000} \\ 
\quad Tax Payable & \underline{271550.92} \\ 
Total Increase (Decrease) in Current Liabilities & \underline{1861550.92} \\ 
\hline
\hline
\textbf{\textit{Net Cash Flow From Operations}} & \underline{-7259994.1} \\ \cline{2-2}
\hline
\hline
\textbf{Cash Flows from Investing Activities} &  \\ \cline{1-2}
Purchase of Fixed Assets & \underline{305354.43} \\ \cline{2-2}
\hline
\hline
\textbf{\textit{Net Cash Flows from Investing Activities}} & \underline{(305354.43)} \\ \cline{1-2}
\hline
\hline
\cline{1-2}
\textbf{Beginning Cash and Cash Equivalents Balance} & \underline{8000000} \\ \cline{2-2}
\textbf{Ending Cash and Cash Equivalents Balance} & \underline{434651.47} \\ \cline{2-2}
\textbf{Net Increase} & \underline{(7565348.53)} \\ 
\hline
\hline
\end{tabular}
\end{table}

\clearpage
\section{FinSim}

\subsection{Types of Companies}
\label{app:company}
The configurations of different types of companies are displayed in Table~\ref{tab:company}. Specifically, for initial capital, we use the sum of the initial bank deposit, the initial fixed assets, and the purchase unit price to represent; for features including profit margin and different frequencies, we display both minimum and maximum values.

\begin{table}[ht]
    \centering
    \caption{Types of Companies.}
    \resizebox{0.9\textwidth}{!}{
    \begin{tabular}{c|ccccc}
    \toprule
        &  Type \uppercase\expandafter{\romannumeral 1} & Type \uppercase\expandafter{\romannumeral 2} & Type \uppercase\expandafter{\romannumeral 3} & Type \uppercase\expandafter{\romannumeral 4} & Type \uppercase\expandafter{\romannumeral 5} \\
         
    \midrule
    Initial Capital & 28M & 13M & 13M & 13M & 16M\\
    Fixed Asset Purchase Frequency & 
    0.00/2.00 & 1.00/2.00 &  1.00/2.00 & 0.00/1.00 & 0.00/2.00\\
    Purchase Unit Price & 950,000 & 45,000 & 21,250 & 31,500 & 1,823\\
    Profit Margin & 0.30/0.50 & 0.10/0.40 & 0.70/1.00 & 0.80/2.00 & 0.30/0.80\\
    Quantity Per Purchase & 1.00 & 15.00 & 5.00 & 2.00 & 500.00\\
    Purchase Frequency & 1.00/2.00 & 1.00/3.00 & 2.00/4.00 & 0.00/2.00 & 1.00/3.00\\
    Credit Purchase Ratio & 0.1 & 0.1 & 0.3 & 0.3 & 0.6\\
    Quantity Per Sale  & 1.00 & 5.00 & 3.00 & 1.00 & 5.00\\
    Sales Frequency  & 0.00/1.00 & 1.00/2.00 & 2.00/4.00 & 0.00/3.00 & 2.00/4.00\\
    Credit Sales Ratio & 0.6 & 0.4 & 0.3 & 0.7 & 0.4\\
    Expense Frequency & 1.00/2.00 & 2.00/4.00 & 2.00/3.00 & 1.00/2.00 & 1.00/2.00\\
    \bottomrule
    \end{tabular}}
    \label{tab:company}
\end{table}

\begin{itemize}[leftmargin=*, topsep=0pt, partopsep=0pt, parsep=0pt, itemsep=0pt]
    \item Type \uppercase\expandafter{\romannumeral 1} considers capital goods manufacturers, e.g., heavy machinery and shipbuilding companies, which are characterized by capital-intensive operations
    with low sales frequency but premium purchase unit prices, reflecting specialized, high-value products. 
    
    % Their large accounts receivable and concentrated client base indicate reliance on long-term contracts with limited buyers. 
    \item Type \uppercase\expandafter{\romannumeral 2} considers transaction-driven companies, e.g., Chemical trader and industrial product distributor.
    These companies often face stable procurement costs but lack pricing power, leading to low, volatile gross margins. To maintain sales and market share, they rely on bulk purchasing and large-scale sales, despite high selling and administrative costs.
    
    % where sales occur at a stable frequency with varied pricing across products. The have large fixed assets and large inventory to adapt to market price shifts, which directly impacts profit margins. 
    
% A key characteristic of these companies is the relative stability of procurement costs, combined with a lack of pricing power, which results in low and volatile gross margins. To sustain sales volume and maintain market share, they typically rely on bulk purchasing and large-scale sales, while incurring substantial selling and administrative expenses.

    \item Type \uppercase\expandafter{\romannumeral 3} considers companies that offer high value-added consumer goods, such as luxury brands or premium electronics manufacturers. 
    These businesses are characterized by high gross margins and low production costs. To sustain high revenue levels, they often make significant investments in selling expenses, particularly in branding and marketing efforts.

    \item Type \uppercase\expandafter{\romannumeral 4} considers asset-light companies, e.g., consulting and designing companies, which operate light-asset models with minimal fixed assets, high profit margin, and even no inventory. These businesses typically have a high purchase-on-credit rate, relying on credit for procurement, while maintaining robust profitability due to their low capital requirements and high-margin service models.

    \item Type \uppercase\expandafter{\romannumeral 5} considers high-turnover companies, e.g., hotel and catering enterprises, which are characterized by high sales frequency, low unit prices, large quantity per purchase, and a dispersed customer base.
\end{itemize}

\clearpage

\section{FinSuite}

In this section, we present the complete information of the tasks for financial literacy, accounting, auditing and consulting considered in \textbf{FinMaster}, detailing each task's name, difficulty, description, and input and output specifications.
\subsection{Financial Statement Items Definition}

\begin{table}[ht]
\raggedright  
\caption{Definition of Balance Sheet Items}
\resizebox{\textwidth}{!}{
\begin{tabular}{@{}p{6cm}p{12cm}@{}} 
\toprule
\midrule
\textbf{Item Name } & \textbf{Item Definition}  \\
\midrule
\midrule
Cash on Hand & Cash held by an entity that is available for use in its day-to-day operations. \\
\midrule
Bank Deposits & Bank deposits are funds deposited into a bank or other financial institution. \\
\midrule
Interest Receivable & Amounts of interest accrued but not yet received. \\
\midrule
Accounts Receivable & Amounts owed to the entity for goods or services sold or provided on credit. \\
\midrule
Inventory & Assets held for sale in the ordinary course of business, in production for such sale, or in the process of being manufactured. \\
\midrule
Total Current Assets & The total amount of assets that are expected to be realised or intended for sale or consumption in the normal course of the entity's operating cycle. \\
\midrule
Fixed Assets & Tangible items that are held for use in the production or supply of goods or services, for rental to others, or for administrative purposes. \\
\midrule
Accumulated Depreciation & The total amount of depreciation recognised as an expense in the statement of profit or loss and other comprehensive income. \\
\midrule
Total Non-current Assets & The total amount of assets that are not expected to be realised or intended for sale or consumption in the normal course of the entity's operating cycle. \\
\midrule
Total Assets & The total present economic resources controlled by the entity as a result of past events, which also means the sum of all assets owned by an entity, both current and non-current, that are expected to bring future economic benefits to the company. \\
\midrule
Accounts Payable & Amounts owed by the entity for goods or services received or purchased on credit. \\
\midrule
Taxes Payable & Amounts of taxes accrued but not yet paid. \\
\midrule
Total Current Liabilities & The total amount of liabilities that are expected to be settled in the normal course of the entity's operating cycle. \\
\midrule
Paid-in Capital & The amount of capital contributed by shareholders in exchange for shares. \\
\midrule
Retained Earnings & The amount of profit or loss retained in the entity, rather than being distributed to shareholders. \\
\midrule
Total Owner's Equity & The total amount of equity recognised in the statement of financial position. \\
\midrule
Total Liabilities and Owner's Equity & The total amount of liabilities and equity recognised in the statement of financial position. \\
\bottomrule
\bottomrule
\end{tabular}}
\end{table}

\begin{table}[ht]
\raggedright  
\caption{Definition of Income Statement Items}
\resizebox{\textwidth}{!}{
\begin{tabular}{@{}p{6cm}p{12cm}@{}} 
\toprule
\midrule
\textbf{Item Name } & \textbf{Item Definition}  \\
\midrule
\midrule
Main Business Revenue & Income arising in the course of the entity's core operating activities. \\
\midrule
Total Revenue & Total income arising in the course of an entity’s ordinary activities. \\
\midrule
Cost of Goods Sold & Carrying amount of inventories sold during the reporting period. \\
\midrule
Total Cost & The aggregate of all expenses incurred by a company to generate its revenues during a specific accounting period. \\
\midrule
Gross Profit & Gross profit is the difference between sales revenue and the cost of goods sold. Gross profit is the cleanest accounting measure of true economic profitability. \\
\midrule
Administrative Expenses & The costs of distribution or administrative activities; costs of general management and administration of the entity as a whole. \\
\midrule
Selling Expenses & Costs incurred to secure customer orders and to deliver the goods and services to customers. \\
\midrule
Depreciation & The systematic allocation of the depreciable amount of an asset over its useful life. \\
\midrule
Financial Expenses & Financing costs incurred by an enterprise to raise funds needed for production and operation. \\
\midrule
Total Expenses & The total amount of expenses incurred by an entity during a reporting period. \\
\midrule
Interest Income & Income earned by an entity from financial assets. \\
\midrule
Profit Before Tax & Profit or loss for a period before deducting tax expense. It represents the company's earnings from all activities—operating and non-operating—before the effects of tax expenses. \\
\midrule
Tax Expense & Total amount of taxes an entity is expected to pay or recover during a reporting period. \\
\midrule
Net Profit & The amount of profit an entity retains after all expenses, including operating costs, interest, taxes, depreciation, and amortization, have been deducted from total revenue. \\
\bottomrule
\bottomrule
\end{tabular}}
\end{table}

\begin{table}[ht]
\raggedright  
\caption{Definition of Cash Flow Statement Items}
\resizebox{\textwidth}{!}{
\begin{tabular}{@{}p{6cm}p{12cm}@{}} 
\toprule
\midrule
\textbf{Item Name } & \textbf{Item Definition}  \\
\midrule
\midrule
Cash Flow From Operating Activities & The cash inflows and outflows generated by a company's core business operations during a specific period. \\
\midrule
Net Profit & The amount of profit an entity retains after all expenses, including operating costs, interest, taxes, depreciation, and amortization, have been deducted from total revenue. \\
\midrule
Depreciation & The systematic allocation of the depreciable amount of an asset over its useful life. \\
\midrule
(Increase) Decrease in Accounts Receivable & The reduction in the amounts owed to the entity for goods or services sold or provided on credit during the period. \\
\midrule
(Increase) Decrease in Interest Receivable & The reduction in the amount of interest accrued but not yet received during the period. \\
\midrule
(Increase) Decrease in Inventory & The reduction in the amount of assets held for sale in the ordinary course of business, in production for such sale, or in the process of being manufactured during the period. \\
\midrule
Increase (Decrease) in Accounts Payable & The addition in the amount owed by the entity for goods or services received or purchased on credit during the period. \\
\midrule
Increase (Decrease) in Tax Payable & The addition in the amount of taxes accrued but not yet paid during the period. \\
\midrule
Net Cash Flow from Operating Activities & The total cash generated or used by a company's core business operations after accounting for all cash inflows and outflows within a specific period. \\
\midrule
Cash Flow from Investing Activities & Investing activities are the acquisition and disposal of long-term assets and other investments not included in cash equivalents and the receipt of interest and dividends. \\
\midrule
Purchase of Fixed Assets & The acquisition of property, plant and equipment. \\
\midrule
Net Cash Flow from Investing Activities & The net amount of cash and cash equivalents generated from an entity's activities that are the acquisition and disposal of long-term assets and other investments not included in cash equivalents and the receipt of interest and dividends. \\
\midrule
Beginning Balance & The amount of cash and cash equivalents at the beginning of the period. \\
\midrule
Ending Balance & The amount of cash and cash equivalents at the end of the period. \\
\midrule
Net Increase & The net addition in the amount of cash and cash equivalents during the period. \\
\bottomrule
\bottomrule
\end{tabular}}
\end{table}

% \clearpage

% \subsection{Critical Financial Indicators Display}

\clearpage

\subsection{Element Categorization}

\begin{table}[ht]
    \centering
    \caption{Audit Basic Error Classification}
    \begin{tabular}{c|c}
        \toprule
         Task Name & Error Category \\
        \midrule
         Transaction TYPE Record Error & Record Error \\ 
         Transaction DATE Record Error & Record Error\\
         Transaction PAYMENT/RECEIPT\_STATUS Record Error & Record Error \\
         Transaction PAYMENT\_METHOD Record Error & Record Error\\
         Transaction QUANTITY Record Error & Record Error\\
         Transaction UNIT\_PRICE Record Error & Record Error\\
         Transaction RECEIVE\_METHOD Record Error & Record Error\\
         Transaction AMOUNT Calculation Error & Calculation Error\\
         Transaction TAX\_AMOUNT Calculation Error & Calculation Error\\
         Transaction PROFIT Calculation Error & Calculation Error\\
         Transaction Without PREPARER Error & Transaction Approval Mismatch\\
         Transaction Without APPROVER Error& Transaction Approval Mismatch\\
        \bottomrule
    \end{tabular}
\end{table}

\begin{table}[ht]
    \centering
    \caption{Financial Indicators Classification}
    \begin{tabular}{c|c}
        \toprule
         Indicator &  Category \\
        \midrule
         Free Cash Flow (FCF) & Cash Flow Quality \\ 
         Operating Cash Flow to Net Income Ratio & Cash Flow Quality\\
         Operating Cash Flow Ratio & Cash Flow Quality \\
         Gross Profit Margin & Profitability\\
         Net Profit Margin & Profitability\\
         Return on Assets (ROA) & Profitability\\
         Return on Equity (ROE) & Profitability\\
         Current Ratio & Liquidity\\
         Quick Ratio & Liquidity\\
         Cash to Current Debt Ratio & Liquidity\\
         Operating Cash Flow to Current Liabilities Ratio & Liquidity\\
         Debt to Asset Ratio & Solvency\\
         Debt to Equity Ratio & Solvency\\
         Cash Flow to Debt Ratio & Solvency\\
         Inventory Turnover Ratio & Operational Efficiency\\
         Accounts Receivable Turnover Ratio & Operational Efficiency\\
         Current Assets Turnover Ratio & Operational Efficiency\\
         Total Asset Turnover Ratio & Operational Efficiency\\
        \bottomrule
    \end{tabular}
\end{table}

\clearpage

\subsection{Critical Financial Indicators Display}

\begin{table}[ht]
\centering
\caption{Critical Financial Indicators Description and Formula}
\resizebox{\textwidth}{!}{
\begin{tabular}{>{\centering\arraybackslash}m{3.5cm}||>{\centering\arraybackslash}m{6cm}|c}
\toprule
\midrule
\textbf{Indicator} & \textbf{Description} & \textbf{Formula}  \\
\midrule
\midrule
Free Cash Flow (FCF) & The cash remaining after a company has paid for its operating expenses and capital expenditures. & $
\text{Net Cash Flow from Operating Activities} - \text{Purchase of Fixed Assets}
$\\
\midrule
Operating Cash Flow to Net Income Ratio & A ratio that evaluates the relationship between cash generated from operating activities and net income. & $
\text{Net Cash Flow from Operating Activities}/{\text{Net Profit}}
$\\
\midrule
Operating Cash Flow Ratio & A liquidity metric that measures the adequacy of operating cash flow in covering a company's short-term liabilities. & $
\text{Net Cash Flow from Operating Activities}/{\text{Current Liabilities}}
$\\
\midrule
Gross Profit Margin & A profitability ratio calculated as gross profit divided by revenue, expressed as a percentage. & $
((\text{Revenue} - \text{COGS})/{\text{Revenue}}) \times 100\%
$\\
\midrule
Net Profit Margin & A financial metric that shows the percentage of net income derived from total revenue. & $
(\text{Net Profit}/\text{Revenue})\times 100\%
$ \\
\midrule
Return on Assets (ROA) & A profitability ratio that measures the efficiency with which a company utilizes its total assets to generate net income.  & $
((2 * \text{Net Profit})/(\text{Beginning Total Assets} + \text{Ending Total Assets})) \times 100\%
$\\
\midrule
Return on Equity (ROE) & A performance metric that quantifies the return generated on shareholders' equity.  & $
((2 * \text{Net Profit})/(\text{Beginning Owner's Equity} + \text{Ending Owner's Equity})) \times 100\%
$ \\
\midrule
Current Ratio & A liquidity ratio calculated as current assets divided by current liabilities. & $
\text{Current Assets}/\text{Current Liabilities}
$ \\
\midrule
Quick Ratio & A stringent liquidity measure that assesses a company's ability to pay off its current liabilities using its most liquid assets. & $
(\text{Current Assets} - \text{Inventory})/\text{Current Liabilities}
$ \\
\midrule
Cash to Current Debt Ratio & A liquidity ratio that evaluates the proportion of cash and cash equivalents available to settle current liabilities, indicating short-term financial stability. & $
(\text{Cash and Cash Equivalents – Ending Balance})/\text{Current Liabilities}
$  \\
\midrule
Operating Cash Flow to Current Liabilities Ratio & A ratio that measures the sufficiency of cash generated from operating activities to cover current liabilities, reflecting operational efficiency and liquidity. & $
\text{Net Cash Flow from Operating Activities}/\text{Ending Current Liabilities}
$\\
\midrule
Debt to Asset Ratio & A leverage ratio that calculates the percentage of a company's total assets financed through debt. It is determined by dividing total debt by total assets. & $
\text{Total Liabilities}/\text{Total Assets}
$ \\
\midrule
Debt to Equity Ratio & A financial leverage metric that compares a company's total debt to its shareholders' equity, illustrating the proportion of debt used relative to equity financing. & $
\text{Total Liabilities}/\text{Owner's Equity}
$\\
\midrule
Cash Flow to Debt Ratio & A solvency ratio that measures a company's ability to repay its total debt using cash generated from operating activities. & $\text{Net Cash Flows from Operating Activities}/\text{Total Liabilities}$\\
\midrule
Inventory Turnover Ratio & An efficiency metric that calculates how many times a company sells and replaces its inventory over a specific period. & $
(2 * \text{COGS})/(\text{Beginning Inventory} + \text{Ending Inventory})
$ \\
\midrule
Accounts Receivable Turnover Ratio & A ratio that measures the efficiency of a company in collecting its accounts receivable. & $
(2 * \text{Revenue})/(\text{Beginning Accounts Receivable} + \text{Ending Accounts Receivable)}
$  \\
\midrule
Current Assets Turnover Ratio & An efficiency ratio that evaluates how effectively a company utilizes its current assets to generate revenue. & $
(2 * \text{Revenue})/(\text{Beginning Current Assets} + \text{Ending Current Assets})
$ \\
\midrule
Total Asset Turnover Ratio & A financial efficiency metric that measures the ability of a company to generate revenue from its total assets. & $
(2 * \text{Revenue})/(\text{Beginning Total Assets} + \text{Ending Total Assets})
$  \\

\bottomrule
\bottomrule
\end{tabular}
}
\end{table}

\clearpage
\subsection{Financial Statement Tasks Information}

\begin{table}[htbp]
\raggedright  
\caption{Task Information Table - Balance Sheet Detection in Financial Literacy}
\resizebox{\textwidth}{!}{
\begin{tabular}{@{}p{4cm}p{1.5cm}p{10cm}p{2.0cm}p{5cm}@{}} 
\toprule
\midrule
\textbf{Task Name} & \textbf{\{$\alpha$,$\beta$,$\gamma$\}} & \textbf{Task Description} & \textbf{Input} & \textbf{Output} \\
\midrule
\midrule
Financial Literacy Detection-Cash on Hand & \{2,1,2\} & Based on the balance sheet, identify and extract the specific line items and value of cash held by an entity that is available for use in its day-to-day operations, including initial and final value & Balance sheet & The value of cash on hand, including initial and final value \\
\midrule
Financial Literacy Detection-Bank Deposits & \{2,1,2\} & Based on the balance sheet, identify and extract the specific line items and value of funds deposit into a bank, including initial and final value & Balance sheet & The value of bank deposits, including initial and final value \\
\midrule
Financial Literacy Detection-Accounts Receivable & \{2,1,2\} & Based on the balance sheet, identify and extract the specific line items and value of amounts owed to the entity for goods or services sold or provided on credit, including initial and final value & Balance sheet & The value of accounts receivable, including initial and final value \\
\midrule
Financial Literacy Detection-Interest Receivable & \{2,1,2\} & Based on the balance sheet, identify and extract the specific line items and value of amounts of interest accrued but not yet received, including initial and final value & Balance sheet & The value of interest receivable, including initial and final value \\
\midrule
Financial Literacy Detection-Inventory & \{2,1,2\} & Based on the balance sheet, identify and extract the specific line items and value of assets held for sale in the ordinary course of business, in production for such sale, or in the process of being manufactured, including initial and final value & Balance sheet & The value of inventory, including initial and final value \\
\midrule
Financial Literacy Detection-Fixed Assets & \{2,1,2\} & Based on the balance sheet, identify and extract the specific line items and value of tangible items that are held for use in the production or supply of goods or services, for rental to others, or for administrative purposes, including initial and final value & Balance sheet & The value of fixed assets, including initial and final value \\
\midrule
Financial Literacy Detection-Accumulated Depreciation & \{2,1,2\} & Based on the balance sheet, identify and extract the specific line items and value of expense in the statement of profit or loss and other comprehensive income, including initial and final value & Balance sheet & The value of accumulated depreciation, including initial and final value \\
\midrule
Financial Literacy Detection-Accounts Payable & \{2,1,2\} & Based on the balance sheet, identify and extract the specific line items and value of the amounts owed by the entity for goods or services received or purchased on credit, including initial and final value & Balance sheet & The value of accounts payable, including initial and final value \\
\midrule
Financial Literacy Detection-Taxes Payable & \{2,1,2\} & Based on the balance sheet, identify and extract the specific line items and value of taxes accrued but not yet paid, including initial and final value & Balance sheet & The value of taxes payable, including initial and final value \\
\midrule
Financial Literacy Detection-Paid-in Capital & \{2,1,2\} & Based on the balance sheet, identify and extract the specific line items and value of capital contributed by shareholders in exchange for shares, including initial and final value & Balance sheet & The value of paid-in capital, including initial and final value \\
\midrule
Financial Literacy Detection-Retained Earnings & \{2,1,2\} & Based on the balance sheet, identify and extract the specific line items and value of profit or loss retained in the entity, rather than being distributed to shareholders, including initial and final value & Balance sheet & The value of retained earnings, including initial and final value \\
\midrule
Financial Literacy Detection-Current Assets & \{2,1,2\} & Based on the balance sheet, identify and extract the specific line items and value of total assets that are expected to be realised or intended for sale or consumption in the normal course of the entity's operating cycle, including initial and final value & Balance sheet & The value of current assets, including initial and final value \\
\midrule
Financial Literacy Detection-Non-current Assets & \{2,1,2\} & Based on the balance sheet, identify and extract the specific line items and value of total assets that are not expected to be realised or intended for sale or consumption in the normal course of the entity's operating cycle, including initial and final value & Balance sheet & The value of non-current assets, including initial and final value \\
\midrule
Financial Literacy Detection-Current Liabilities & \{2,1,2\} & Based on the balance sheet, identify and extract the specific line items and value of total amount of liabilities that are expected to be settled in the normal course of the entity's operating cycle, including initial and final value & Balance sheet & The value of current liabilities, including initial and final value \\
\midrule
Financial Literacy Detection-Owner's Equity & \{2,1,2\} & Based on the balance sheet, identify and extract the specific line items and value of total amount of equity recognised in the statement of financial position, including initial and final value & Balance sheet & The value of owner's equity, including initial and final value \\
\midrule
Financial Literacy Detection-Total Liabilities and Owner's Equity & \{5,1,5\} & Based on the balance sheet, identify and extract the specific line items and value of the total amount of liabilities and equity recognised in the statement of financial position, along with the relevant financial data involved in the calculation, including initial and final value. In addition, decompose this item into 2 component sub-items, all of which must also originate from the input statement. For each sub-item, output its initial and final values. & Balance sheet & i) The value of liabilities and owner's equity, including initial and final value\newline ii) The value of each core sub-item under owner's equity, including initial and final value \\
\midrule
Financial Literacy Detection-Accounts Receivable \& Accounts Payable & \{2,1,2\} & Based on the balance sheet, identify and extract the two specific line items and value of amounts owed to the entity for goods or services sold or provided on credit and the amounts owed by the entity for goods or services received or purchased on credit, including initial and final value. For multiple outputs, maintain the original line item order as shown in the input statement. & Balance sheet & The value of accounts receivable and accounts payable, including initial and final value \\
\midrule
Financial Literacy Detection-Cash on Hand \& Fixed Assets \& Taxes Payable & \{3,1,3\} & Based on the balance sheet, identify and extract the three specific line items and value of cash held by an entity that is available for use in its day-to-day operations, tangible items that are held for use in the production or supply of goods or services, for rental to others, or for administrative purposes and the amounts of taxes accrued but not yet paid, including initial and final value. For multiple outputs, maintain the original line item order as shown in the input statement. & Balance sheet & The value of cash on hand, fixed assets and taxes payable, including initial and final value \\
\midrule
Financial Literacy Detection-Interest Receivable \& Accumulated Depreciation \& Taxes Payable \& Paid-in Capital & \{4,1,4\} & Based on the balance sheet, identify and extract the four specific line items and value of amounts of interest accrued but not yet received, accumulated the systematic allocation of the depreciable amount of an asset over its useful life, tax payable and capital contributed by shareholders in exchange for shares, including initial and final value. For multiple outputs, maintain the original line item order as shown in the input statement. & Balance sheet & The value of interest receivable, accumulated depreciation, tax payable and paid-in capital, including initial and final value \\

\bottomrule
\bottomrule
\end{tabular}}
\end{table}

\begin{table}[htbp]
\raggedright  
\caption{Task Information Table - Income Statement Detection in Financial Literacy}
\resizebox{\textwidth}{!}{
\begin{tabular}{@{}p{4cm}p{1.5cm}p{10cm}p{2.0cm}p{5cm}@{}} 
\toprule
\midrule
\textbf{Task Name} & \textbf{\{$\alpha$,$\beta$,$\gamma$\}} & \textbf{Task Description} & \textbf{Input} & \textbf{Output} \\
\midrule
\midrule
Financial Literacy Detection-Cost of Goods Sold & \{1,1,1\} & Based on the income statement, identify and extract the specific line items and value of carrying amount of inventories sold during the reporting period & Income statement & The value of cost of goods sold \\
\midrule
Financial Literacy Detection-Main Business Revenue & \{1,1,1\} & Based on the income statement, identify and extract the specific line items and value of income arising in the course of the entity's core operating activities & Income statement & The value of main business revenue \\
\midrule
Financial Literacy Detection-Gross Profit & \{1,1,1\} & Based on the income statement, identify and extract the specific line items and value of difference between revenue and cost & Income statement & The value of gross profit \\
\midrule
Financial Literacy Detection-Interest Income & \{1,1,1\} & Based on the income statement, identify and extract the specific line items and value of income earned by an entity from financial assets & Income statement & The value of interest income \\
\midrule
Financial Literacy Detection-Administrative Expenses & \{1,1,1\} & Based on the income statement, identify and extract the specific line items and value of costs of general management and administration of the entity as a whole & Income statement & The value of administrative expenses \\
\midrule
Financial Literacy Detection-Selling Expenses & \{1,1,1\} & Based on the income statement, identify and extract the specific line items and value of costs incurred to secure customer orders and to deliver the goods and services to customers & Income statement & The value of selling expenses \\
\midrule
Financial Literacy Detection-Financial Expenses & \{1,1,1\} & Based on the income statement, identify and extract the specific line items and value of financing costs incurred by an enterprise to raise funds needed for production and operation & Income statement & The value of financial expenses \\
\midrule
Financial Literacy Detection-Accumulated Depreciation & \{1,1,1\} & Based on the income statement, identify and extract the specific line items and value of accumulated the systematic allocation of the depreciable amount of an asset over its useful life & Income statement & The value of accumulated depreciation \\
\midrule
Financial Literacy Detection-Tax Expense & \{1,1,1\} & Based on the income statement, identify and extract the specific line items and value of total amount of taxes an entity is expected to pay or recover during a reporting period & Income statement & The value of tax expense \\
\midrule
Financial Literacy Detection-Total Revenue & \{2,1,2\} & Based on the income statement, identify and extract the specific line items and value of total income arising in the course of an entity's ordinary activities, along with the values and names of its constituent line items. In addition, decompose this item into 1 component sub-items, all of which must also originate from the input statement. For each sub-item, output its initial and final values. & Income statement & i) The value of revenue\newline ii) The value of each sub-item under revenue \\
\midrule
Financial Literacy Detection-Total Expenses & \{5,1,5\} & Based on the income statement, identify and extract the specific line items and value of operating expenses, along with the relevant financial data involved in the calculation. In addition, decompose this item into 4 component sub-items, all of which must also originate from the input statement. For each sub-item, output its initial and final values. & Income statement & i) The value of total expenses\newline ii) The value of each sub-item under operating expenses \\
\midrule
Financial Literacy Detection-Profit Before Tax & \{5,1,5\} & Based on the income statement, identify and extract the specific line items and value of profit or loss for a period before deducting tax expense, along with the relevant financial data involved in the calculation. In addition, decompose this item into 4 component sub-items, all of which must also originate from the input statement. For each sub-item, output its initial and final values. & Income statement & i) The value of profit before tax\newline ii) The value of each sub-item under profit before tax \\
\midrule
Financial Literacy Detection-Net Profit & \{6,1,6\} & Based on the income statement, identify and extract the specific line items and value of the amount of profit an entity retains after all expenses, including operating costs, interest, taxes, depreciation, and amortization, have been deducted from total revenue, along with the relevant financial data involved in the calculation. In addition, decompose this item into 5 component sub-items, all of which must also originate from the input statement. For each sub-item, output its initial and final values. & Income statement & i) The value of net profit\newline ii) The value of core sub-item under operating expenses \\
\midrule
Financial Literacy Detection-Main Business Revenue \& Cost of Goods Sold & \{2,1,2\} & Based on the income statement, identify and extract the two specific line items and value of income arising in the course of the entity's core operating activities and carrying amount of inventories sold during the reporting period. For multiple outputs, maintain the original line item order as shown in the input statement. & Income statement & The value of main business revenue and cost of goods sold \\
\midrule
Financial Literacy Detection-Total Revenue \& Cost of Goods Sold \& Administrative Expenses & \{3,1,3\} & Based on the income statement, identify and extract the three specific line items and value of total income arising in the course of an entity's ordinary activities, carrying amount of inventories sold during the reporting period, and costs of general management and administration of the entity as a whole. For multiple outputs, maintain the original line item order as shown in the input statement. & Income statement & The value of revenue, cost of goods sold and administrative expenses \\
\midrule
Financial Literacy Detection-Selling Expenses \& Depreciation \& Financial Expenses \& Tax Expense & \{4,1,4\} & Based on the income statement, identify and extract the four specific line items and value of financing costs incurred by an enterprise to raise funds needed for production and operation, the systematic allocation of the depreciable amount of an asset over its useful life, and total amount of taxes an entity is expected to pay or recover during a reporting period. For multiple outputs, maintain the original line item order as shown in the input statement. & Income statement & The value of financial expenses, selling expenses, depreciation and tax expense \\
\bottomrule
\bottomrule
\end{tabular}}
\end{table}

\begin{table}[htbp]
\raggedright  
\caption{Task Information Table - Cash Flow Statement Detection in Financial Literacy}
\resizebox{\textwidth}{!}{
\begin{tabular}{@{}p{4cm}p{1.5cm}p{10cm}p{2.2cm}p{5cm}@{}} 
\toprule
\midrule
\textbf{Task Name} & \textbf{\{$\alpha$,$\beta$,$\gamma$\}} & \textbf{Task Description} & \textbf{Input} & \textbf{Output} \\
\midrule
\midrule
Financial Literacy Detection-Net Profit & \{1,1,1\} & Based on the cash flow statement, identify and extract the specific line items and value of the amount of profit an entity retains after all expenses, including operating costs, interest, taxes, depreciation, and amortization, have been deducted from total revenue & Cash flow statement & The value of net profit \\
\midrule
Financial Literacy Detection-Depreciation & \{1,1,1\} & Based on the cash flow statement, identify and extract the specific line items and value of the systematic allocation of the depreciable amount of an asset over its useful life & Cash flow statement & The value of depreciation \\
\midrule
Financial Literacy Detection-Decrease in Accounts Receivable & \{1,1,1\} & Based on the cash flow statement, identify and extract the specific line items and value of decrease in amounts owed to the entity for goods or services sold or provided on credit & Cash flow statement & The value of decrease in accounts receivable \\
\midrule
Financial Literacy Detection-Decrease in Inventory & \{1,1,1\} & Based on the cash flow statement, identify and extract the specific line items and value of reduction in the amount of assets held for sale in the ordinary course of business, in production for such sale, or in the process of being manufactured during the period & Cash flow statement & The value of decrease in inventory \\
\midrule
Financial Literacy Detection-Increase in Accounts Payable & \{1,1,1\} & Based on the cash flow statement, identify and extract the specific line items and value of the addition in the amount owed by the entity for goods or services received or purchased on credit during the period & Cash flow statement & The value of increase in accounts payable \\
\midrule
Financial Literacy Detection-Increase in Taxes Payable & \{1,1,1\} & Based on the cash flow statement, identify and extract the specific line items and value of increase in the amounts of taxes accrued but not yet paid & Cash flow statement & The value of increase in taxes payable \\
\midrule
Financial Literacy Detection-Purchase of Fixed Assets & \{1,1,1\} & Based on the cash flow statement, identify and extract the specific line items and value of cash payments to acquire property, plant and equipment and other long-term assets & Cash flow statement & The value of purchased of fixed assets \\
\midrule
Financial Literacy Detection-Beginning Cash and Cash Equivalents Balance & \{1,1,1\} & Based on the cash flow statement, identify and extract the specific line items and value of cash and cash equivalents at the beginning of the period & Cash flow statement & The value of Beginning Cash and Cash Equivalents Balance \\
\midrule
Financial Literacy Detection-Ending Cash and Cash Equivalents Balance & \{1,1,1\} & Based on the cash flow statement, identify and extract the specific line items and value of cash and cash equivalents at the end of the period & Cash flow statement & The value of Ending Cash and Cash Equivalents Balance \\
\midrule
Financial Literacy Detection-Net Cash Flow from Operating Activities & \{7,1,7\} & Based on the net cash flow statement, identify and extract the specific line items and value of net amount of cash and cash equivalents generated from an entity's activities that are the principal revenue-producing activities of the entity and other activities that are not investing or financing activities, along with the relevant financial data involved in the calculation. In addition, decompose this item into 7 component sub-items, all of which must also originate from the input statement. For each sub-item, output its initial and final values. & Cash flow statement & i) The value of net cash flow from operating activities\newline ii) The value of core sub-item under net cash flow from operating activities \\
\midrule
Financial Literacy Detection-Net Cash Flow from Investing Activities & \{2,1,2\} & Based on the cash flow statement, identify and extract the specific line items and value of net amount of cash and cash equivalents generated from an entity's activities that are the acquisition and disposal of long-term assets and other investments not included in cash equivalents and the receipt of interest and dividends, along with the relevant financial data involved in the calculation. In addition, decompose this item into 1 component sub-items, all of which must also originate from the input statement. For each sub-item, output its initial and final values. & Cash flow statement & i) The value of net cash flow from investing activities\newline ii) The value of core sub-item under net cash flow from operating activities \\
\midrule
Financial Literacy Detection-Net Increase in Cash and Cash Equivalents & \{3,1,3\} & Based on the cash flow statement, identify and extract the specific line items and value of net addition in the amount of cash and cash equivalents during the period, along with the relevant financial data involved in the calculation. In addition, decompose this item into 2 component sub-items, all of which must also originate from the input statement. For each sub-item, output its initial and final values. & Cash flow statement & i) The value of cash flow from net increase in cash and cash equivalents\newline ii) The value of core sub-item under net cash flow from operating activities \\
\midrule
Financial Literacy Detection-Net Profit \& Purchase of Fixed Assets & \{2,1,2\} & Based on the cash flow statement, identify and extract the two specific line items and value of the amount of profit an entity retains after all expenses, including operating costs, interest, taxes, depreciation, and amortization, have been deducted from total revenue and cash payments to acquire property, plant and equipment and other long-term assets. For multiple outputs, maintain the original line item order as shown in the input statement. & Cash flow statement & The value of net profit and purchased in fixed assets \\
\midrule
Financial Literacy Detection-Increase in Accounts Payable \& Purchase of Fixed Assets \& Beginning Cash Balance & \{3,1,3\} & Based on the cash flow statement, identify and extract the three specific line items and value of the addition in the amount owed by the entity for goods or services received or purchased on credit during the period, acquisition of property, plant and equipment, and cash and cash equivalents at the beginning of the period. For multiple outputs, maintain the original line item order as shown in the input statement. & Cash flow statement & The value of increase in accounts payable, purchase of fixed assets and beginning cash and cash equivalents balance \\
\midrule
Financial Literacy Detection-Depreciation \& Decrease in Inventory \& Net Cash Flow from Investing \& Net Increase & \{4,1,4\} & Based on the cash flow statement, identify and extract the four specific line items and value of the systematic allocation of the depreciable amount of an asset over its useful life, reduction in the amount of assets held for sale in the ordinary course of business, in production for such sale, or in the process of being manufactured during the period, net amount of cash and cash equivalents generated from an entity's activities that are the acquisition and disposal of long-term assets and other investments not included in cash equivalents and the receipt of interest and dividends, and net addition in the amount of cash and cash equivalents during the period. For multiple outputs, maintain the original line item order as shown in the input statement. & Cash flow statement & The value of depreciation, decrease in inventory, net cash flow from investing activities and net increase \\
\bottomrule
\bottomrule
\end{tabular}}
\end{table}

\begin{table}[htbp]
\raggedright  
\caption{Task Information Table - Financial Statement Detection in Financial Literacy}
\resizebox{\textwidth}{!}{
\begin{tabular}{@{}p{4cm}p{1.5cm}p{10cm}p{2cm}p{5cm}@{}} 
\toprule
\midrule
\textbf{Task Name} & \textbf{\{$\alpha$,$\beta$,$\gamma$\}} & \textbf{Task Description} & \textbf{Input} & \textbf{Output} \\
\midrule
\midrule
Financial Literacy Detection-Interest Receivable & \{1,3,1\} & Based on the all financial statements, identify and extract the specific line items and value of amounts of interest accrued but not yet received, including initial and final value & All financial statements & The value of interest receivable, including initial and final value \\
\midrule
Financial Literacy Detection-Paid-in Capital & \{1,3,1\} & Based on the financial statement, identify and extract the specific line items and value of capital contributed by shareholders in exchange for shares, including initial and final value & All financial statements & The value of paid-in capital, including initial and final value \\
\midrule
Financial Literacy Detection-Cost of Goods Sold & \{1,3,1\} & Based on the financial statement, identify and extract the specific line items and value of carrying amount of inventories sold during the reporting period & All financial statements & The value of cost of goods sold \\
\midrule
Financial Literacy Detection-Selling Expenses & \{1,3,1\} & Based on the financial statement, identify and extract the specific line items and value of selling expenses & All financial statements & The value of selling expenses \\
\midrule
Financial Literacy Detection-Tax Expense & \{1,3,1\} & Based on the financial statement, identify and extract the specific line items and value of total amount of taxes an entity is expected to pay or recover during a reporting period & All financial statements & The value of tax expense \\
\midrule
Financial Literacy Detection-Depreciation & \{1,3,1\} & Based on the financial statement, identify and extract the specific line items and value of the systematic allocation of the depreciable amount of an asset over its useful life & All financial statements & The value of depreciation \\
\midrule
Financial Literacy Detection-Increase in Accounts Payable & \{1,3,1\} & Based on the financial statement, identify and extract the specific line items and value of the addition in the amount owed by the entity for goods or services received or purchased on credit during the period & All financial statements & The value of increase in accounts payable \\
\midrule
Financial Literacy Detection-Beginning Cash and Cash Equivalents Balance & \{1,3,1\} & Based on the financial statement, identify and extract the specific line items and value of cash and cash equivalents at the beginning of the period & All financial statements & The value of Beginning Cash and Cash Equivalents Balance \\
\midrule
Financial Literacy Detection-Interest Receivable \& Net Increase in Cash & \{2,3,2\} & Based on the all financial statements, identify and extract the two specific line items and end value of amounts of interest accrued but not yet received, and net addition in the amount of cash and cash equivalents during the period. For multiple outputs, group them by financial statement in the following order: 1.Balance Sheet 2.Income Statement 3.Cash Flow Statement. Within each group, maintain the original line item order as shown in the input statement. & All financial statements & The end value of interest receivable and net increase of cash and cash equivalents \\
\midrule
Financial Literacy Detection-Bank Deposits \& Interest Income & \{2,3,2\} & Based on the all financial statements, identify and extract the two specific line items and end value of funds deposit into a bank, and income earned by an entity from financial assets. For multiple outputs, group them by financial statement in the following order: 1.Balance Sheet 2.Income Statement 3.Cash Flow Statement. Within each group, maintain the original line item order as shown in the input statement. & All financial statements & The end value of bank deposits and interest income \\
\midrule
Financial Literacy Detection-Selling Expenses \& Purchase of Fixed Assets & \{2,3,2\} & Based on the all financial statements, identify and extract the two specific line items and value of costs incurred to secure customer orders and to deliver the goods and services to customers, and cash payments to acquire property, plant and equipment and other long-term assets. For multiple outputs, group them by financial statement in the following order: 1.Balance Sheet 2.Income Statement 3.Cash Flow Statement. Within each group, maintain the original line item order as shown in the input statement. & All financial statements & The value of selling expenses and purchased of fixed assets \\
\midrule
Financial Literacy Detection-Accounts Receivable \& Financial Expenses \& Fixed Assets & \{3,3,3\} & Based on the all financial statements, identify and extract the three specific line items and end value of amounts owed to the entity for goods or services sold or provided on credit, financing costs incurred by an enterprise to raise funds needed for production and operation, and cash payments to acquire property, plant and equipment and other long-term assets. For multiple outputs, group them by financial statement in the following order: 1.Balance Sheet 2.Income Statement 3.Cash Flow Statement. Within each group, maintain the original line item order as shown in the input statement. & All financial statements & The end value of accounts receivable, financial expenses and purchased of fixed assets \\
\midrule
Financial Literacy Detection-Taxes Payable \& Revenue \& Operating Cash Flow & \{3,3,3\} & Based on the all financial statements, identify and extract the three specific line items and end value of net amount of cash and cash equivalents generated from an entity's activities that are the principal revenue-producing activities of the entity and other activities that are not investing or financing activities, the amounts of taxes accrued but not yet paid, and total income arising in the course of an entity's ordinary activities. For multiple outputs, group them by financial statement in the following order: 1.Balance Sheet 2.Income Statement 3.Cash Flow Statement. Within each group, maintain the original line item order as shown in the input statement. & All financial statements & The end value of net cash flow from operating activities, taxes payable and revenue \\
\midrule
Financial Literacy Detection-Paid-in Capital \& Profit Before Tax \& Accounts Payable & \{3,3,3\} & Based on the all financial statements, identify and extract the three specific line items and end value of profit or loss for a period before deducting tax expense, the addition in the amount owed by the entity for goods or services received or purchased on credit during the period, and capital contributed by shareholders in exchange for shares. For multiple outputs, group them by financial statement in the following order: 1.Balance Sheet 2.Income Statement 3.Cash Flow Statement. Within each group, maintain the original line item order as shown in the input statement. & All financial statements & The end value of profit before tax, increase in accounts payable and paid-in capital \\

\bottomrule
\bottomrule
\end{tabular}}
\end{table}

\begin{table}[htbp]
\raggedright  
\caption{Task Information Table - Balance Sheet Generation in Accounting}
\resizebox{\textwidth}{!}{
\begin{tabular}{@{}p{3cm}p{1.5cm}p{6.5cm}p{3.0cm}p{4.5cm}@{}} 
\toprule
\midrule
\textbf{Task Name} & \textbf{\{$\alpha$,$\beta$,$\gamma$\}} & \textbf{Task Description} & \textbf{Input} & \textbf{Output} \\
\midrule
\midrule
Balance Sheet-Cash on Hand &  \{1,1,2\} & Based on transactions data, calculate the total amount of cash on hand item in the balance sheet, including both the initial and final amounts. & All transactions data & The total value of Cash on Hand, including initial and final value \\
\midrule
Balance Sheet-Bank Deposits &  \{1,1,2\} & Based on transactions data, calculate the bank deposits item in the balance sheet, including both the initial and final amounts. & All transactions data & The total value of Bank Deposits, including initial and final value \\
\midrule
Balance Sheet-Inventory &  \{1,1,2\} & Based on transactions data, calculate the inventory item in the balance sheet, including both the initial and final amounts. & All transactions data & The total value of Inventory, including initial and final value \\
\midrule
Balance Sheet-Accounts Receivable &  \{1,1,2\} & Based on transactions data, calculate the accounts receivable item in the balance sheet, including both the initial and final amounts. & All transactions data & The total value of Accounts Receivable, including initial and final value \\
\midrule
Balance Sheet-Interest Receivable &  \{1,1,2\} & Based on transactions data, calculate the Interest Receivable item in the balance sheet, including both the initial and final amounts. & All transactions data & The total value of Interest Receivable, including initial and final value \\
\midrule
Balance Sheet-Current Assets &  \{5,1,2\} & Based on transactions data, calculate the current assets item in the balance sheet, including both the initial and final amounts. & All transactions data & The total value of Current assets, including initial and final value\\
\midrule
Balance Sheet-Accumulated Depreciation &  \{1,1,2\} & Based on transactions data, calculate the Accumulated Depreciation item in the balance sheet, including both the initial and final amounts. & All transactions data & The total value of Accumulated Depreciation, including initial and final value \\
\midrule
Balance Sheet-Fixed Assets net &  \{1,1,2\} & Based on transactions data, calculate the Fixed Assets net item in the balance sheet, including both the initial and final amounts. & All transactions data & The total value of Fixed Assets net, including initial and final value \\
\midrule
Balance Sheet-Non-current Assets & \{2,1,2\} & Based on transactions data, calculate the property, plant and non-current assets item in the balance sheet, including both the initial and final amounts. & All transactions data & The total value of Non-current Assets, including initial and final value \\
\midrule
Balance Sheet-Total Assets & \{7,1,2\} & Based on transactions data, calculate the total assets item in the balance sheet, including both the initial and final amounts. & All transactions data & The total value of Total Assets, including initial and final value \\
\midrule
Balance Sheet-Accounts Payable & \{1,1,2\} & Based on transactions data, calculate the accounts payable item in the balance sheet, including both the initial and final amounts. & All transactions data & The total value of Accounts Payable, including initial and final value \\
\midrule
Balance Sheet-Taxes Payable & \{1,1,2\} & Based on transactions data, calculate the taxes payable item in the balance sheet, including both the initial and final amounts. & All transactions data & The total value of Taxes Payable, including initial and final value \\
\midrule
Balance Sheet-Current Liabilities &  \{2,1,2\} & Based on transactions data, calculate the current liabilities item in the balance sheet, including both the initial and final amounts. & All transactions data & The total value of Current Liabilities,including initial and final value\\
\midrule
Balance Sheet-Total Liabilities &  \{2,1,2\} & Based on transactions data, calculate the total liabilities item in the balance sheet, including both the initial and final amounts. & All transactions data & The total value of Total Liabilities, including initial and final value \\
\midrule
Balance Sheet-Paid-in Capital &  \{1,1,2\} & Based on transactions data, calculate the Paid-in Capital item in the balance sheet, including both the initial and final amounts. & All transactions data & The total value of Paid-in Capital, including initial and final value \\
\midrule
Balance Sheet-Retained Earnings &  \{1,1,2\} & Based on transactions data,calculate the retained earnings item in the balance sheet, including both the initial and final amounts. & All transactions data & The total value of Retained Earnings, including initial and final value \\
\midrule
Balance Sheet-Total Owner's Equity &  \{2,1,2\} & Based on transactions data,calculate the total owner's equity item in the balance sheet, including both the initial and final amounts. & All transactions data & The total value of Total Owner's Equity, including initial and final value \\
\midrule
Balance Sheet-Total Liabilities and Owner's Equity &  \{4,1,2\} & Based on transactions data,calculate the total liabilities and owner's equity item in the balance sheet, including both the initial and final amounts. & All transactions data & The total value of Total Liabilities and Owner's Equity, including initial and final value \\
\midrule

Balance Sheet-Balance Sheet &  \{37,1,2\} & Based on transactions data, directly generate a complete balance sheet, including both the initial and final amounts. & All transactions data & The complete balance sheet \\

\bottomrule
\bottomrule
\end{tabular}}
\end{table}

\clearpage

\begin{table}[htbp]
\centering 
\caption{Task Information Table - Income Statement Generation in Accounting}
\resizebox{\textwidth}{!}{
\begin{tabular}{@{}p{3cm}p{1.5cm}p{6.5cm}p{3.0cm}p{4.5cm}@{}} 
\toprule
\midrule
\textbf{Task Name} & \textbf{\{$\alpha$,$\beta$,$\gamma$\}} & \textbf{Task Description} & \textbf{Input} & \textbf{Output} \\
\midrule

\midrule
Income Statement-Main Business Revenue & \{1,1,1\} & Based on transactions data, calculate the final Main Business Revenue item in the Income.  & Statement All transactions data & The final total value of Main Business Revenue \\
\midrule
Income Statement-Total Revenue & \{1,1,1\} & Based on transactions data, calculate the final Total Revenue item in the Income Statement. & All transactions data & The final total value of Total Revenue \\
\midrule
Income Statement-Cost of Goods Sold & \{1,1,1\} & Based on transactions data, calculate the final Cost of Goods Sold item in the Income Statement. & All transactions data & The final total value of Cost of Goods Sold \\
\midrule
Income Statement-Total Cost & \{1,1,1\} & Based on transactions data, calculate the final Total Cost item in the Income Statement. & All transactions data & The final total value of Total Cost \\
\midrule
Income Statement-Gross Profit & \{2,1,1\} & Based on transactions data, calculate the final Gross Profit item in the Income Statement. & All transactions data & The final total value of Gross Profit \\
\midrule
Income Statement-Depreciation & \{1,1,1\} & Based on transactions data, calculate the final Depreciation item in the Income Statement. & All transactions data & The final total value of Depreciation \\
\midrule
Income Statement-Administrative Expenses & \{1,1,1\} & Based on transactions data, calculate the final Administrative Expenses item in the Income Statement. & All transactions data & The final total value of Administrative Expenses \\
\midrule
Income Statement-Sales Expenses & \{1,1,1\} & Based on transactions data, calculate the final Sales Expenses item in the Income Statement. & All transactions data & The final total value of Sales Expenses \\
\midrule
Income Statement-Financial Expenses & \{1,1,1\} & Based on transactions data, calculate the final Financial Expenses item in the Income Statement. & All transactions data & The final total value of Financial Expenses \\
\midrule
Income Statement-Total Expenses & \{4,1,1\} & Based on transactions data, calculate the final Total Expenses item in the Income Statement. & All transactions data & The final total value of Total Expenses \\
\midrule
Income Statement-Interest Income & \{1,1,1\} & Based on transactions data, calculate the final Interest Income item in the Income Statement. & All transactions data & The final total value of Interest Income \\
\midrule
Income Statement-Profit Before Tax & \{7,1,1\} & Based on transactions data, calculate the final Profit Before Tax item in the Income Statement. & All transactions data & The final total value of Profit Before Tax \\
\midrule
Income Statement-Tax Expense & \{1,1,1\} & Based on transactions data, calculate the final Tax Expense item in the Income Statement. & All transactions data & The final total value of Tax Expense \\
\midrule
Income Statement-Net Profit & \{8,1,1\} & Based on transactions data, calculate the final Net Profit item in the Income Statement. & All transactions data & The final total value of Net Profit \\
\midrule
Income Statement-Income Statement & \{31,1,1\} & Based on transactions data, directly generate a complete income statement. & All transactions data & The complete income statement \\
\bottomrule
\bottomrule
\end{tabular}}
\end{table}

\begin{table}[htbp]
\raggedright  
\caption{Task Information Table - Cash Flow Statement Generation in Accounting}
\resizebox{\textwidth}{!}{
\begin{tabular}{@{}p{3cm}p{1.5cm}p{6.5cm}p{3.0cm}p{4.5cm}@{}} 
\toprule
\midrule
\textbf{Task Name} & \textbf{\{$\alpha$,$\beta$,$\gamma$\}} & \textbf{Task Description} & \textbf{Input} & \textbf{Output} \\
\midrule
\midrule

Cash Flow Statement-Net profit & \{8,1,1\} & Based on transactions data, calculate the final Net profit item in the Cash Flow Statement. & All transactions data & The final total value of Net profit \\
\midrule
Cash Flow Statement-Depreciation & \{1,1,1\} & Based on transactions data, calculate the final Depreciation item in the Cash Flow Statement. & All transactions data & The final total value of Depreciation \\
\midrule
Cash Flow Statement-Accounts Receivable & \{1,1,1\} & Based on transactions data, calculate the final Accounts Receivable item in the Cash Flow Statement. & All transactions data & The final total value of Accounts Receivable \\
\midrule
Cash Flow Statement-Interest Receivable & \{1,1,1\} & Based on transactions data, calculate the final Interest Receivable item in the Cash Flow Statement. & All transactions data & The final total value of Interest Receivable \\
\midrule
Cash Flow Statement-Inventory & \{1,1,1\} & Based on transactions data, calculate the final Inventory item in the Cash Flow Statement. & All transactions data & The final total value of Inventory \\
\midrule
Cash Flow Statement-Total (Increase) Decrease in Current Assets & \{1,1,1\} & Based on transactions data, calculate the final Total (Increase) Decrease in Current Assets item in the Cash Flow Statement. & All transactions data & The final total value of Total (Increase) Decrease in Current Assets \\
\midrule
Cash Flow Statement-Accounts Payable & \{1,1,1\} & Based on transactions data, calculate the final Accounts Payable item in the Cash Flow Statement. & All transactions data & The final total value of Accounts Payable \\
\midrule
Cash Flow Statement-Tax Payable & \{14,1,1\} & Based on transactions data, calculate the final Tax Payable item in the Cash Flow Statement. & All transactions data & The final total value of Tax Payable \\
\midrule
Cash Flow Statement-Total Increase (Decrease) in Current Liabilities & \{1,1,1\} & Based on transactions data, calculate the final Total Increase (Decrease) in Current Liabilities item in the Cash Flow Statement. & All transactions data & The final total value of Total Increase (Decrease) in Current Liabilities \\
\midrule
Cash Flow Statement-Net Cash Flow from Operating Activities & \{1,1,1\} & Based on transactions data, calculate the final Net Cash Flow from Operating Activities item in the Cash Flow Statement. & All transactions data & The final total value of Net Cash Flow from Operating Activities \\
\midrule
Cash Flow Statement-Purchase of Fixed Assets & \{1,1,1\} & Based on transactions data, calculate the final Purchase of Fixed Assets item in the Cash Flow Statement. & All transactions data & The final total value of Purchase of Fixed Assets \\
\midrule
Cash Flow Statement-Net Cash Flows from Investing Activities & \{1,1,1\} & Based on transactions data, calculate the final Net Cash Flows from Investing Activities item in the Cash Flow Statement. & All transactions data & The final total value of Net Cash Flows from Investing Activities \\
\midrule
Cash Flow Statement-Beginning Cash and Cash Equivalents Balance & \{2,1,1\} & Based on transactions data, calculate the final Beginning Cash and Cash Equivalents Balance item in the Cash Flow Statement. & All transactions data & The final total value of Beginning Cash and Cash Equivalents Balance \\
\midrule
Cash Flow Statement-Ending Cash and Cash Equivalents Balance & \{2,1,1\} & Based on transactions data, calculate the final Ending Cash and Cash Equivalents Balance item in the Cash Flow Statement. & All transactions data & The final total value of Ending Cash and Cash Equivalents Balance \\
\midrule
Cash Flow Statement-Net Increase & \{4,1,1\} & Based on transactions data, calculate the final Net Increase item in the Cash Flow Statement. & All transactions data & The final total value of Net Increase \\
\midrule
Cash Flow Statement-Cash Flow Statement & \{38,1,1\} & Based on transactions data, directly generate a complete Cash Flow Statement. & All transactions data & The complete cash flow statement \\
\bottomrule
\bottomrule

\end{tabular}}
\end{table}

\begin{table}[htbp]
\raggedright
\caption{Task Information Table - Single-Error in Auditing}
\resizebox{\textwidth}{!}{
\begin{tabular}{@{}p{5cm}p{1.5cm}p{10cm}p{3cm}p{4cm}@{}}
\toprule
\midrule
\textbf{Task Name} & \textbf{\{$\alpha$,$\beta$,$\gamma$\}} & \textbf{Task Description} & \textbf{Input} & \textbf{Output} \\
\midrule
\midrule
Find Record Error-Transaction TYPE Record Error & \{13,1,3\} & Analyze each transaction entry and identify any internal inconsistencies or Errors in the recorded information. Some fields—such as invoice—are generally considered more reliable due to their standardized and regulated nature. For each inconsistency you find, output transaction ID, (the incorrect field(s),) their recorded values, and your best estimate of the correct value(s)  based on the other fields in that row. & All transactions data & ID; Recorded Type; Original Type \\
\midrule
Find Record Error-Transaction DATE Record Error & \{13,1,3\} & Analyze each transaction entry and identify any internal inconsistencies or Errors in the recorded information. Some fields—such as invoice—are generally considered more reliable due to their standardized and regulated nature. For each inconsistency you find, output transaction ID, (the incorrect field(s),) their recorded values, and your best estimate of the correct value(s)  based on the other fields in that row. & All transactions data & ID; Recorded Date; Original Date \\
\midrule
Find Record Error-Transaction PAYMENT/RECEIPT\_STATUS Record Error & \{13,1,3\} & Analyze each transaction entry and identify any internal inconsistencies or Errors in the recorded information. Some fields—such as invoice—are generally considered more reliable due to their standardized and regulated nature. For each inconsistency you find, output transaction ID, (the incorrect field(s),) their recorded values, and your best estimate of the correct value(s)  based on the other fields in that row. & All transactions data & ID; Recorded Payment/Receipt Status; Original Payment/Receipt Status \\
\midrule
Find Record Error-Transaction PAYMENT\_METHOD Record Error & \{13,1,3\} & Analyze each transaction entry and identify any internal inconsistencies or Errors in the recorded information. Some fields—such as invoice—are generally considered more reliable due to their standardized and regulated nature. For each inconsistency you find, output transaction ID, (the incorrect field(s),) their recorded values, and your best estimate of the correct value(s)  based on the other fields in that row. & All transactions data & ID; Recorded Payment Method; Original Payment Method \\
\midrule
Find Record Error-Transaction QUANTITY Record Error & \{13,1,3\} & Analyze each transaction entry and identify any internal inconsistencies or Errors in the recorded information. Some fields—such as invoice—are generally considered more reliable due to their standardized and regulated nature. For each inconsistency you find, output transaction ID, (the incorrect field(s),) their recorded values, and your best estimate of the correct value(s)  based on the other fields in that row. & All transactions data & ID; Recorded Quantity; Original Quantity \\
\midrule
Find Record Error-Transaction UNIT\_PRICE Record Error & \{13,1,3\} & Analyze each transaction entry and identify any internal inconsistencies or Errors in the recorded information. Some fields—such as invoice—are generally considered more reliable due to their standardized and regulated nature. For each inconsistency you find, output transaction ID, (the incorrect field(s),) their recorded values, and your best estimate of the correct value(s)  based on the other fields in that row. & All transactions data & ID; Recorded Unit Price; Original Unit Price \\
\midrule
Find Record Error-Transaction RECEIVE\_METHOD Record Error & \{13,1,3\} & Analyze each transaction entry and identify any internal inconsistencies or Errors in the recorded information. Some fields—such as invoice—are generally considered more reliable due to their standardized and regulated nature. For each inconsistency you find, output transaction ID, (the incorrect field(s),) their recorded values, and your best estimate of the correct value(s)  based on the other fields in that row. & All transactions data & ID; Recorded Receive Method; Original Receive Method \\
\midrule
Find Calculation Error-Transaction AMOUNT Calculation Error & \{13,1,3\} & Analyze each transaction entry and identify any internal inconsistencies or Errors in the recorded information. Some fields—such as invoice—are generally considered more reliable due to their standardized and regulated nature. For each inconsistency you find, output transaction ID, (the incorrect field(s),) their recorded values, and your best estimate of the correct value(s)  based on the other fields in that row. & All transactions data & ID; Recorded Amount; Original Amount \\
\midrule
Find Calculation Error-Transaction TAX\_AMOUNT Calculation Error & \{13,1,3\} & Analyze each transaction entry and identify any internal inconsistencies or Errors in the recorded information. Some fields—such as invoice—are generally considered more reliable due to their standardized and regulated nature. For each inconsistency you find, output transaction ID, (the incorrect field(s),) their recorded values, and your best estimate of the correct value(s)  based on the other fields in that row.  & All transactions data & ID; Recorded Tax Amount; Original Tax Amount \\
\midrule
Find Calculation Error-Transaction PROFIT Calculation Error & \{13,1,2\} & Analyze each transaction entry and identify any internal inconsistencies or Errors in the recorded information. Some fields—such as invoice—are generally considered more reliable due to their standardized and regulated nature. For each inconsistency you find, output transaction ID, (the incorrect field(s),) their recorded values, and your best estimate of the correct value(s)  based on the other fields in that row. & All transactions data & ID; Recorded Profit; Original Profit \\
\midrule
Find Transaction Approval Mismatch-Transaction Without PREPARER Error & \{13,1,2\} & Analyze each transaction entry and identify any internal inconsistencies or Errors in the recorded information. Some fields—such as invoice—are generally considered more reliable due to their standardized and regulated nature. For each inconsistency you find, output transaction ID, (the incorrect field(s),) their recorded values, and your best estimate of the correct value(s)  based on the other fields in that row. & All transactions data & ID; Original Preparer \\
\midrule
Find Transaction Approval Mismatch-Transaction Without APPROVER Error & \{13,1,4\} & Analyze each transaction entry and identify any internal inconsistencies or Errors in the recorded information. Some fields—such as invoice—are generally considered more reliable due to their standardized and regulated nature. For each inconsistency you find, output transaction ID, (the incorrect field(s),) their recorded values, and your best estimate of the correct value(s)  based on the other fields in that row. & All transactions data & ID; Recorded Preparer; Original Approver \\
\bottomrule
\bottomrule
\end{tabular}}
\end{table}

\begin{table}[htbp]
\raggedright
\caption{Task Information Table - Double-Error in Auditing}
\resizebox{\textwidth}{!}{
\begin{tabular}{@{}p{5cm}p{1.5cm}p{10cm}p{3cm}p{4cm}@{}}
\toprule
\midrule
\textbf{Task Name} & \textbf{\{$\alpha$,$\beta$,$\gamma$\}} & \textbf{Task Description} & \textbf{Input} & \textbf{Output} \\
\midrule
\midrule
Find Record Error-Transaction TYPE Record Error \& Calculation Error-Transaction TAX\_AMOUNT Calculation Error & \{13,1,4\} & Analyze each transaction entry and identify any internal inconsistencies or Errors in the recorded information. Some fields—such as invoice—are generally considered more reliable due to their standardized and regulated nature. For each inconsistency you find, output transaction ID, (the incorrect field(s),) their recorded values, and your best estimate of the correct value(s)  based on the other fields in that row. & All transactions data & ID; Recorded Tax Amount; Original Tax Amount; Recorded Type; Original Type \\
\midrule
Find Record Error-Transaction PAYMENT/RECEIPT\_STATUS Record Error \& Record Error-Transaction QUANTITY Record Error & \{13,1,4\} & Analyze each transaction entry and identify any internal inconsistencies or Errors in the recorded information. Some fields—such as invoice—are generally considered more reliable due to their standardized and regulated nature. For each inconsistency you find, output transaction ID, (the incorrect field(s),) their recorded values, and your best estimate of the correct value(s)  based on the other fields in that row. & All transactions data & ID; Recorded Payment/Receipt Status; Original Payment/Receipt Status; Recorded Quantity; Original Quantity \\
\midrule
Find Record Error-Transaction QUANTITY Record Error \& Record Error-Transaction TYPE Record Error & \{13,1,4\} & Analyze each transaction entry and identify any internal inconsistencies or Errors in the recorded information. Some fields—such as invoice—are generally considered more reliable due to their standardized and regulated nature. For each inconsistency you find, output transaction ID, (the incorrect field(s),) their recorded values, and your best estimate of the correct value(s)  based on the other fields in that row. & All transactions data & ID; Recorded Quantity; Original Quantity; Recorded Type; Original Type \\
\midrule
Find Record Error-Transaction PAYMENT/RECEIPT\_STATUS Record Error \& Calculation Error-Transaction AMOUNT Calculation Error & \{13,1,5\} & Analyze each transaction entry and identify any internal inconsistencies or Errors in the recorded information. Some fields—such as invoice—are generally considered more reliable due to their standardized and regulated nature. For each inconsistency you find, output transaction ID, (the incorrect field(s),) their recorded values, and your best estimate of the correct value(s)  based on the other fields in that row. & All transactions data & ID; Recorded Payment/Receipt Status; Original Payment/Receipt Status; Recorded Amount; Original Amount \\
\midrule
Find Record Error-Transaction RECEIVE\_METHOD Record Error \& Record Error-Transaction TYPE Record Error & \{13,1,7\} & Analyze each transaction entry and identify any internal inconsistencies or Errors in the recorded information. Some fields—such as invoice—are generally considered more reliable due to their standardized and regulated nature. For each inconsistency you find, output transaction ID, (the incorrect field(s),) their recorded values, and your best estimate of the correct value(s)  based on the other fields in that row. & All transactions data & ID; Recorded Receive Method; Original Receive Method; Recorded Transaction Type; Original Transaction Type \\
\midrule

Find Error-TYPE MISCLASSIFICATION Error \& RECORDING DELAY Error & \{13,1,5\} & Analyze each transaction entry and identify any internal inconsistencies or Errors in the recorded information. Some fields—such as invoice—are generally considered more reliable due to their standardized and regulated nature. For each inconsistency you find, output transaction ID, (the incorrect field(s),) their recorded values, and your best estimate of the correct value(s)  based on the other fields in that row. & All transactions data & ID; Recorded Type; Original Type; Recorded Date; Original Date \\
\midrule
Find Error-TYPE MISCLASSIFICATION Error \& PRICE ANOMALY Error & \{13,1,5\} & Analyze each transaction entry and identify any internal inconsistencies or Errors in the recorded information. Some fields—such as invoice—are generally considered more reliable due to their standardized and regulated nature. For each inconsistency you find, output transaction ID, (the incorrect field(s),) their recorded values, and your best estimate of the correct value(s)  based on the other fields in that row. & All transactions data & ID; Recorded Type; Original Type; Recorded Price; Original Price \\
\midrule
Find Error-TYPE MISCLASSIFICATION Error \& AMOUNT DISCREPANCY Error & \{13,1,5\} & Analyze each transaction entry and identify any internal inconsistencies or Errors in the recorded information. Some fields—such as invoice—are generally considered more reliable due to their standardized and regulated nature. For each inconsistency you find, output transaction ID, (the incorrect field(s),) their recorded values, and your best estimate of the correct value(s)  based on the other fields in that row. & All transactions data & ID; Recorded Type; Original Type; Recorded Amount; Original Amount \\
\midrule
Find Error-RECORDING DELAY Error \& PRICE ANOMALY Error & \{13,1,5\} & Analyze each transaction entry and identify any internal inconsistencies or Errors in the recorded information. Some fields—such as invoice—are generally considered more reliable due to their standardized and regulated nature. For each inconsistency you find, output transaction ID, (the incorrect field(s),) their recorded values, and your best estimate of the correct value(s)  based on the other fields in that row. & All transactions data & ID; Recorded Date; Original Date; Recorded Price; Original Price \\
\midrule
Find Error-RECORDING DELAY Error \& AMOUNT DISCREPANCY Error & \{13,1,5\} & Analyze each transaction entry and identify any internal inconsistencies or Errors in the recorded information. Some fields—such as invoice—are generally considered more reliable due to their standardized and regulated nature. For each inconsistency you find, output transaction ID, (the incorrect field(s),) their recorded values, and your best estimate of the correct value(s)  based on the other fields in that row. & All transactions data & ID; Recorded Date; Original Date; Recorded Amount; Original Amount \\
\midrule
Find Error-PRICE ANOMALY Error \& AMOUNT DISCREPANCY Error & \{13,1,5\} & Analyze each transaction entry and identify any internal inconsistencies or Errors in the recorded information. Some fields—such as invoice—are generally considered more reliable due to their standardized and regulated nature. For each inconsistency you find, output transaction ID, (the incorrect field(s),) their recorded values, and your best estimate of the correct value(s)  based on the other fields in that row. & All transactions data & ID; Recorded Price; Original Price; Recorded Amount; Original Amount \\
\bottomrule
\bottomrule
\end{tabular}}
\end{table}

\begin{table}[htbp]
\raggedright
\caption{Task Information Table - Multi-Error in Auditing}
\resizebox{\textwidth}{!}{
\begin{tabular}{@{}p{5cm}p{1.5cm}p{10cm}p{3cm}p{4cm}@{}}
\toprule
\midrule
\textbf{Task Name} & \textbf{\{$\alpha$,$\beta$,$\gamma$\}} & \textbf{Task Description} & \textbf{Input} & \textbf{Output} \\
\midrule
\midrule

Find Record Error-Transaction PAYMENT/RECEIPT\_STATUS Record Error \& Record Error-Transaction QUANTITY Record Error \& Calculation Error-Transaction PROFIT Calculation Error & \{13,1,5\} & Analyze each transaction entry and identify any internal inconsistencies or Errors in the recorded information. Some fields—such as invoice—are generally considered more reliable due to their standardized and regulated nature. For each inconsistency you find, output transaction ID, (the incorrect field(s),) their recorded values, and your best estimate of the correct value(s)  based on the other fields in that row. & All transactions data & ID; Recorded Payment/Receipt Status; Original Payment/Receipt Status; Recorded Quantity; Original Quantity; Recorded Profit; Original Profit \\

\midrule
Find Error-TYPE MISCLASSIFICATION Error \& RECORDING DELAY Error \& PRICE ANOMALY Error & \{13,1,7\} & Analyze each transaction entry and identify any internal inconsistencies or Errors in the recorded information. Some fields—such as invoice—are generally considered more reliable due to their standardized and regulated nature. For each inconsistency you find, output transaction ID, (the incorrect field(s),) their recorded values, and your best estimate of the correct value(s)  based on the other fields in that row. & All transactions data & ID; Recorded Type; Original Type; Recorded Date; Original Date; Recorded Price; Original Price \\
\midrule
Find Error-TYPE MISCLASSIFICATION Error \& RECORDING DELAY Error \& AMOUNT DISCREPANCY Error & \{13,1,7\} & Analyze each transaction entry and identify any internal inconsistencies or Errors in the recorded information. Some fields—such as invoice—are generally considered more reliable due to their standardized and regulated nature. For each inconsistency you find, output transaction ID, (the incorrect field(s),) their recorded values, and your best estimate of the correct value(s)  based on the other fields in that row. & All transactions data & ID; Recorded Type; Original Type; Recorded Date; Original Date; Recorded Amount; Original Amount \\
\midrule
Find Error-TYPE MISCLASSIFICATION Error \& PRICE ANOMALY Error \&
AMOUNT DISCREPANCY Error & \{13,1,7\} & Analyze each transaction entry and identify any internal inconsistencies or Errors in the recorded information. Some fields—such as invoice—are generally considered more reliable due to their standardized and regulated nature. For each inconsistency you find, output transaction ID, (the incorrect field(s),) their recorded values, and your best estimate of the correct value(s)  based on the other fields in that row. & All transactions data & ID; Recorded Type; Original Type; Recorded Price; Original Price; Recorded Amount; Original Amount \\
\midrule
Find Error-RECORDING DELAY Error \& PRICE ANOMALY Error \& AMOUNT DISCREPANCY Error & \{13,1,7\} & Analyze each transaction entry and identify any internal inconsistencies or Errors in the recorded information. Some fields—such as invoice—are generally considered more reliable due to their standardized and regulated nature. For each inconsistency you find, output transaction ID, (the incorrect field(s),) their recorded values, and your best estimate of the correct value(s)  based on the other fields in that row. & All transactions data & ID; Recorded Date; Original Date; Recorded Price; Original Price; Recorded Amount; Original Amount \\
\midrule
Find Error-TAX Error \& PRICE ANOMALY Error \& AMOUNT DISCREPANCY Error \& RECORDING DELAY Error & \{13,1,9\} & Analyze each transaction entry and identify any internal inconsistencies or Errors in the recorded information. Some fields—such as invoice—are generally considered more reliable due to their standardized and regulated nature. For each inconsistency you find, output transaction ID, (the incorrect field(s),) their recorded values, and your best estimate of the correct value(s)  based on the other fields in that row. & All transactions data & ID; Recorded Tax; Original Tax; Recorded Price; Original Price; Recorded Amount; Original Amount; Recorded Date; Original Date \\
\midrule
Find Error-TAX Error \& PRICE ANOMALY Error \& AMOUNT DISCREPANCY Error \& TYPE MISCLASSIFICATION Error & \{13,1,9\} & Analyze each transaction entry and identify any internal inconsistencies or Errors in the recorded information. Some fields—such as invoice—are generally considered more reliable due to their standardized and regulated nature. For each inconsistency you find, output transaction ID, (the incorrect field(s),) their recorded values, and your best estimate of the correct value(s)  based on the other fields in that row. & All transactions data & ID; Recorded Tax; Original Tax; Recorded Price; Original Price; Recorded Amount; Original Amount; Recorded Type; Original Type \\
\midrule
Find Error-TAX Error \& PRICE ANOMALY Error \& AMOUNT DISCREPANCY Error \& QUANTITY MISMATCH Error & \{13,1,9\} & Analyze each transaction entry and identify any internal inconsistencies or Errors in the recorded information. Some fields—such as invoice—are generally considered more reliable due to their standardized and regulated nature. For each inconsistency you find, output transaction ID, (the incorrect field(s),) their recorded values, and your best estimate of the correct value(s)  based on the other fields in that row. & All transactions data & ID; Recorded Tax; Original Tax; Recorded Price; Original Price; Recorded Amount; Original Amount; Recorded Quantity; Original Quantity \\
\midrule
Find Error-PRICE ANOMALY Error \& AMOUNT DISCREPANCY Error \& RECORDING DELAY Error \& QUANTITY MISMATCH Error & \{13,1,9\} & Analyze each transaction entry and identify any internal inconsistencies or Errors in the recorded information. Some fields—such as invoice—are generally considered more reliable due to their standardized and regulated nature. For each inconsistency you find, output transaction ID, (the incorrect field(s),) their recorded values, and your best estimate of the correct value(s)  based on the other fields in that row. & All transactions data & ID; Recorded Price; Original Price; Recorded Amount; Original Amount; Recorded Date; Original Date; Recorded Quantity; Original Quantity \\
\midrule
Find Error-TAX Error \& PRICE ANOMALY Error \& AMOUNT DISCREPANCY Error \& RECORDING DELAY Error  \& TYPE MISCLASSIFICATION Error & \{13,1,11\} & Analyze each transaction entry and identify any internal inconsistencies or Errors in the recorded information. Some fields—such as invoice—are generally considered more reliable due to their standardized and regulated nature. For each inconsistency you find, output transaction ID, (the incorrect field(s),) their recorded values, and your best estimate of the correct value(s)  based on the other fields in that row. & All transactions data & ID; Recorded Tax; Original Tax; Recorded Price; Original Price; Recorded Amount; Original Amount; Recorded Date; Original Date; Recorded Type; Original Type \\
\midrule
Find Error-TAX Error \& PRICE ANOMALY Error \& RECORDING DELAY Error \& TYPE MISCLASSIFICATION Error \& QUANTITY MISMATCH Error & \{13,1,11\} & Analyze each transaction entry and identify any internal inconsistencies or Errors in the recorded information. Some fields—such as invoice—are generally considered more reliable due to their standardized and regulated nature. For each inconsistency you find, output transaction ID, (the incorrect field(s),) their recorded values, and your best estimate of the correct value(s)  based on the other fields in that row. & All transactions data & ID; Recorded Tax; Original Tax; Recorded Price; Original Price; Recorded Amount; Original Amount; Recorded Date; Original Date; Recorded Type; Original Type \\
\midrule
Find Error-PRICE ANOMALY Error \& AMOUNT DISCREPANCY Error \& RECORDING DELAY Error \& TYPE MISCLASSIFICATION Error \& QUANTITY MISMATCH Error & \{13,1,11\} & Analyze each transaction entry and identify any internal inconsistencies or Errors in the recorded information. Some fields—such as invoice—are generally considered more reliable due to their standardized and regulated nature. For each inconsistency you find, output transaction ID, (the incorrect field(s),) their recorded values, and your best estimate of the correct value(s)  based on the other fields in that row. & All transactions data & ID; Recorded Price; Original Price; Recorded Amount; Original Amount; Recorded Date; Original Date; Recorded Type; Original Type; Recorded Quantity; Original Quantity \\
\bottomrule
\bottomrule
\end{tabular}}
\end{table}

\begin{table}[htbp]
\raggedright
\caption{Task Information Table - Single-Capability in Consulting }
\resizebox{\textwidth}{!}{
\begin{tabular}{@{}p{6cm}p{1.5cm}p{6cm}p{4cm}p{4cm}@{}}

\toprule
\midrule
\textbf{Task Name} & \textbf{\{$\alpha$,$\beta$,$\gamma$\}} & \textbf{Task Description} & \textbf{Input} & \textbf{Output} \\
\midrule
\midrule
Analyze Balance Sheet-Calculate Current Ratio & \{2,1,1\} & Based on the balance sheet, calculate the Current Ratio as of the end of the reporting period & Balance sheet & The value of Current Ratio \\
\midrule
Analyze Balance Sheet-calculate Quick Ratio & \{6,1,1\} & Based on the balance sheet as of the end of the reporting period & Balance Sheet & The value of Quick Ratio\\
\midrule
Analyze Balance Sheet-calculate Debt to Asset Ratio & \{2,1,1\} & Based on the balance sheet as of the end of the reporting period & Balance Sheet & The value of Debt to Asset Ratio\\
\midrule
Analyze Balance Sheet-calculate Debt to Equity Ratio & \{2,1,1\} & Based on the balance sheet as of the end of the reporting period & Balance Sheet & The value of Debt to Equity Ratio \\
\midrule
Analyze Income Statement-Gross Profit Margin & \{2,1,1\} & Based on the income statement, calculate the Gross Profit Margin & Income Statement & The value of Gross Profit Margin\\
\midrule
Analyze Income Statement-Net Profit Margin & \{2,1,1\} & Based on the income statement, calculate the Net Profit Margin & Income Statement  & The value of Net Profit Margin \\
\midrule
Analyze Cash Flow Statement-FCF & \{2,1,1\} & Based on the cash flow statement, calculate the FCF & Cash Flow Statement & The value of FCF \\
\midrule
Analyze Cash Flow Statement-Net Cash Ratio & \{2,1,1\} & Based on the cash flow statement, calculate the Net Cash Ratio & Cash Flow Statement & The value of Net Cash Ratio \\
\midrule
Analyze Financial Statement-Cash to Current Debt Ratio & \{2,1,1\} & Based on the three financial statements, calculate the Cash to Current Debt Ratio & All Financial Statements & The value of Cash to Current Debt Ratio \\
\midrule
Analyze Financial Statement-Operating Cash Flow to Current Liabilities Ratio & \{3,3,1\} & Based on the three financial statements, calculate the Operating Cash Flow to Current Liabilities Ratio & All Financial Statements & The value of Operating Cash Flow to Current Liabilities Ratio \\
\midrule
Analyze Financial Statement-ROA & \{3,3,1\} & Based on the three financial statements, calculate the ROA & All Financial Statements & The value of ROA\\
\midrule
Analyze Financial Statement-ROE & \{3,3,1\} & Based on the three financial statements, calculate the ROE & All Financial Statements & The value of ROE\\
\midrule
Analyze Financial Statement-Inventory Turnover Ratio & \{3,3,1\} & Based on the three financial statements, calculate the Inventory Turnover Ratio & All Financial Statements & The value of Inventory Turnover Ratio\\
\midrule
Analyze Financial Statement-Accounts Receivable Turnover Ratio & \{3,3,1\} & Based on the three financial statements, calculate the Accounts Receivable Turnover Ratio & All Financial Statements & The value of Accounts Receivable Turnover Ratio \\
\midrule
Analyze Financial Statement-Current Assets Turnover Ratio & \{3,3,1\} & Based on the three financial statements, calculate the Current Assets Turnover Ratio & All Financial Statements & The value of Current Assets Turnover Ratio \\
\midrule
Analyze Financial Statement-Total Asset Turnover Ratio & \{3,3,1\} & Based on the three financial statements, calculate the Total Asset Turnover Ratio & All Financial Statements & The value of Total Asset Turnover Ratio  \\
\midrule
Analyze Financial Statement-Cash Flow to Debt Ratio & \{2,3,1\} & Based on the three financial statements, calculate the Cash Flow to Debt Ratio & All Financial Statements & The value of Cash Flow to Debt Ratio \\
\midrule
Analyze Financial Statement-Operating Cash Flow Ratio & \{2,3,1\} & Based on the three financial statements, calculate the Operating Cash Flow Ratio & All Financial Statements & The value of Operating Cash Flow Ratio \\

\midrule
\bottomrule
\end{tabular}}
\end{table}

\begin{table}[htbp]
\raggedright
\caption{Task Information Table - Multi-Capability in Consulting}
\resizebox{\textwidth}{!}{
\begin{tabular}{@{}p{6cm}p{1.5cm}p{6cm}p{4cm}p{4cm}@{}}

\midrule
Analyze Financial Statement-Current Ratio \& Inventory Turnover Ratio & \{5,3,2\} & Based on the three financial statements, calculate the Current Ratio and Inventory Turnover Ratio & All Financial Statements & The value of Current Ratio and Inventory Turnover Ratio \\
\midrule
Analyze Financial Statement-Gross Profit Margin \& Operating Cash Flow Ratio & \{4,3,2\} & Based on the three financial statements, calculate the Gross Profit Margin and Operating Cash Flow Ratio & All Financial Statements & The value of Gross Profit Margin and Operating Cash Flow Ratio \\
\midrule
Analyze Financial Statement-FCF \& Current Assets Turnover Ratio & \{5,3,2\} & Based on the three financial statements, calculate the FCF and Current Assets Turnover Ratio & All Financial Statements & The value of FCF and Current Assets Turnover Ratio  \\
\midrule
Analyze Financial Statement-Quick Ratio \& Net Profit Margin & \{8,3,2\} & Based on the three financial statements, calculate the Quick Ratio and Net Profit Margin & All Financial Statements & The value of Quick Ratio and Net Profit Margin \\
\midrule
Analyze Financial Statement-Gross Profit Margin \& Current Liabilities Ratio & \{4,3,2\} & Based on the three financial statements, calculate the Gross Profit Margin and Current Liabilities Ratio & All Financial Statements & The value of Gross Profit Margin and Current Liabilities Ratio \\
\midrule
Analyze Financial Statement-Debt to Asset Ratio \& Net Cash Ratio & \{4,3,2\} & Based on the three financial statements, calculate the Debt to Asset Ratio and Net Cash Ratio & All Financial Statements & The value of Debt to Asset Ratio and Net Cash Ratio \\
\midrule
Analyze Financial Statement-Debt to Equity Ratio \& Net Profit Margin \& Operating Cash Flow to Current Liabilities Ratio & \{6,3,3\} & Based on the three financial statements, calculate the Debt to Equity Ratio, Net Profit Margin and Operating Cash Flow to Current Liabilities Ratio & All Financial Statements & The value of Debt to Equity Ratio, Net Profit Margin and Operating Cash Flow to Current Liabilities Ratio \\
\midrule
Analyze Financial Statement-ROE \& Debt to Asset Ratio \& Gross Profit Margin & \{7,3,3\} & Based on the three financial statements, calculate the ROE, Debt to Asset Ratio and Gross Profit Margin & All Financial Statements & The value of ROE, Debt to Asset Ratio and Gross Profit Margin \\
\midrule
Analyze Financial Statement-Net Cash Ratio \& Turnover Ratio \& Quick Ratio & \{11,3,3\} & Based on the three financial statements, calculate the Net Cash Ratio, Turnover Ratio and Quick Ratio & All Financial Statements & The value of Net Cash Ratio, Turnover Ratio and Quick Ratio \\
\midrule
Analyze Financial Statement-Debt to Asset Ratio \& Gross Profit Margin \& Operating Cash Flow Ratio & \{6,3,3\} & Based on the three financial statements, calculate the Debt to Asset Ratio, Gross Profit Margin and Operating Cash Flow Ratio & All Financial Statements & The value of Debt to Asset Ratio, Gross Profit Margin and Operating Cash Flow Ratio \\
\midrule
Analyze Financial Statement-Debt to Equity Ratio \& Net Profit Margin \& ROA \& Accounts Receivable Turnover Ratio & \{10,3,4\} & Based on the three financial statements, calculate the Debt to Equity Ratio, Net Profit Margin, ROA and Accounts Receivable Turnover Ratio & All Financial Statements & The value of Debt to Equity Ratio, Net Profit Margin, ROA and Accounts Receivable Turnover Ratio\\
\midrule
Analyze Financial Statement-Current Ratio \& Quick Ratio \& Debt to Asset Ratio \& Debt to Equity Ratio \& Cash Flow to Debt Ratio & \{14,3,5\} & Based on the three financial statements, calculate the Current Ratio, Quick Ratio, Debt to Asset Ratio, Debt to Equity Ratio, Cash Flow to Debt Ratio & All Financial Statements & The value of Current Ratio, Quick Ratio, Debt to Asset Ratio, Debt to Equity Ratio, Cash Flow to Debt Ratio \\
\midrule
Analyze Financial Statement-Accounts Receivable Turnover Ratio \& Operating Cash Flow to Current Liabilities Ratio \& Operating Cash Flow Ratio \& Total Asset Turnover Ratio \& Debt to Equity Ratio & \{12,3,5\} & Based on the three financial statements, calculate the Accounts Receivable Turnover Ratio, Operating Cash Flow to Current Liabilities Ratio, Operating Cash Flow Ratio, Total Asset Turnover Ratio and Debt to Equity Ratio & All Financial Statements & The value of Accounts Receivable Turnover Ratio, Operating Cash Flow to Current Liabilities Ratio, Operating Cash Flow Ratio, Total Asset Turnover Ratio and Debt to Equity Ratio \\
\midrule
Analyze Financial Statement-FCF \& ROA \& ROE \& Net Cash Ratio \& Net Profit Margin \& Gross Profit Margin & \{14,3,6\} & Based on the three financial statements, calculate the FCF, ROA, ROE, Net Cash Ratio, Net Profit Margin and Gross Profit Margin & All Financial Statements & The value of FCF, ROA, ROE, Net Cash Ratio, Net Profit Margin and Gross Profit Margin \\
\midrule
Analyze Financial Statement-Operating Cash Flow Ratio \& Cash Flow to Debt Ratio \& Inventory Turnover Ratio \& Debt to Equity Ratio \& Quick Ratio \& Current Ratio & \{17,3,6\} & Based on the three financial statements, calculate the Operating Cash Flow Ratio, Cash Flow to Debt Ratio, Inventory Turnover Ratio, Debt to Equity Ratio, Quick Ratio and Current Ratio & All Financial Statements & The value of Operating Cash Flow Ratio, Cash Flow to Debt Ratio, Inventory Turnover Ratio, Debt to Equity Ratio, Quick Ratio and Current Ratio \\
\midrule
Analyze Financial Statement-Operating Cash Flow to Current Liabilities Ratio \& Debt to Equity Ratio \& Total Asset Turnover Ratio \& Quick Ratio \& Operating Cash Flow Ratio \& ROE \& Accounts Receivable Turnover Ratio & \{21,3,7\} & Based on the three financial statements, calculate the Operating Cash Flow to Current Liabilities Ratio, Debt to Equity Ratio, Total Asset Turnover Ratio, Quick Ratio, Operating Cash Flow Ratio, ROE and Accounts Receivable Turnover Ratio & All Financial Statements & The value of Operating Cash Flow to Current Liabilities Ratio, Debt to Equity Ratio, Total Asset Turnover Ratio, Quick Ratio, Operating Cash Flow Ratio, ROE and Accounts Receivable Turnover Ratio \\
\midrule
Analyze Financial Statement-Current Ratio \& Gross Profit Margin \& Debt to Asset Ratio \& Net Profit Margin \& Cash to Current Debt Ratio \& FCF \& ROA & \{15,3,7\} & Based on the three financial statements, calculate the Current Ratio, Gross Profit Margin, Debt to Asset Ratio, Net Profit Margin, Cash to Current Debt Ratio, FCF and ROA & All Financial Statements & The value of Current Ratio, Gross Profit Margin, Debt to Asset Ratio, Net Profit Margin, Cash to Current Debt Ratio, FCF and ROA \\ 
\midrule
\bottomrule
\end{tabular}}
\end{table}

\clearpage

\section{Extended Experiment Result}
\begin{figure}[ht]
    \centering
    \includegraphics[width=1\textwidth]{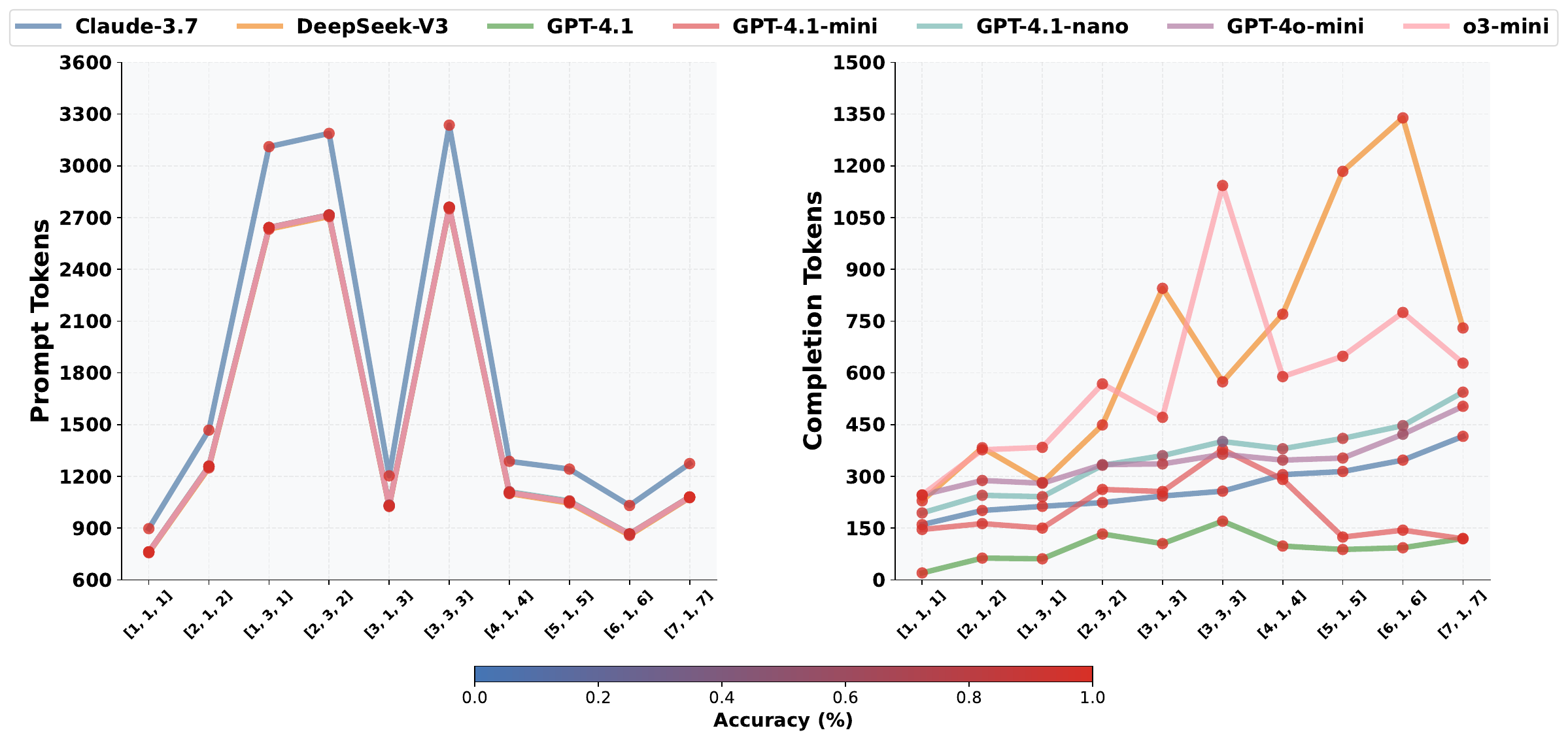}
    \caption{Financial Literacy prompt and completion Token Result}
    \label{fig:token_single}
\end{figure}

\begin{figure}[ht]
    \centering
    \includegraphics[width=1\textwidth]{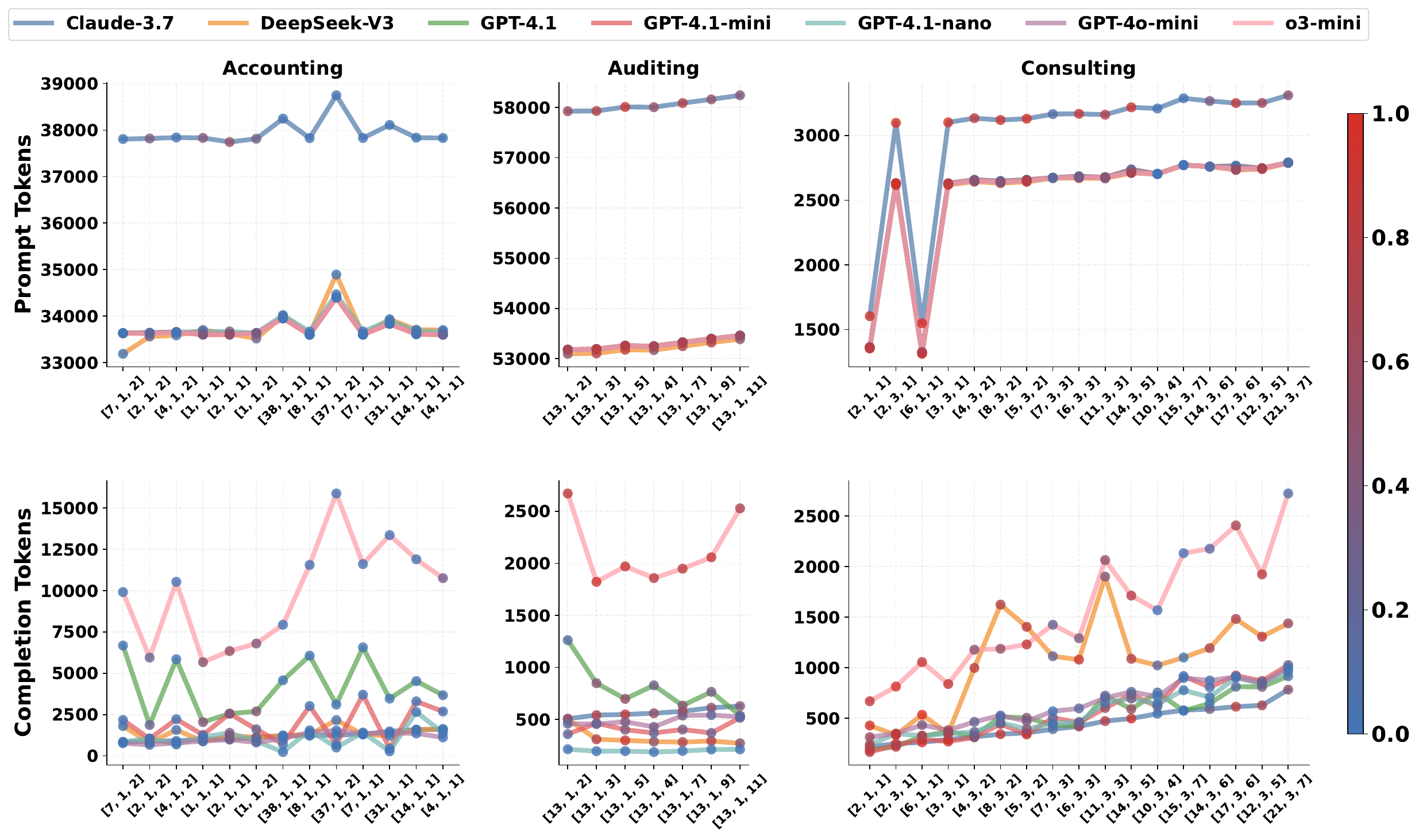}
    \caption{Main task prompt and completion result for model Comparison}
    \label{fig:complete_prompt_token}
\end{figure}
\clearpage

\begin{table}[htbp]
\centering 
\caption{Model accuracy for different operation
time}
\begin{tabular}{@{}p{4cm}ccc@{}} 
\toprule
\midrule
\textbf{Metrics} & \textbf{DeepSeek-V3} & \textbf{GPT-4.1} & \textbf{GPT-4o-mini} \\
\midrule
\midrule
\multicolumn{4}{l}{\textit{Transaction-400 Long Cycle}} \\
\midrule
Financial Literacy & 99.22\% ±0.89\% & 99.53\% ±0.73\% & 88.91\% ±1.94\% \\
Accounting & 15.04\% ±3.49\% & 20.27\% ±4.18\% & 3.81\% ±1.76\% \\
Auditing & 62.35\% ±6.54\% & 37.05\% ±11.56\% & 6.19\% ±3.68\% \\
Consulting & 78.10\% ±8.09\% & 61.52\% ±7.97\% & 39.52\% ±8.19\% \\
\midrule
\multicolumn{4}{l}{\textit{Transaction-200 Short Cycle}} \\
\midrule
Financial Literacy & 99.06\% ±0.88\% & 99.90\% ±0.40\% & 89.01\% ±2.19\% \\
Accounting & 21.33\% ±5.63\% & 32.93\% ±5.59\% & 7.76\% ±3.73\% \\
Auditing & 69.14\% ±5.53\% & 41.43\% ±10.31\% & 27.33\% ±7.22\% \\
Consulting & 80.00\% ±8.52\% & 56.38\% ±11.00\% & 37.43\% ±8.33\% \\
\bottomrule
\end{tabular}
\end{table}
\clearpage

\begin{table}[htbp]
\centering
\caption{Model Comparison across Different Company Types}
\begin{tabular}{@{}p{2.8cm}cccc@{}} 
\toprule
\midrule
\textbf{Model / Company Type} & \textbf{Financial Literacy} & \textbf{Accounting} & \textbf{Auditing} & \textbf{Consulting} \\
\midrule
\midrule
\multicolumn{5}{l}{\textit{Claude-3.7-Sonnet}} \\
\midrule
Type I & 99.38\% & 35.10\% & 73.14\% & 74.86\% \\
Type II & 99.74\% & 29.66\% & 72.86\% & 81.43\% \\
Type III & 99.48\% & 29.35\% & 71.43\% & 77.62\% \\
Type IV & 100.00\% & 26.61\% & 70.48\% & 83.81\% \\
Type V & 99.33\% & 28.53\% & 62.04\% & 82.45\% \\
\midrule
\multicolumn{5}{l}{\textit{DeepSeek-V3}} \\
\midrule
Type I & 98.75\% & 21.80\% & 68.57\% & 73.71\% \\
Type II & 99.22\% & 23.57\% & 68.57\% & 80.00\% \\
Type III & 99.48\% & 25.20\% & 71.43\% & 84.29\% \\
Type IV & 98.96\% & 19.31\% & 69.52\% & 76.67\% \\
Type V & 98.88\% & 17.27\% & 67.76\% & 83.67\% \\
\midrule
\multicolumn{5}{l}{\textit{GPT-4.1}} \\
\midrule
Type I & 99.69\% & 29.80\% & 36.00\% & 56.00\% \\
Type II & 99.74\% & 37.76\% & 45.24\% & 58.10\% \\
Type III & 100.00\% & 32.65\% & 45.71\% & 62.38\% \\
Type IV & 100.00\% & 31.63\% & 46.67\% & 46.67\% \\
Type V & 100.00\% & 32.36\% & 33.88\% & 58.37\% \\
\midrule
\multicolumn{5}{l}{\textit{GPT-4.1-mini}} \\
\midrule
Type I & 98.13\% & 21.22\% & 42.86\% & 63.43\% \\
Type II & 98.96\% & 22.79\% & 16.67\% & 70.00\% \\
Type III & 97.92\% & 21.09\% & 32.86\% & 66.67\% \\
Type IV & 97.92\% & 19.05\% & 16.19\% & 58.10\% \\
Type V & 97.77\% & 15.74\% & 36.73\% & 69.80\% \\
\midrule
\multicolumn{5}{l}{\textit{GPT-4.1-nano}} \\
\midrule
Type I & 86.25\% & 0.93\% & 0.57\% & 40.00\% \\
Type II & 83.33\% & 5.30\% & 0.48\% & 44.29\% \\
Type III & 83.59\% & 6.41\% & 0.00\% & 41.43\% \\
Type IV & 88.28\% & 4.90\% & 0.00\% & 38.62\% \\
Type V & 85.04\% & 2.56\% & 0.00\% & 45.31\% \\
\midrule
\multicolumn{5}{l}{\textit{GPT-4o-mini}} \\
\midrule
Type I & 89.38\% & 5.31\% & 29.14\% & 36.57\% \\
Type II & 89.06\% & 10.54\% & 23.33\% & 40.00\% \\
Type III & 89.06\% & 11.56\% & 27.62\% & 38.10\% \\
Type IV & 88.02\% & 4.76\% & 26.67\% & 33.81\% \\
Type V & 89.51\% & 6.41\% & 29.80\% & 38.37\% \\
\midrule
\multicolumn{5}{l}{\textit{o3-mini}} \\
\midrule
Type I & 100.00\% & 37.55\% & 86.29\% & 64.00\% \\
Type II & 100.00\% & 32.31\% & 85.71\% & 67.62\% \\
Type III & 100.00\% & 29.01\% & 83.81\% & 72.38\% \\
Type IV & 100.00\% & 38.10\% & 82.86\% & 55.71\% \\
Type V & 100.00\% & 36.55\% & 85.71\% & 71.43\% \\
\bottomrule
\end{tabular}
\end{table}

\clearpage
\begin{table}[htbp]
\centering
\caption{Claude-3.7-Sonnet Model Accuracy, Complete Completion  and Prompt Token Result}
\resizebox{\textwidth}{!}{
\begin{tabular}{@{}p{3cm}p{1cm}p{2cm}p{2cm}p{3cm}@{}} 
\toprule
\midrule
\textbf{Task Type} & \textbf{\{$\alpha$,$\beta$,$\gamma$\}} & \textbf{Percentage (\%)} & \textbf{Completion Token} & \textbf{Prompt Token} \\
\midrule
\midrule
Financial Literacy & [1,1,1] & 100\% & 160 & 897 \\
& [1,3,1] & 100\% & 213 & 3,111 \\
& [2,1,2] & 99\% & 201 & 1,468 \\
& [2,3,2] & 100\% & 224 & 3,188 \\
& [3,1,3] & 100\% & 243 & 1,203 \\
& [3,3,3] & 100\% & 257 & 3,236 \\
& [4,1,4] & 97.78\% & 305 & 1,287 \\
& [5,1,5] & 100\% & 314 & 1,242 \\
& [6,1,6] & 100\% & 347 & 1,031 \\
& [7,1,7] & 100\% & 416 & 1,274 \\
\midrule
Accounting & [1,1,1] & 41.9\% & 1,015 & 37,832 \\
& [1,1,2] & 30.45\% & 1,102 & 37,814 \\
& [14,1,1] & 0\% & 1,588 & 37,837 \\
& [2,1,1] & 52\% & 1,018 & 37,744 \\
& [2,1,2] & 39.33\% & 892 & 37,819 \\
& [31,1,1] & 0\% & 1,483 & 38,110 \\
& [37,1,2] & 0\% & 1,262 & 38,748 \\
& [38,1,1] & 0\% & 1,213 & 38,246 \\
& [4,1,1] & 0\% & 1,595 & 37,830 \\
& [4,1,2] & 0\% & 906 & 37,841 \\
& [7,1,1] & 0\% & 1,310 & 37,830 \\
& [7,1,2] & 0\% & 822 & 37,808 \\
& [8,1,1] & 0\% & 1,220 & 37,827 \\
\midrule
Auditing & [13,1,11] & 42.22\% & 629 & 58,243 \\
& [13,1,2] & 61.67\% & 508 & 57,927 \\
& [13,1,3] & 76.33\% & 542 & 57,931 \\
& [13,1,4] & 52.5\% & 560 & 58,006 \\
& [13,1,5] & 89.05\% & 549 & 58,011 \\
& [13,1,7] & 76\% & 579 & 58,089 \\
& [13,1,9] & 52.5\% & 613 & 58,162 \\
\midrule
Consulting & [10,3,4] & 0\% & 546 & 3,209 \\
& [11,3,3] & 76.67\% & 473 & 3,161 \\
& [12,3,5] & 66.67\% & 628 & 3,252 \\
& [14,3,5] & 93.33\% & 498 & 3,218 \\
& [14,3,6] & 36.67\% & 592 & 3,267 \\
& [15,3,7] & 0\% & 580 & 3,288 \\
& [17,3,6] & 83.33\% & 615 & 3,251 \\
& [2,1,1] & 91.11\% & 218 & 1,603 \\
& [2,3,1] & 98.33\% & 250 & 3,098 \\
& [21,3,7] & 46.67\% & 784 & 3,311 \\
& [3,3,1] & 95\% & 291 & 3,102 \\
& [4,3,2] & 83.33\% & 317 & 3,134 \\
& [5,3,2] & 98.33\% & 356 & 3,130 \\
& [6,1,1] & 100\% & 263 & 1,548 \\
& [6,3,3] & 86.67\% & 420 & 3,168 \\
& [7,3,3] & 10\% & 393 & 3,166 \\
& [8,3,2] & 90\% & 343 & 3,120 \\
\bottomrule
\end{tabular}
}
\end{table}

\clearpage

\begin{table}[htbp]
\centering
\caption{Deepseek-V3 Model Accuracy, Complete Completion  and Prompt Token Result}
\resizebox{\textwidth}{!}{
\begin{tabular}{@{}p{3cm}p{1cm}p{2cm}p{2cm}p{3cm}@{}} 
\toprule
\midrule
\textbf{Task Type} & \textbf{\{$\alpha$,$\beta$,$\gamma$\}} & \textbf{Percentage (\%)} & \textbf{Completion Token} & \textbf{Prompt Token} \\
\midrule
\midrule
Financial Literacy & [1,1,1] & 100\% & 229 & 757 \\
& [1,3,1] & 100\% & 282 & 2,633 \\
& [2,1,2] & 97\% & 383 & 1,249 \\
& [2,3,2] & 100\% & 449 & 2,706 \\
& [3,1,3] & 100\% & 845 & 1,025 \\
& [3,3,3] & 100\% & 574 & 2,751 \\
& [4,1,4] & 100\% & 770 & 1,100 \\
& [5,1,5] & 100\% & 1,184 & 1,045 \\
& [6,1,6] & 100\% & 1,339 & 858 \\
& [7,1,7] & 100\% & 730 & 1,076 \\
\midrule
Accounting & [1,1,1] & 23.39\% & 878 & 33,701 \\
& [1,1,2] & 25.87\% & 1,123 & 33,519 \\
& [14,1,1] & 0\% & 1,586 & 33,708 \\
& [2,1,1] & 52.94\% & 1,221 & 33,618 \\
& [2,1,2] & 30.87\% & 792 & 33,562 \\
& [31,1,1] & 0\% & 1,214 & 33,933 \\
& [37,1,2] & 0\% & 2,164 & 34,894 \\
& [38,1,1] & 0\% & 1,237 & 34,003 \\
& [4,1,1] & 1.85\% & 1,625 & 33,699 \\
& [4,1,2] & 0\% & 1,572 & 33,586 \\
& [7,1,1] & 0\% & 1,292 & 33,604 \\
& [7,1,2] & 0\% & 1,805 & 33,189 \\
& [8,1,1] & 0\% & 1,285 & 33,615 \\
\midrule
Auditing & [13,1,11] & 45.56\% & 272 & 53,387 \\
& [13,1,2] & 50\% & 503 & 53,096 \\
& [13,1,3] & 72.33\% & 310 & 53,105 \\
& [13,1,4] & 49.17\% & 288 & 53,170 \\
& [13,1,5] & 87.62\% & 300 & 53,177 \\
& [13,1,7] & 70.67\% & 283 & 53,249 \\
& [13,1,9] & 74.17\% & 294 & 53,321 \\
\midrule
Consulting & [10,3,4] & 16.67\% & 1,023 & 2,701 \\
& [11,3,3] & 43.33\% & 1,900 & 2,668 \\
& [12,3,5] & 80\% & 1,307 & 2,740 \\
& [14,3,5] & 73.33\% & 1,088 & 2,710 \\
& [14,3,6] & 66.67\% & 1,194 & 2,760 \\
& [15,3,7] & 0\% & 1,101 & 2,771 \\
& [17,3,6] & 70\% & 1,482 & 2,733 \\
& [2,1,1] & 92.59\% & 429 & 1,353 \\
& [2,3,1] & 96.67\% & 335 & 2,617 \\
& [21,3,7] & 56.67\% & 1,438 & 2,787 \\
& [3,3,1] & 96.11\% & 350 & 2,619 \\
& [4,3,2] & 81.11\% & 998 & 2,643 \\
& [5,3,2] & 83.33\% & 1,404 & 2,642 \\
& [6,1,1] & 100\% & 535 & 1,312 \\
& [6,3,3] & 91.67\% & 1,080 & 2,671 \\
& [7,3,3] & 23.33\% & 1,114 & 2,670 \\
& [8,3,2] & 73.33\% & 1,624 & 2,632 \\
\bottomrule
\end{tabular}
}
\end{table}
\clearpage

\begin{table}[htbp]
\raggedright  
\caption{GPT-4.1 Model Accuracy, Complete Completion  and Prompt Token Result}
\resizebox{\textwidth}{!}{
\begin{tabular}{@{}p{3cm}p{1cm}p{2cm}p{2cm}p{3cm}@{}} 

\toprule
\midrule
\textbf{Task Type} & \textbf{\{$\alpha$,$\beta$,$\gamma$\}} & \textbf{Percentage (\%)} & \textbf{Completion Token} & \textbf{Prompt Token} \\
\midrule
\midrule
Financial Literacy & [1,1,1] & 100\% & 20 & 762 \\
& [1,3,1] & 100\% & 61 & 2,642 \\
& [2,1,2] & 99.67\% & 63 & 1,257 \\
& [2,3,2] & 100\% & 133 & 2,715 \\
& [3,1,3] & 100\% & 105 & 1,030 \\
& [3,3,3] & 100\% & 170 & 2,760 \\
& [4,1,4] & 100\% & 98 & 1,106 \\
& [5,1,5] & 100\% & 88 & 1,053 \\
& [6,1,6] & 100\% & 93 & 866 \\
& [7,1,7] & 100\% & 120 & 1,080 \\
\midrule
Accounting & [1,1,1] & 43.51\% & 2,044 & 33,607 \\
& [1,1,2] & 45\% & 2,694 & 33,638 \\
& [14,1,1] & 0\% & 4,525 & 33,618 \\
& [2,1,1] & 53.33\% & 2,556 & 33,613 \\
& [2,1,2] & 25.33\% & 1,872 & 33,642 \\
& [31,1,1] & 0\% & 3,464 & 33,837 \\
& [37,1,2] & 0\% & 3,106 & 34,403 \\
& [38,1,1] & 0\% & 4,575 & 33,957 \\
& [4,1,1] & 6.67\% & 3,674 & 33,604 \\
& [4,1,2] & 0\% & 5,847 & 33,659 \\
& [7,1,1] & 0\% & 6,565 & 33,606 \\
& [7,1,2] & 0\% & 6,678 & 33,635 \\
& [8,1,1] & 0\% & 6,065 & 33,602 \\
\midrule
Auditing & [13,1,11] & 34.44\% & 538 & 53,459 \\
& [13,1,2] & 13.33\% & 1,262 & 53,180 \\
& [13,1,3] & 45.67\% & 849 & 53,191 \\
& [13,1,4] & 30\% & 828 & 53,248 \\
& [13,1,5] & 53.81\% & 698 & 53,259 \\
& [13,1,7] & 46.67\% & 634 & 53,326 \\
& [13,1,9] & 33.33\% & 766 & 53,396 \\
\midrule
Consulting & [10,3,4] & 6.67\% & 755 & 2,704 \\
& [11,3,3] & 23.33\% & 724 & 2,677 \\
& [12,3,5] & 36.67\% & 811 & 2,747 \\
& [14,3,5] & 56.67\% & 592 & 2,737 \\
& [14,3,6] & 0\% & 647 & 2,760 \\
& [15,3,7] & 0\% & 574 & 2,772 \\
& [17,3,6] & 13.33\% & 812 & 2,766 \\
& [2,1,1] & 82.96\% & 188 & 1,364 \\
& [2,3,1] & 95\% & 219 & 2,632 \\
& [21,3,7] & 6.67\% & 916 & 2,791 \\
& [3,3,1] & 75\% & 366 & 2,630 \\
& [4,3,2] & 37.78\% & 327 & 2,657 \\
& [5,3,2] & 46.67\% & 505 & 2,658 \\
& [6,1,1] & 90\% & 325 & 1,325 \\
& [6,3,3] & 46.67\% & 424 & 2,685 \\
& [7,3,3] & 10\% & 435 & 2,674 \\
& [8,3,2] & 43.33\% & 514 & 2,648 \\
\bottomrule
\end{tabular}
}
\end{table}
\clearpage
\begin{table}[htbp]
\raggedright  
\caption{GPT-4.1-mini Accuracy, Complete Completion  and Prompt Token Result}
\resizebox{\textwidth}{!}{
\begin{tabular}{@{}p{3cm}p{1cm}p{2cm}p{2cm}p{3cm}@{}} 

\toprule
\midrule
\textbf{Task Type} & \textbf{\{$\alpha$,$\beta$,$\gamma$\}} & \textbf{Percentage (\%)} & \textbf{Completion Token} & \textbf{Prompt Token} \\
\midrule
\midrule
Financial Literacy & [1,1,1] & 97.96\% & 146 & 762 \\
& [1,3,1] & 100\% & 150 & 2,642 \\
& [2,1,2] & 95.83\% & 163 & 1,257 \\
& [2,3,2] & 100\% & 262 & 2,715 \\
& [3,1,3] & 100\% & 256 & 1,030 \\
& [3,3,3] & 100\% & 378 & 2,760 \\
& [4,1,4] & 100\% & 291 & 1,106 \\
& [5,1,5] & 100\% & 124 & 1,053 \\
& [6,1,6] & 100\% & 144 & 866 \\
& [7,1,7] & 100\% & 119 & 1,080 \\
\midrule
Accounting & [1,1,1] & 24.91\% & 1,271 & 33,603 \\
& [1,1,2] & 25.28\% & 1,608 & 33,634 \\
& [14,1,1] & 0\% & 3,298 & 33,611 \\
& [2,1,1] & 40\% & 2,555 & 33,602 \\
& [2,1,2] & 22\% & 1,030 & 33,638 \\
& [31,1,1] & 0\% & 439 & 33,834 \\
& [37,1,2] & 0\% & 681 & 34,400 \\
& [38,1,1] & 0\% & 715 & 33,952 \\
& [4,1,1] & 1.67\% & 2,685 & 33,601 \\
& [4,1,2] & 0\% & 2,219 & 33,651 \\
& [7,1,1] & 0\% & 3,705 & 33,602 \\
& [7,1,2] & 0\% & 2,172 & 33,632 \\
& [8,1,1] & 0\% & 3,017 & 33,599 \\
\midrule
Auditing & [13,1,11] & 22.22\% & 517 & 53,459 \\
& [13,1,2] & 16.67\% & 360 & 53,180 \\
& [13,1,3] & 29.67\% & 465 & 53,191 \\
& [13,1,4] & 19.17\% & 371 & 53,248 \\
& [13,1,5] & 38.10\% & 402 & 53,259 \\
& [13,1,7] & 33.33\% & 404 & 53,326 \\
& [13,1,9] & 25.83\% & 371 & 53,396 \\
\midrule
Consulting & [10,3,4] & 23.33\% & 617 & 2,704 \\
& [11,3,3] & 56.67\% & 602 & 2,677 \\
& [12,3,5] & 46.67\% & 869 & 2,747 \\
& [14,3,5] & 76.67\% & 735 & 2,715 \\
& [14,3,6] & 3.33\% & 807 & 2,760 \\
& [15,3,7] & 0\% & 918 & 2,772 \\
& [17,3,6] & 63.33\% & 923 & 2,740 \\
& [2,1,1] & 80\% & 167 & 1,360 \\
& [2,3,1] & 66.67\% & 234 & 2,626 \\
& [21,3,7] & 23.33\% & 1,027 & 2,791 \\
& [3,3,1] & 91.67\% & 270 & 2,627 \\
& [4,3,2] & 41.11\% & 312 & 2,651 \\
& [5,3,2] & 85\% & 340 & 2,650 \\
& [6,1,1] & 93.33\% & 293 & 1,321 \\
& [6,3,3] & 63.33\% & 458 & 2,677 \\
& [7,3,3] & 23.33\% & 505 & 2,674 \\
& [8,3,2] & 70\% & 444 & 2,641 \\
\bottomrule
\end{tabular}
}
\end{table}
\clearpage

\begin{table}[htbp]
\raggedright  
\caption{GPT-4.1-nano Model Accuracy, Complete Completion  and Prompt Token Result}
\resizebox{\textwidth}{!}{
\begin{tabular}{@{}p{3cm}p{1cm}p{2cm}p{2cm}p{3cm}@{}} 
\toprule
\midrule
\textbf{Task Type} & \textbf{\{$\alpha$,$\beta$,$\gamma$\}} & \textbf{Percentage (\%)} & \textbf{Completion Token} & \textbf{Prompt Token} \\
\midrule
\midrule
Financial Literacy & [1,1,1] & 93.22\% & 194 & 761 \\
& [1,3,1] & 93.53\% & 241 & 2,642 \\
& [2,1,2] & 88.42\% & 245 & 1,260 \\
& [2,3,2] & 71.26\% & 332 & 2,715 \\
& [3,1,3] & 64.96\% & 360 & 1,034 \\
& [3,3,3] & 40.23\% & 401 & 2,759 \\
& [4,1,4] & 78.41\% & 380 & 1,111 \\
& [5,1,5] & 87.50\% & 410 & 1,058 \\
& [6,1,6] & 68.97\% & 447 & 866 \\
& [7,1,7] & 100\% & 544 & 1,080 \\
\midrule
Accounting & [1,1,1] & 2.63\% & 1,146 & 33,672 \\
& [1,1,2] & 6.11\% & 892 & 33,634 \\
& [14,1,1] & 0\% & 2,658 & 33,680 \\
& [2,1,1] & 36.36\% & 1,393 & 33,671 \\
& [2,1,2] & 0\% & 1,057 & 33,638 \\
& [31,1,1] & 0\% & 277 & 33,903 \\
& [37,1,2] & 0\% & 506 & 34,469 \\
& [38,1,1] & 0\% & 233 & 34,021 \\
& [4,1,1] & 0\% & 1,392 & 33,670 \\
& [4,1,2] & 0\% & 688 & 33,651 \\
& [7,1,1] & 0\% & 1,412 & 33,671 \\
& [7,1,2] & 0\% & 852 & 33,632 \\
& [8,1,1] & 0\% & 1,547 & 33,668 \\
\midrule
Auditing & [13,1,11] & 0\% & 213 & 53,459 \\
& [13,1,2] & 0\% & 215 & 53,180 \\
& [13,1,3] & 0\% & 196 & 53,191 \\
& [13,1,4] & 0\% & 188 & 53,248 \\
& [13,1,5] & 0.95\% & 197 & 53,259 \\
& [13,1,7] & 0\% & 198 & 53,326 \\
& [13,1,9] & 0\% & 212 & 53,396 \\
\midrule
Consulting & [10,3,4] & 3.45\% & 641 & 2,704 \\
& [11,3,3] & 6.90\% & 651 & 2,678 \\
& [12,3,5] & 24.14\% & 839 & 2,747 \\
& [14,3,5] & 31.03\% & 696 & 2,715 \\
& [14,3,6] & 3.45\% & 708 & 2,761 \\
& [15,3,7] & 0\% & 778 & 2,772 \\
& [17,3,6] & 13.79\% & 903 & 2,740 \\
& [2,1,1] & 64.07\% & 241 & 1,360 \\
& [2,3,1] & 72.41\% & 346 & 2,626 \\
& [21,3,7] & 3.45\% & 965 & 2,791 \\
& [3,3,1] & 64.04\% & 348 & 2,627 \\
& [4,3,2] & 24.14\% & 366 & 2,651 \\
& [5,3,2] & 44.83\% & 394 & 2,650 \\
& [6,1,1] & 63.33\% & 329 & 1,321 \\
& [6,3,3] & 17.24\% & 463 & 2,677 \\
& [7,3,3] & 3.45\% & 452 & 2,674 \\
& [8,3,2] & 10.34\% & 458 & 2,641 \\
\bottomrule
\end{tabular}
}
\end{table}
\clearpage
\begin{table}[htbp]
\raggedright  
\caption{GPT-4o-mini Model Accuracy, Complete Completion  and Prompt Token Result}
\resizebox{\textwidth}{!}{
\begin{tabular}{@{}p{3cm}p{1cm}p{2cm}p{2cm}p{3cm}@{}} 
\toprule
\midrule
\textbf{Task Type} & \textbf{\{$\alpha$,$\beta$,$\gamma$\}} & \textbf{Percentage (\%)} & \textbf{Completion Token} & \textbf{Prompt Token} \\
\midrule
\midrule
Financial Literacy & [1,1,1] & 96.48\% & 246 & 762 \\
& [1,3,1] & 87.50\% & 280 & 2,642 \\
& [2,1,2] & 89.67\% & 288 & 1,257 \\
& [2,3,2] & 67.78\% & 334 & 2,715 \\
& [3,1,3] & 80.00\% & 336 & 1,030 \\
& [3,3,3] & 91.11\% & 364 & 2,760 \\
& [4,1,4] & 78.89\% & 347 & 1,106 \\
& [5,1,5] & 91.11\% & 353 & 1,053 \\
& [6,1,6] & 60.00\% & 422 & 866 \\
& [7,1,7] & 100\% & 503 & 1,080 \\
\midrule
Accounting & [1,1,1] & 7.37\% & 890 & 33,607 \\
& [1,1,2] & 14.17\% & 808 & 33,638 \\
& [14,1,1] & 0\% & 1,357 & 33,618 \\
& [2,1,1] & 26.67\% & 984 & 33,613 \\
& [2,1,2] & 3.33\% & 665 & 33,642 \\
& [31,1,1] & 0\% & 1,384 & 33,837 \\
& [37,1,2] & 0\% & 1,537 & 34,403 \\
& [38,1,1] & 0\% & 1,053 & 33,957 \\
& [4,1,1] & 0\% & 1,119 & 33,604 \\
& [4,1,2] & 0\% & 788 & 33,659 \\
& [7,1,1] & 0\% & 1,346 & 33,606 \\
& [7,1,2] & 0\% & 767 & 33,635 \\
& [8,1,1] & 0\% & 1,428 & 33,602 \\
\midrule
Auditing & [13,1,11] & 14.44\% & 526 & 53,459 \\
& [13,1,2] & 20.00\% & 460 & 53,180 \\
& [13,1,3] & 24.00\% & 456 & 53,191 \\
& [13,1,4] & 33.33\% & 429 & 53,248 \\
& [13,1,5] & 36.67\% & 478 & 53,259 \\
& [13,1,7] & 31.33\% & 538 & 53,326 \\
& [13,1,9] & 21.67\% & 543 & 53,396 \\
\midrule
Consulting & [10,3,4] & 0\% & 714 & 2,704 \\
& [11,3,3] & 16.67\% & 702 & 2,677 \\
& [12,3,5] & 23.33\% & 849 & 2,747 \\
& [14,3,5] & 13.33\% & 762 & 2,738 \\
& [14,3,6] & 0\% & 874 & 2,760 \\
& [15,3,7] & 0\% & 898 & 2,772 \\
& [17,3,6] & 0\% & 909 & 2,767 \\
& [2,1,1] & 58.89\% & 314 & 1,364 \\
& [2,3,1] & 85.00\% & 349 & 2,632 \\
& [21,3,7] & 0\% & 999 & 2,791 \\
& [3,3,1] & 57.22\% & 383 & 2,631 \\
& [4,3,2] & 21.11\% & 465 & 2,658 \\
& [5,3,2] & 36.67\% & 472 & 2,658 \\
& [6,1,1] & 40.00\% & 434 & 1,325 \\
& [6,3,3] & 15.00\% & 598 & 2,685 \\
& [7,3,3] & 0\% & 570 & 2,674 \\
& [8,3,2] & 6.67\% & 528 & 2,649 \\
\bottomrule
\end{tabular}
}
\end{table}

\clearpage
\begin{table}[htbp]
\raggedright  
\caption{o3-mini Model Accuracy, Complete Completion and Prompt Token Result}
\resizebox{\textwidth}{!}{
\begin{tabular}{@{}p{3cm}p{1cm}p{2cm}p{2cm}p{3cm}@{}} 
\toprule
\midrule
\textbf{Task Type} & \textbf{\{$\alpha$,$\beta$,$\gamma$\}} & \textbf{Percentage (\%)} & \textbf{Completion Token} & \textbf{Prompt Token} \\
\midrule
\midrule
Financial Literacy & [1,1,1] & 100\% & 246 & 761 \\
& [1,3,1] & 100\% & 384 & 2,641 \\
& [2,1,2] & 100\% & 377 & 1,256 \\
& [2,3,2] & 100\% & 568 & 2,714 \\
& [3,1,3] & 100\% & 471 & 1,029 \\
& [3,3,3] & 100\% & 1,143 & 2,759 \\
& [4,1,4] & 100\% & 589 & 1,105 \\
& [5,1,5] & 100\% & 648 & 1,052 \\
& [6,1,6] & 100\% & 775 & 865 \\
& [7,1,7] & 100\% & 628 & 1,079 \\
\midrule
Accounting & [1,1,1] & 47.02\% & 5,673 & 33,602 \\
& [1,1,2] & 42.62\% & 6,798 & 33,628 \\
& [14,1,1] & 3.33\% & 11,896 & 33,610 \\
& [2,1,1] & 55\% & 6,345 & 33,601 \\
& [2,1,2] & 24.83\% & 5,943 & 33,621 \\
& [31,1,1] & 0\% & 13,361 & 33,833 \\
& [37,1,2] & 0\% & 15,881 & 34,399 \\
& [38,1,1] & 0\% & 7,936 & 33,951 \\
& [4,1,1] & 28.33\% & 10,756 & 33,600 \\
& [4,1,2] & 0\% & 10,533 & 33,650 \\
& [7,1,1] & 0\% & 11,607 & 33,601 \\
& [7,1,2] & 0\% & 9,914 & 33,631 \\
& [8,1,1] & 0\% & 11,553 & 33,598 \\
\midrule
Auditing & [13,1,11] & 68.89\% & 2,527 & 53,458 \\
& [13,1,2] & 81.67\% & 2,670 & 53,179 \\
& [13,1,3] & 93\% & 1,822 & 53,190 \\
& [13,1,4] & 78.33\% & 1,858 & 53,247 \\
& [13,1,5] & 87.62\% & 1,970 & 53,258 \\
& [13,1,7] & 81.33\% & 1,948 & 53,325 \\
& [13,1,9] & 84.17\% & 2,057 & 53,395 \\
\midrule
Consulting & [10,3,4] & 0\% & 1,568 & 2,703 \\
& [11,3,3] & 53.33\% & 2,064 & 2,676 \\
& [12,3,5] & 80\% & 1,923 & 2,746 \\
& [14,3,5] & 76.67\% & 1,713 & 2,714 \\
& [14,3,6] & 13.33\% & 2,176 & 2,759 \\
& [15,3,7] & 0\% & 2,132 & 2,771 \\
& [17,3,6] & 66.67\% & 2,406 & 2,739 \\
& [2,1,1] & 84.44\% & 669 & 1,359 \\
& [2,3,1] & 95\% & 814 & 2,625 \\
& [21,3,7] & 10\% & 2,722 & 2,790 \\
& [3,3,1] & 86.11\% & 839 & 2,626 \\
& [4,3,2] & 50\% & 1,178 & 2,650 \\
& [5,3,2] & 86.67\% & 1,230 & 2,649 \\
& [6,1,1] & 86.67\% & 1,056 & 1,320 \\
& [6,3,3] & 36.67\% & 1,291 & 2,676 \\
& [7,3,3] & 23.33\% & 1,425 & 2,673 \\
& [8,3,2] & 53.33\% & 1,187 & 2,640 \\
\bottomrule
\end{tabular}
}
\end{table}

\clearpage
\section{FinEval}
\label{app:fineval}
\subsection{Models}

\begin{table}[ht]
    \centering
    \caption{API-based LLMs considered in this paper via \emph{FinEval}.}
    \label{tab:models}
    \begin{tabular}{c|c|c|c}
        \toprule
         Type & Models & Version & Provider \\
        \midrule
        \multirow{7}{*}{Online} 
         & GPT-4o-mini & gpt-4o-mini-2024-07-18 &OpenAI \\
         & GPT-4.1 &gpt-4.1-2025-04-14 &OpenAI \\
         & GPT-4.1-mini &gpt-4.1-mini-2025-04-14 &OpenAI \\
         & GPT-4.1-nano &gpt-4.1-nano-2025-04-14 &OpenAI \\   
         & o3-mini & o3-mini-2025-01-31 & OpenAI \\
         & DeepSeek-V3 & deepseek-v3-250324 & Huoshan \\
         & Claude-3.7-Sonnet &claude-3-7-sonnet-20250219 & Anthropic \\
        \bottomrule
    \end{tabular}
\end{table}

\subsection{Prompt Template}
\label{template}

\begin{lstlisting}[language=Python, backgroundcolor=\color{lightgray!20}]
finmaster_template = """
# <task_name> Task Description:
<task_description>

# Examples:
<in_context_examples>
# Problem to Solve: 
{"problem": <task_to_solve>}

# Instruction:
Now please solve the above task. Reason step by step and present your answer in the "solution" field in the following json format:
```json
{"solution": "___" }
```
"""

example_and_solution = """{"problem": <example_problem>}
{"solution": <example_solution>}
"""
\end{lstlisting}

\clearpage

\clearpage

\clearpage

\section{Accuracy of Each Company Type for Specific Task }

\begin{figure}[H]
    \centering
    \includegraphics[width=\textwidth]{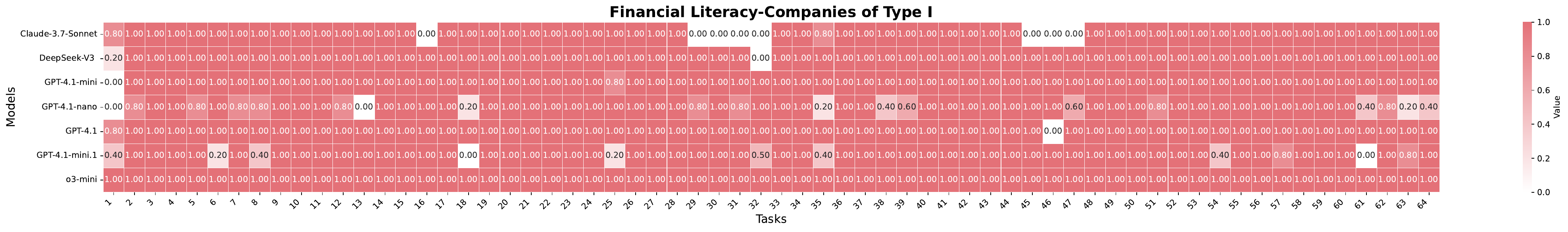}
    \caption{Accuracy of type I companies in financial literacy}
    \label{fig:AM1}
\end{figure}

\begin{figure}[H]
    \centering
    \includegraphics[width=\textwidth]{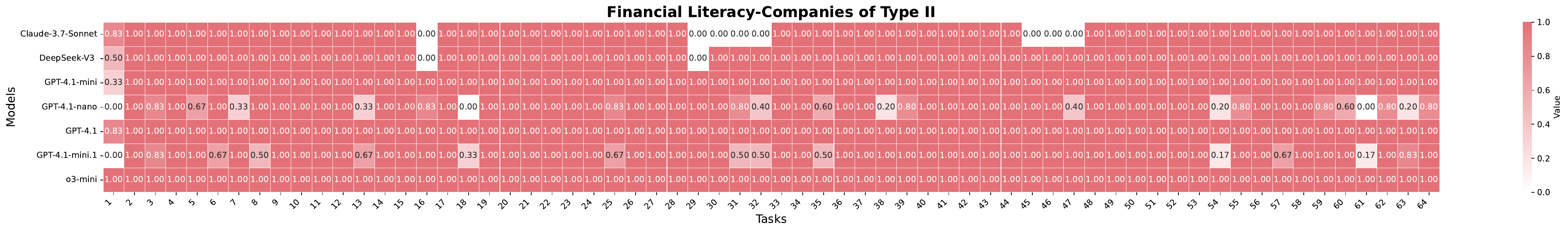}
    \caption{Accuracy of Type II Companies in Financial Literacy}
    \label{fig:AM2}
\end{figure}

\begin{figure}[H]
    \centering
    \includegraphics[width=\textwidth]{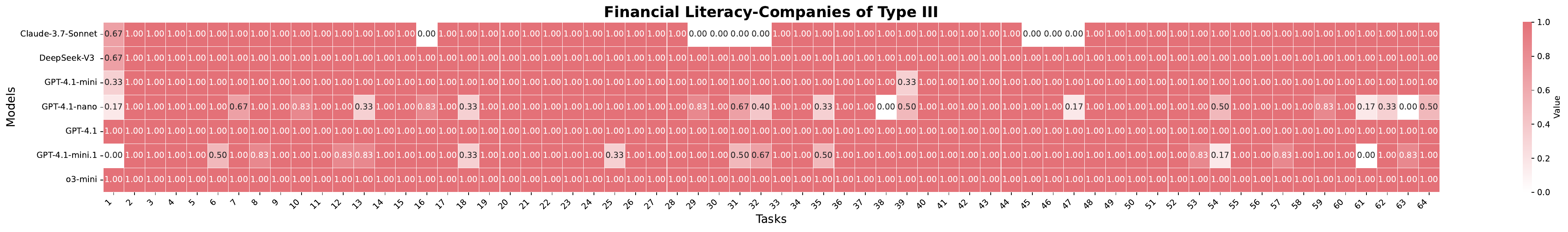}
    \caption{Accuracy of Type III Companies in Financial Literacy}
    \label{fig:AM3}
\end{figure}

\begin{figure}[H]
    \centering
    \includegraphics[width=\textwidth]{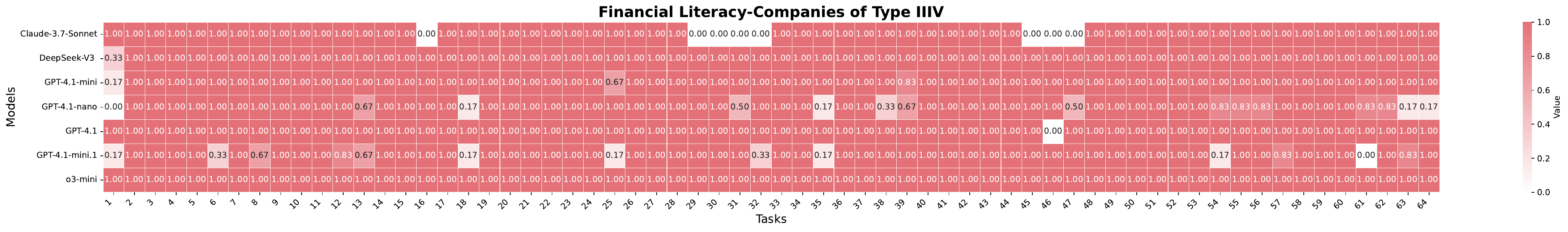}
    \caption{Accuracy of Type IIIV Companies in Financial Literacy}
    \label{fig:AM4}
\end{figure}

\begin{figure}[H]
    \centering
    \includegraphics[width=\textwidth]{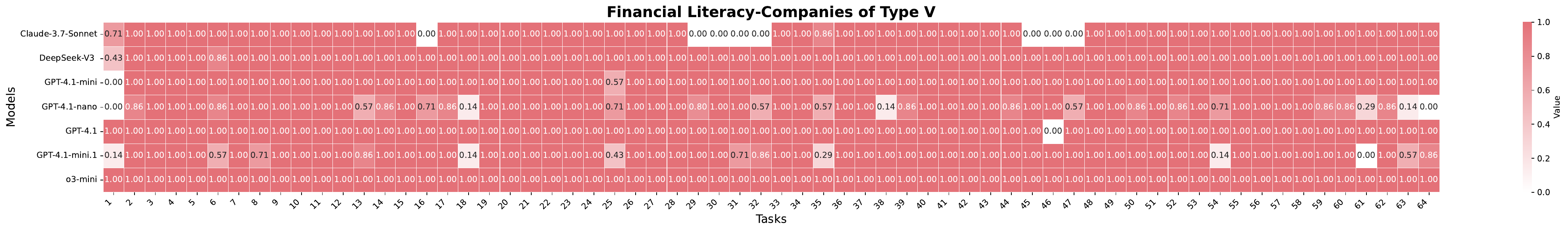}
    \caption{Accuracy of Type V Companies in Financial Literacy}
    \label{fig:AM5}
\end{figure}

\clearpage

\begin{figure}
    \centering
    \includegraphics[width=\textwidth]{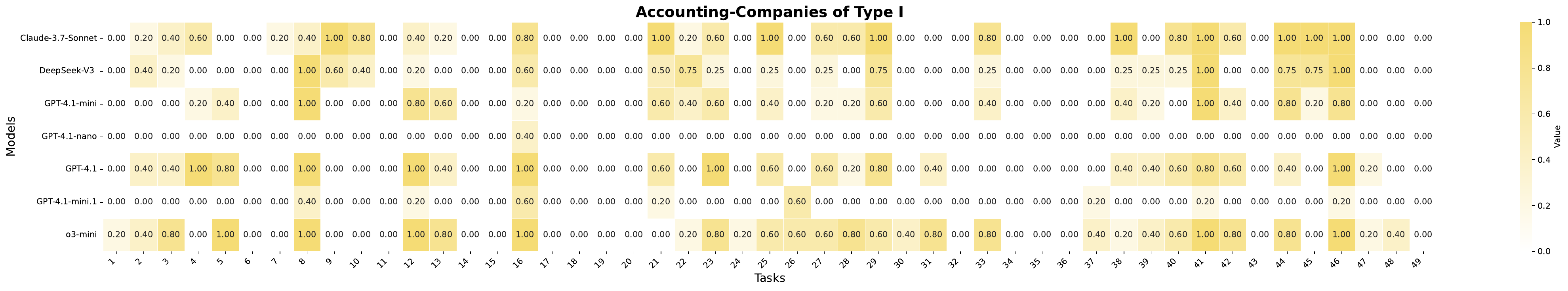}
    \caption{Accuracy of Type I Companies in Accounting}
    \label{fig:AM6}
\end{figure}

\begin{figure}
    \centering
    \includegraphics[width=\textwidth]{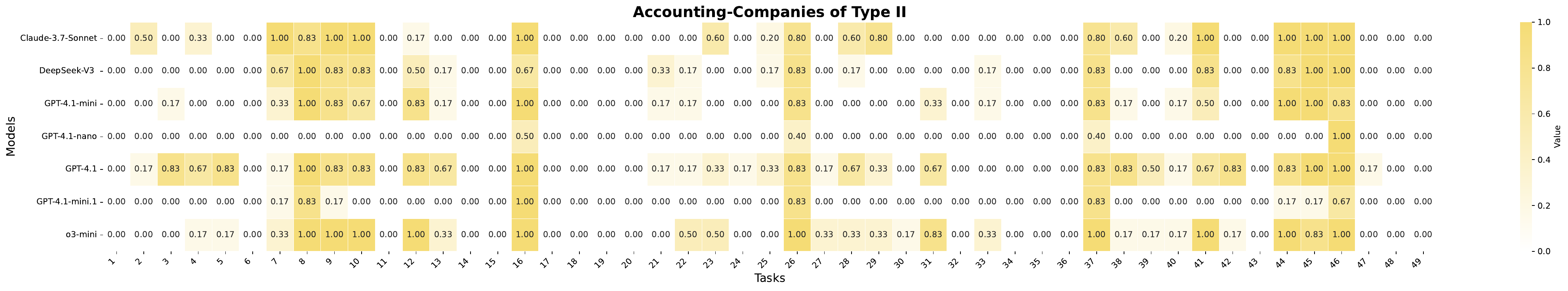}
    \caption{Accuracy of Type II Companies in Accounting}
    \label{fig:AM7}
\end{figure}

\begin{figure}
    \centering
    \includegraphics[width=\textwidth]{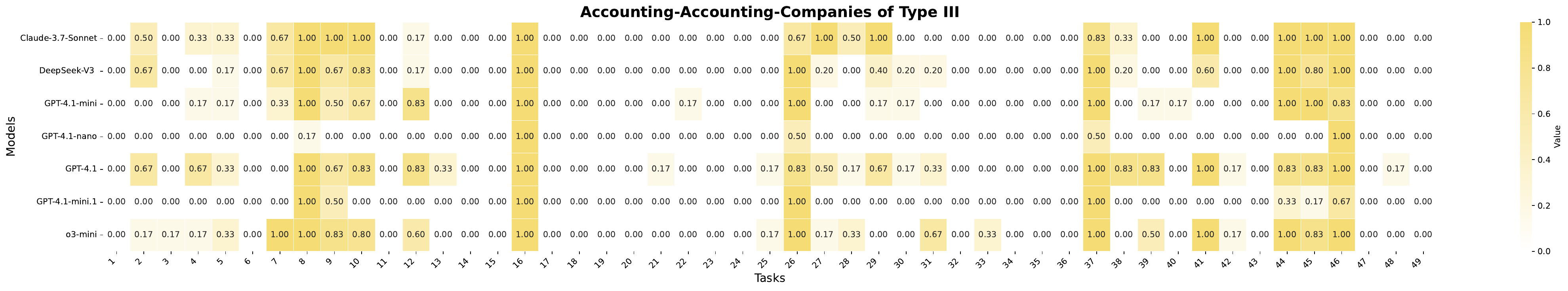}
    \caption{Accuracy of Type III Companies in Accounting}
    \label{fig:AM8}
\end{figure}

\begin{figure}
    \centering
    \includegraphics[width=\textwidth]{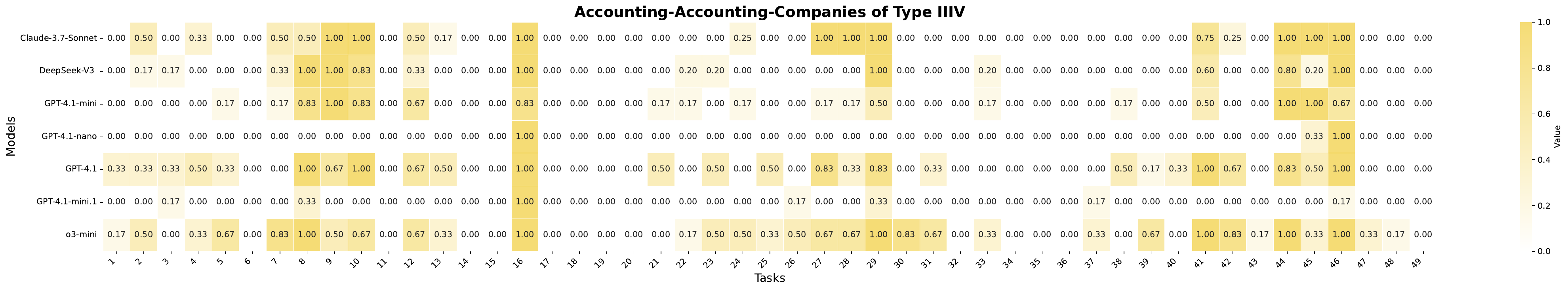}
    \caption{Accuracy of Type IIIV Companies in Accounting}
    \label{fig:AM9}
\end{figure}

\begin{figure}
    \centering
    \includegraphics[width=\textwidth]{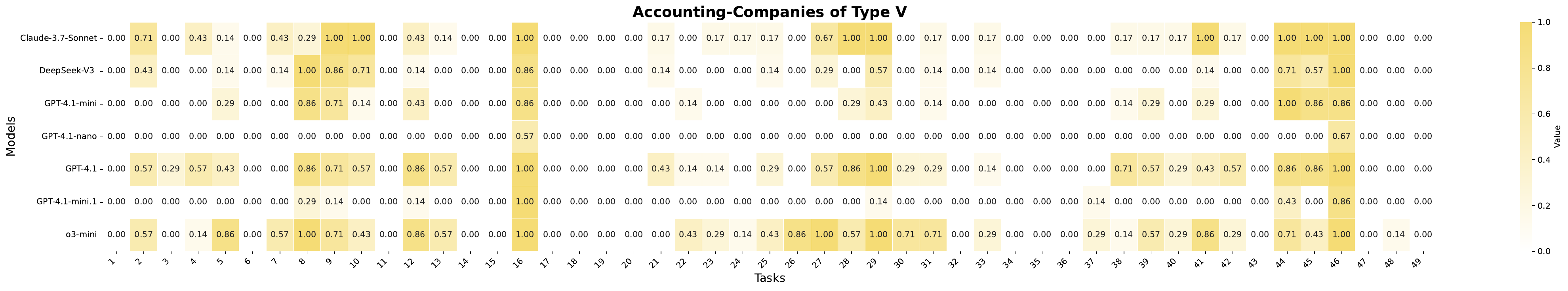}
    \caption{Accuracy of Type V Companies in Accounting}
    \label{fig:AM10}
\end{figure}

\clearpage

\begin{figure}
    \centering
    \includegraphics[width=\textwidth]{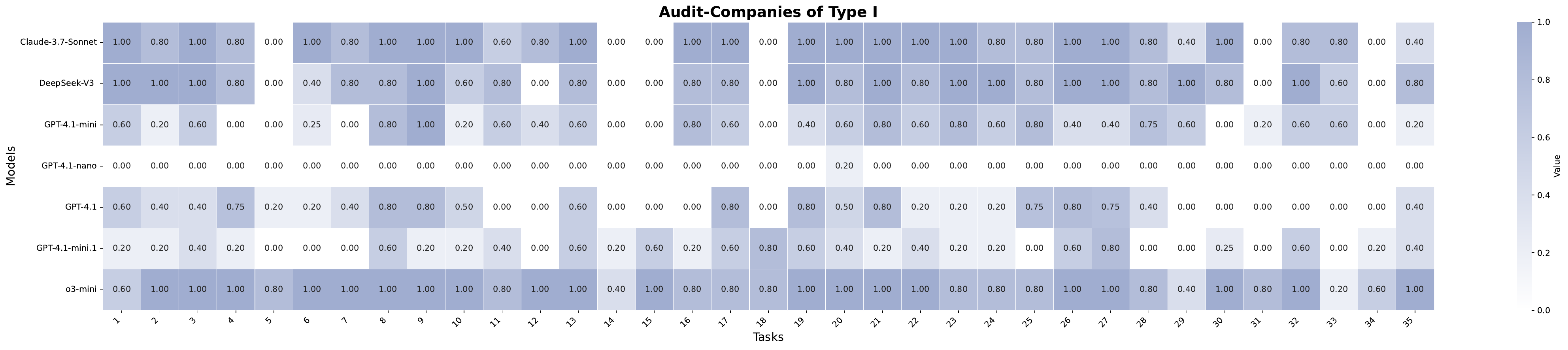}
    \caption{Accuracy of Type I Companies in Audit}
    \label{fig:AM11}
\end{figure}

\begin{figure}
    \centering
    \includegraphics[width=\textwidth]{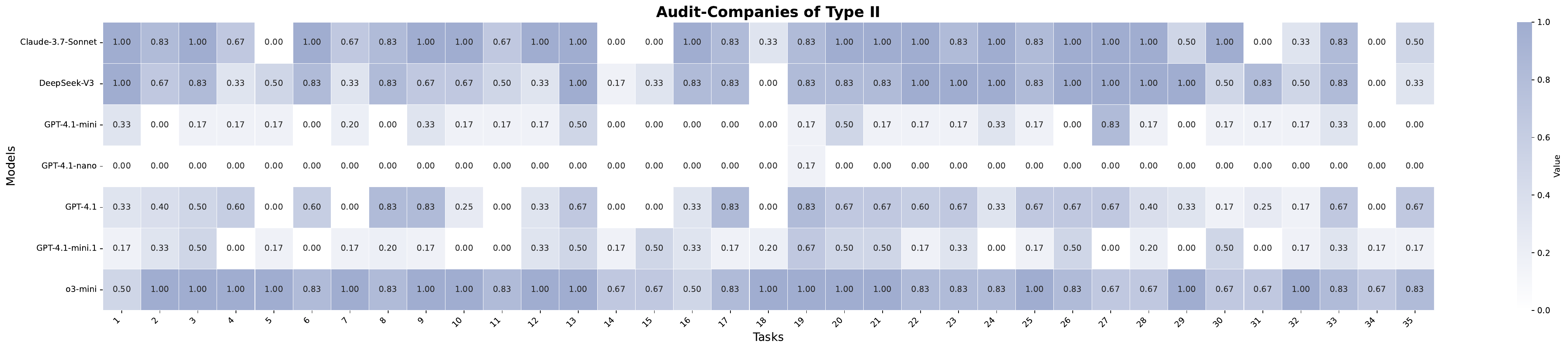}
    \caption{Accuracy of Type II Companies in Audit}
    \label{fig:AM12}
\end{figure}

\begin{figure}
    \centering
    \includegraphics[width=\textwidth]{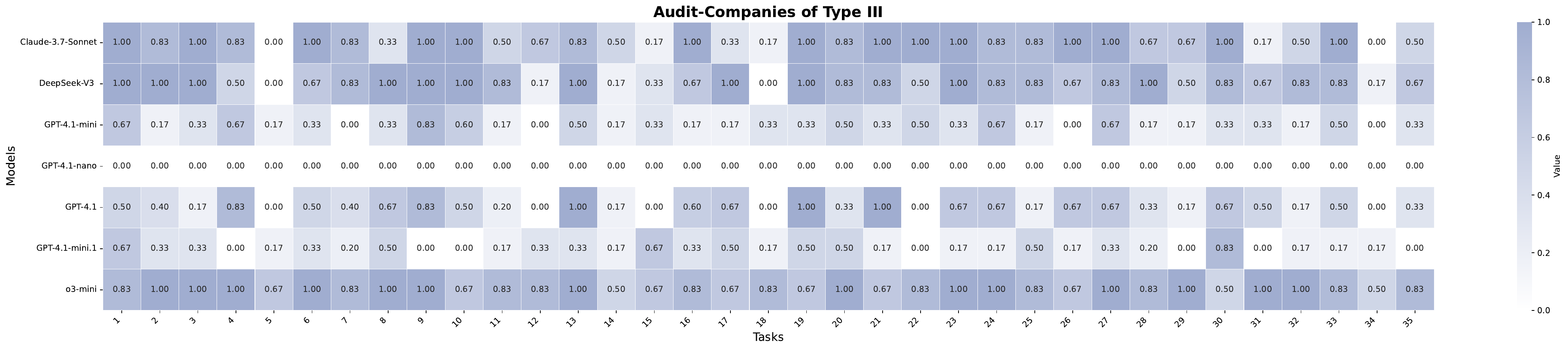}
    \caption{Accuracy of Type III Companies in Audit}
    \label{fig:AM13}
\end{figure}

\begin{figure}
    \centering
    \includegraphics[width=\textwidth]{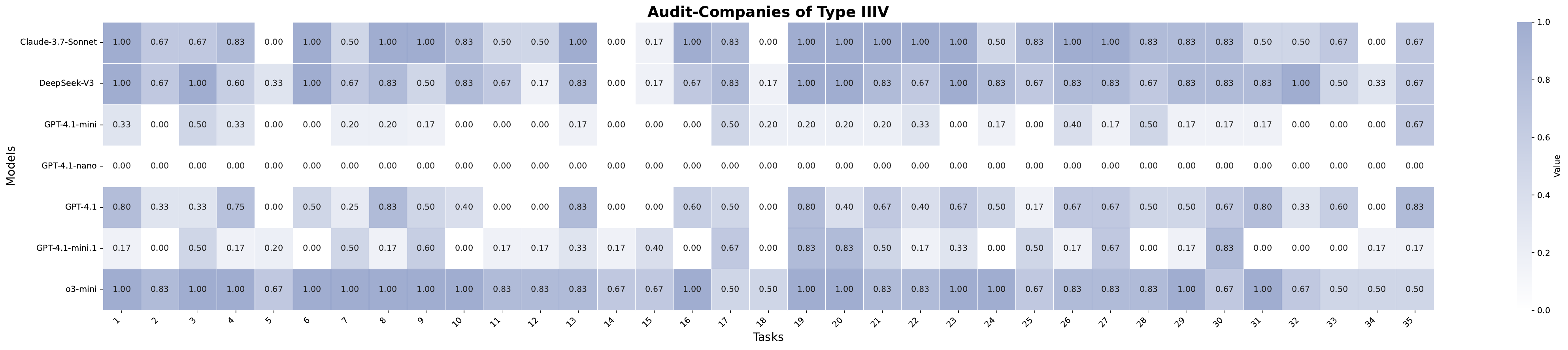}
    \caption{Accuracy of Type IIIV Companies in Audit}
    \label{fig:AM14}
\end{figure}

\begin{figure}
    \centering
    \includegraphics[width=\textwidth]{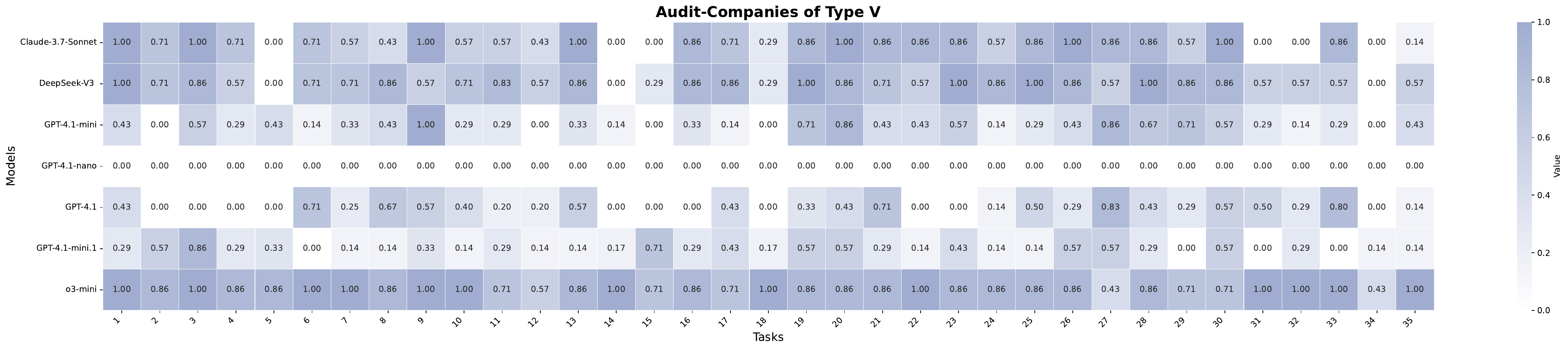}
    \caption{Accuracy of Type V Companies in Audit}
    \label{fig:AM15}
\end{figure}

\clearpage

\begin{figure}
    \centering
    \includegraphics[width=\textwidth]{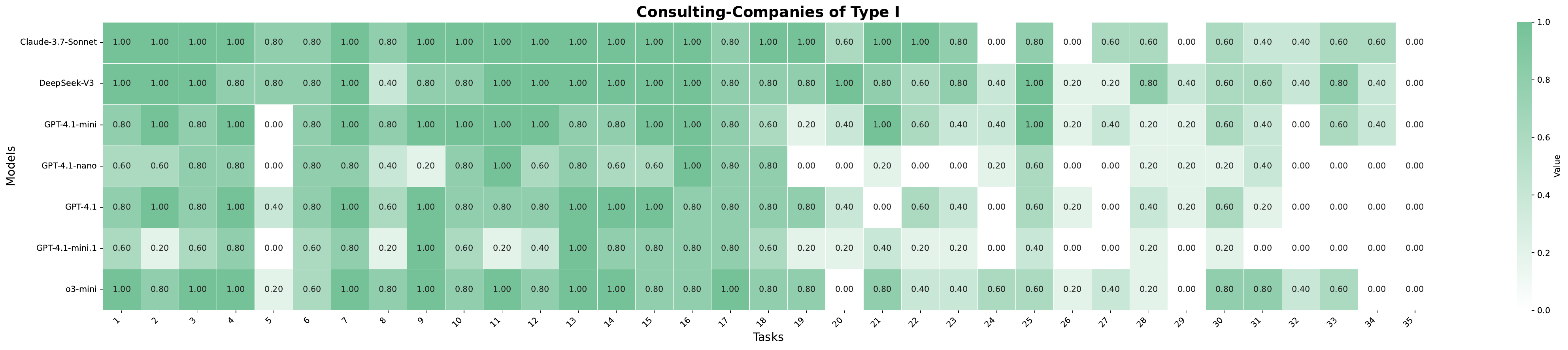}
    \caption{Accuracy of Type I Companies in Consulting}
    \label{fig:AM16}
\end{figure}

\begin{figure}
    \centering
    \includegraphics[width=\textwidth]{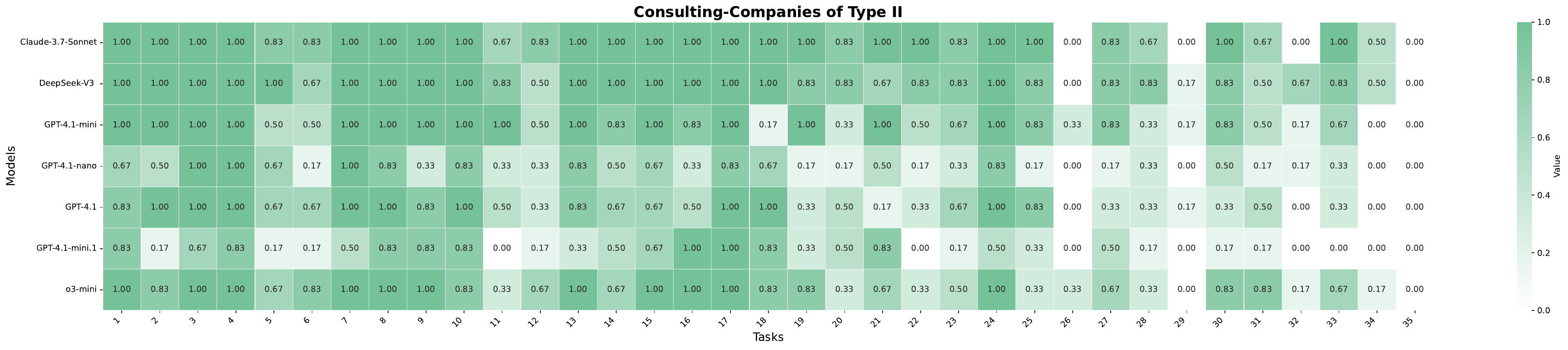}
    \caption{Accuracy of Type II Companies in Consulting}
    \label{fig:AM17}
\end{figure}

\begin{figure}
    \centering
    \includegraphics[width=\textwidth]{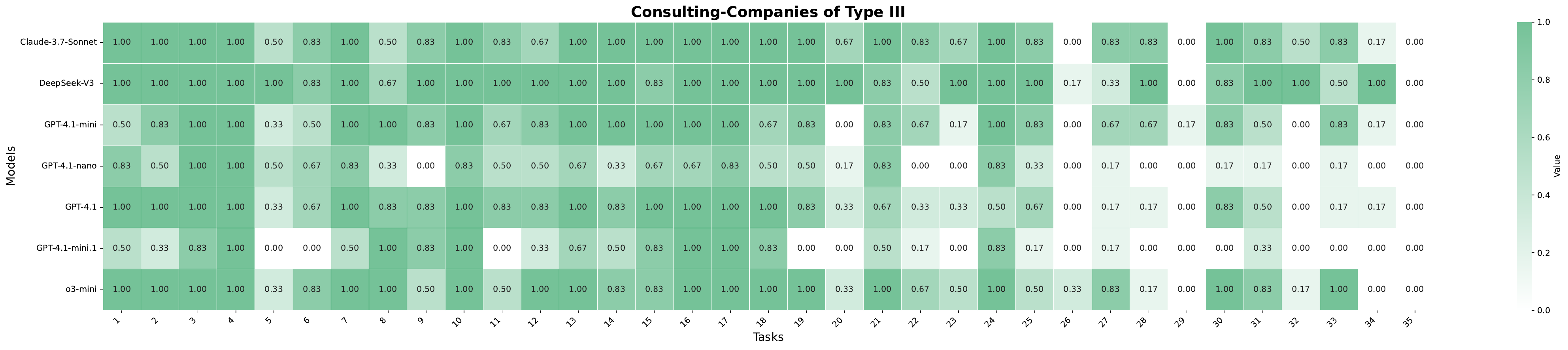}
    \caption{Accuracy of Type III Companies in Consulting}
    \label{fig:AM18}
\end{figure}

\begin{figure}
    \centering
    \includegraphics[width=\textwidth]{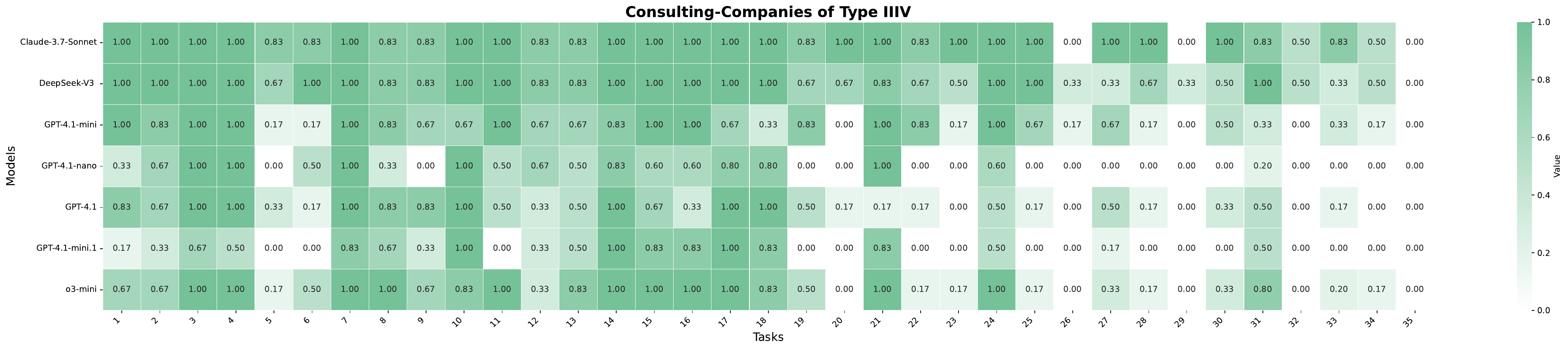}
    \caption{Accuracy of Type IIIV Companies in Consulting}
    \label{fig:AM19}
\end{figure}

\begin{figure}
    \centering
    \includegraphics[width=\textwidth]{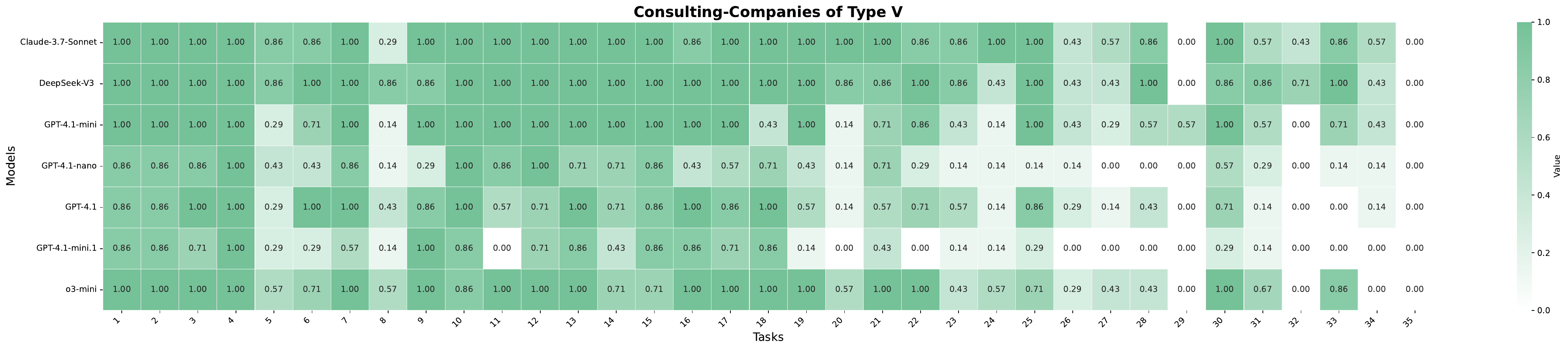}
    \caption{Accuracy of Type V Companies in Consulting}
    \label{fig:AM20}
\end{figure}

\clearpage

\section{Analysis of Reasoning Failures}
\label{app:reasoning_failures}

\begin{table}[htbp]
\centering
\caption{Comparative Analysis of Critical Errors Across Large Language Models}
\label{tab:model_error}
\resizebox{\textwidth}{!}{%
\begin{tabular}{>{\raggedright\arraybackslash}p{2cm}>{\raggedright\arraybackslash}p{3.5cm}>{\raggedright\arraybackslash}p{7cm}}
\toprule
\textbf{Model} & \textbf{Error Type} & \textbf{Description} \\
\midrule
\textbf{Claude-3.7-Sonnet} 
    & Floating Points Error & Incorrect rounding in Accounts Receivable\\
    & Reasoning Consistency & Inconsistent cash balance calculation\\
    & Methodology Chain Breaking & Missed bank transfer in cash calculation\\
\midrule
\textbf{DeepSeek-V3}
    & Factual Deviation & Added non-existent payable interest\\
    & Knowledge Retrieval Deficiency & Failed to identify audit date error\\
    & Floating Points Error & Lost precision in Taxes Payable\\
\midrule
\textbf{GPT-4.1}
    & Methodology Chain Breaking & Ignored depreciation in cash flow\\
    & Critical Data Omission & Omitted prepaid expenses in assets\\
\midrule
\textbf{GPT-4.1-mini}
   & Contextual Inconsistency & Misclassified audit opinion type\\
   & Logical Calculation Error & Inventory turnover ratio reversed\\
   & Multi-Step Calculation Error & Skipped Account Receivabl in Total Assets End Value calculation\\
\midrule
\textbf{GPT-4o-mini}
    & Factual Deviation & Misreported Account Payable as equity\\
    & Format Handling & Failed parsing complex raw transaction data\\
\midrule
\textbf{GPT-4.1-nano}
   & Logical Calculation Error & Fix asset purchase misclassified as cash inflow\\
   & Arithmetic Error & Trial balance summation error\\
   & Format Handling & Failed to perform cross financial statement analysis\\
\bottomrule
\end{tabular}%
}
\end{table}

\clearpage
\section{Costs}
\label{app:cost}
The costs of LLMs for running all the tasks once are calculated based on the input token and completion token counts with their corresponding prices, which are shown in Table~\ref{tab:cost}. 

\begin{table}[ht]
\centering
\caption{Cost for LLMs}
\label{tab:cost}
\begin{tabular}{c|c|c|c}
\toprule
Model & Prompt & Completion & Cost \\
\midrule
GPT-4o-mini&  (\$0.15/MTok) &  (\$0.6/MTok) & \$18.35 \\
GPT-4.1 & (\$2/MTok) & 8/MTok) & \$265.47\\
GPT-4.1-mini & (\$0.4/MTok) & (\$1.6/MTok) & \$49.87 \\
GPT-4.1-nano & (\$0.1/MTok) & (\$0.4/MTok) & \$11.22 \\
o3-mini&  (\$1.1/MTok) & (\$4.4/MTok) & \$186.94 \\
Claude-3.7-Sonnet&  (\$3/MTok) &  (\$15/MTok) &   \$396.07 \\
DeepSeek-V3-2503& (2RMB/MTok)&  (8RMB/MTok)&  239.27RMB \\

\bottomrule
\end{tabular}
\end{table}

\end{document}